\documentclass[final,onefignum,onetabnum]{siamart250211} % ARXIV
\usepackage{pdflscape}

\usepackage{lipsum}
\usepackage{amsfonts}
\usepackage{graphicx}
\usepackage{epstopdf}
\usepackage{algorithmic}
\ifpdf
  \DeclareGraphicsExtensions{.eps,.pdf,.png,.jpg}
\else
  \DeclareGraphicsExtensions{.eps}
\fi

\newsiamremark{remark}{Remark}
\newsiamremark{hypothesis}{Hypothesis}
\crefname{hypothesis}{Hypothesis}{Hypotheses}
\newsiamthm{claim}{Claim}
\newsiamremark{fact}{Fact}
\crefname{fact}{Fact}{Facts}

\headers{Weak Form Learning for Mean-Field PDEs: an Appl. to Insect Mvmnt.}{S. Minor, B. D. Elderd, B. Van Allen, D. M. Bortz, and V. Dukic}

\title{Weak Form Learning for Mean-Field Partial Differential Equations: an Application to Insect Movement\thanks{Submitted to the editors DATE.
\funding{This research was supported in part by the NIFA Biological Sciences Grant 2019-67014-29919, in part by the NSF Division Of Environmental Biology Grant 2109774, and in part by the NIGMS Division of Biophysics, Biomedical Technology and Computational Biosciences grant R35GM149335. This study was also funded in part by USDA grant 2019-67014-29919 and NSF grant 1316334 as part of the joint NSF–NIH–USDA Ecology and Evolution of Infectious Diseases program. This work utilized the Blanca condo computing resource at the University of Colorado Boulder. Blanca is jointly funded by computing users and the University of Colorado Boulder.}}}

\author{Seth Minor\thanks{Department of Applied Mathematics, University of Colorado, Boulder, CO 80309-0526.}
\and Bret D. Elderd\thanks{Department of Biological Sciences, Louisiana State University, Baton Rouge, LA 70803.}
\and Benjamin Van Allen\footnotemark[3]
\and David M. Bortz\footnotemark[2]
\and Vanja Dukic\footnotemark[2]}

\usepackage{amsopn}

\ifpdf
\hypersetup{
  pdftitle={Weak Form Learning for Mean-Field Partial Differential Equations: an Application to Insect Movement},
  pdfauthor={S. Minor, B. D. Elderd, D. Van Allen, D. M. Bortz, and V. Dukic}
}
\fi

\begin{document}

\maketitle

% REQUIRED
\begin{abstract}
Insect species subject to infection, predation, and anisotropic environmental conditions may exhibit preferential movement patterns. Given the innate stochasticity of exogenous factors driving these patterns over short timescales, individual insect trajectories typically obey overdamped stochastic dynamics. In practice, data-driven modeling approaches designed to learn the underlying Fokker-Planck equations from observed insect distributions serve as ideal tools for understanding and predicting such behavior. Understanding dispersal dynamics of crop and silvicultural pests can lead to a better forecasting of outbreak intensity and location, which can result in better pest management. In this work, we extend weak-form equation learning techniques, coupled with kernel density estimation, to learn effective models for lepidopteran larval population movement from highly sparse experimental data. Galerkin methods such as the Weak form Sparse Identification of Nonlinear Dynamics (WSINDy) algorithm have recently proven useful for learning governing equations in several scientific contexts. We demonstrate the utility of the method on a sparse dataset of position measurements of fall armyworms ({\it Spodoptera frugiperda}) obtained in simulated agricultural conditions with varied plant resources and infection status.
\end{abstract}

% REQUIRED

% REQUIRED

\newpage

\section{Introduction}
Insect populations subject to viral infection, predation, and anisotropic environmental conditions may exhibit preferential movement patterns \cite{Kareiva1983Oecologia, Turchin1998,GasqueVanOersRos2019CurrOpinInsectSci}. Given the inherent stochasticity of exogenous factors driving these patterns over short timescales, individual insect trajectories typically obey overdamped stochastic dynamics. In practice, modern data-driven modeling approaches designed to learn the underlying Fokker-Planck equations from observed insect distributions may serve as ideal tools for understanding, predicting, and in the case of economically important pests, controlling such behavior.

As many insect pest populations can be controlled by their natural viral or fungal pathogen \cite{Elderd2019Ecology, LiuKyleWangEtAl2025NatClimChang}, it is natural to ask what role, if any, the dispersal behavior may play in the epizootics.  Infectious agents that cause epizootics in insect populations can spread over time and space, with the spread of disease involving a contact between a susceptible individual and the pathogen. Such contact is either a direct contact between a susceptible and an infected individual, or due to contact with the pathogen contained in an environmental reservoir. In all of these situations, contact between the pathogen and the host requires movement. When considering a rapidly spreading disease or relatively local outbreak, disease transmission can be captured by a simple set of mass-action equations that assumes that movement is random and that any individual can come into contact with any other individual with equal probability \cite{KeelingRohani2008}. However, these simplifying assumptions do not hold for all outbreaks, where movement rate and direction may be non-random. Thus, to understand how a disease spreads across the landscape or between population centers, accurately capturing the movement dynamics of both infected and susceptible individuals becomes increasingly important. 

Similarly, it is important to understand whether, and to what extent, the disease status itself alters movement patterns.  For example, reduced  movement of infected individuals could slow down the disease spread as seen in the migratory monarch butterfly (\textit{Danaus plexippus}) \cite{BradleyAltizer2005EcolLett}. The level of infection or the pathogen's virulence can also be important factors in limiting infected host movement \cite{OsnasHurtadoDobson2015AmNat}. Yet, pathogens may also increase the movement rates of infected hosts in other settings \cite{Goulson1997Oecologia}. Regardless of whether infection increases or decreases host movement, its impact on disease transmission can be an important factor in determining disease spread  and optimal intervention strategies; see, e.g., \cite{BrosiDelaplaneBootsEtAl2017NatEcolEvol, Dwyer1992Ecology}.

The movement of individuals through the environment can be influenced by other factors besides infection status. For instance, organisms move through the environment to seek out food or other essential resources. Thus, movement can also depend on the habitat in which an organism finds itself \cite{Kareiva1983Oecologia}, and specifically for herbivores like forest or agricultural defoliators, the quality of food resources can affect movement across the landscape. Similarly, plants producing chemical or physical defenses in response to herbivory can negatively affect resource quality.  From a theoretical perspective, an increased level of such plant defenses could increase the rate at which herbivores spread across the landscape, as organisms move at a faster rate away from areas with poorer quality resources \cite{MorrisDwyer1997AmNat}.

Given that a number of herbivores that are agricultural or silvicultural pests can cause a great deal of damage \cite{LiuKyleWangEtAl2025NatClimChang,Elderd2019Ecology}, understanding movement dynamics becomes particularly important from an applied perspective. %As densities increase and resources diminish in the local area, these herbivores will spread across the landscape as they continue to defoliate a field or forest. Infestations of some of these pests can be held in check as their densities rise by natural enemies such as their pathogens. Thus, 
Understanding the movement dynamics of these pests as they travel through the environment can lend important insight into the spatial dynamics of pest infestations and how to control them. This is particularly true for the fall armyworm (\textit{Spodoptera frugiperda}), a world-wide agricultural pest whose larval stage readily feeds on a wide variety of crops.

From a mathematical modeling perspective, Galerkin approaches such as the Weak Identification of Nonlinear Dynamics (WSINDy) algorithm \cite{MessengerBortz2021MultiscaleModelSimul, MessengerBortz2021JComputPhys} have recently proven useful for learning sophisticated and interpretable governing equations directly from empirical data in several relevant biological contexts. For example, \cite{MessengerDwyerDukic2024JRSocInterface} introduced a weak-form hybrid modeling paradigm to the context of epizootics for the North American Spongy Moth (\textit{Lymantria dispar dispar}). Moreover, \cite{MessengerBortz2022PhysicaD} demonstrated that WSINDy can retrieve accurate mean-field governing equations from noisy interacting particle data.

In this paper, we use weak-form equation learning techniques \cite{MessengerBortz2021JComputPhys, MessengerBortz2022PhysicaD,MessengerDwyerDukic2024JRSocInterface}, coupled with kernel density estimation, to learn effective models for insect population movement from experimental data. We demonstrate the utility of the method on a sparse set of position measurements of the fall armyworm obtained over several regimes of interest, with varied environmental (two plant genotypes) and infection conditions (infected/not infected larvae). We learn the best effective population movement model for each of the four experimental settings, and compare the individual results in order to assess whether and how infection status and plant genotype (i.e., resource quality) affect dispersal. 

We organize the paper as follows. In Section~\ref{Sec:MB}, we review the experimental setting, and give an overview of the mathematics of the weak form methodology used for the analysis. In Section~\ref{Sec:results}, we discuss the learned dispersal models and compare them across the infection status and soybean genotypes. Finally, Section~\ref{Sec:disc} provides concluding remarks. Supplemental results and details about our numerical implementation are given in the appendix.

\section{Methods and Background}\label{Sec:MB}
In this section, we provide a brief biological background in \S\ref{sec:background} before giving an overview of our experimental setup in \S\ref{sec:experimental_methods}, as well as the relevant mathematical and numerical background behind our methods and their implementation in \S\ref{sec:mathematical_methods} and \S\ref{sec:numerical_methods}, respectively. Our approach couples kernel density estimation with the WSINDy methodology of \cite{MessengerBortz2022PhysicaD} to learn effective models for lepidopteran larval population movement from highly sparse and irregularly-spaced experimental data exploring various combinations of plant resource quality and infection status.

\begin{figure*}
    \centering
    \fbox{\includegraphics[width=0.38\linewidth]{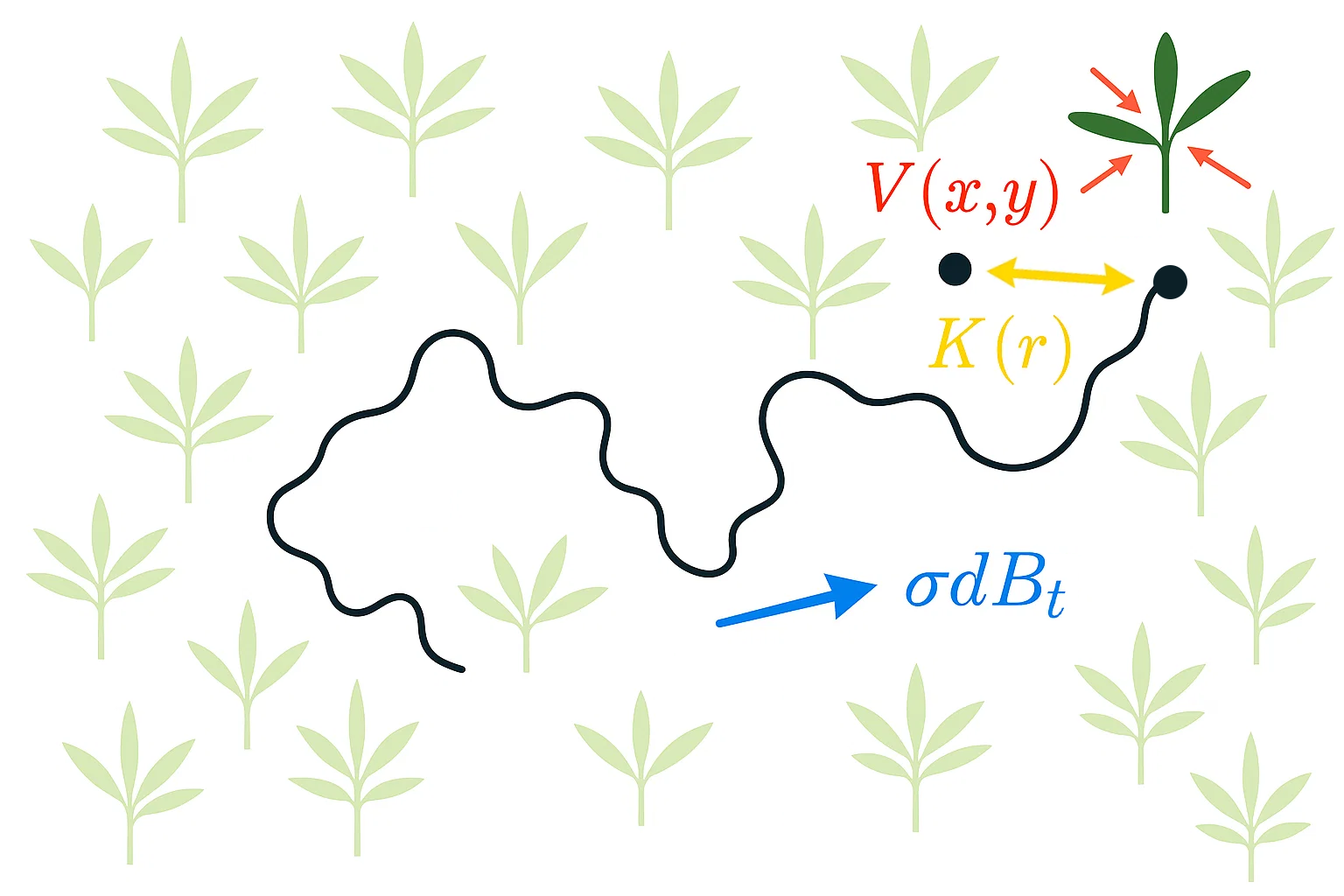}}
    \qquad
    \raisebox{-1.5mm}{\includegraphics[width=0.28\linewidth]{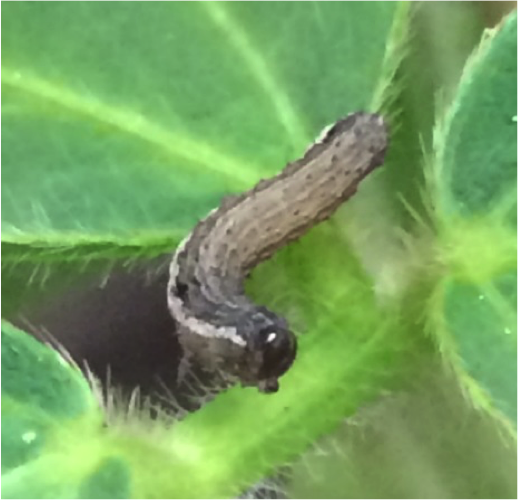}}
    \caption{(Left) Illustrating the forces at play in eq.~(\ref{eq:SDE}). (Right) A fall armyworm larva.}
    %\caption{(Left) Illustrating the forces at play in eq.~(\ref{eq:SDE}). (Right) A fall armyworm larva (photo credit: Ben Van Allen).}
    \label{fig:FAW_picture}
\end{figure*}

\subsection{Biological Background }\label{sec:background}
Our tritrophic pathogen-herbivore-plant study system consists of: (1) a species-specific lethal baculovirus known as \textit{Spodoptera frugiperda multiple-nucleopolyhedrovirus} (SfMNPV), (2) an agricultural pest, the larval stage of the fall armyworm (Figure~\ref{fig:FAW_picture}), which serves as the disease host, and (3) one of the two genotypes/varieties of soybean plant (Glycine max) on which the host feeds, which vary in resource quality. 

%For the host to become infected, it must consume a lethal dose of SfMNPV viral particles. These viral particles (also known as occlusion bodies or OBs) can be found on leaf surfaces, left over by a previously infected caterpillar cadaver tissue. When fall armyworm larvae feed on contaminated leaf tissue, they inadvertently consume enough viral particles to become lethally infected.  At the conclusion of the infection process, the larvae disintegrate (lyse) thus spreading newly formed OBs on nearby leaf tissue \cite{Elderd2013PLoSPathog}. %The  process from consumption to lysing takes about a week.

The fall armyworm is a multivoltine agricultural pest (i.e., multiple generations per year) that goes through six larval growth stages or instars. They are polyphagous and consume several different agricultural crops including soybeans. This pest is native to North and South America but has recently been introduced to Africa and Asia, where it is currently causing billions of dollars of damage \cite{Stokstad2017Science,RaneWalshLenanckerEtAl2023SciRep}. Their life cycle begins when the larvae emerge from their egg casings and begin to feed on leaf tissue. Once they have reached the sixth larval instar, the larvae pupate in the soil. After pupation, they eclose and mate to continue the next generation. During the winter, freezing temperatures kill the pupae before they eclose. In North America, the fall armyworm overwinters in southern Texas and Florida where the pupae can survive during the winter months. Over the growing season from spring to fall, the adult moths steadily migrate northward and can cause infestations as far north as southern Canada during the late summer and early fall. At a more local scale, larvae will move from field to field as resources run low and, thus, spread across the landscape as they continue to devastate crops \cite{Sparks1979FlaEntomol}.

Fall armyworm populations traditionally go through boom-and-bust cycles where the population collapses are often driven by the baculovirus. During the collapse, upwards of 60\% of a population can be infected with SfMNPV \cite{Fuxa1982EnvironEntomol}. The infection cycle begins when recently emerged first instars become lethally infected.  The virus stops the molting process and the infected first instars cease to grow. After a number of days (this number depends on temperature), the infected larvae liquefy and lyse, spreading viral particles onto the leaf tissue that they are feeding on. Uninfected larvae, which have grown to the fourth instar by this time, feed on the contaminated leaf tissue and the infection cycle continues. Due to UV light exposure, the virus will degrade over time \cite{Elderd2013PLoSPathog}, reducing the risk of environmental exposure. Since SfMNPV is species specific, the virus can be and has been used as a biocontrol agent [\url{agbitech.us/fawligen}].

It is well known that pathogen infection can cause changes in animal behavior and, particularly, in insects \cite{GasqueVanOersRos2019CurrOpinInsectSci}.  The behavioral changes include those exhibited by "zombie" ants infected with fungal pathogens.  Prior to death, infected individuals climb up in the vegetation to help facilitate the spread of fungal spores from the fruiting body that emerges from their corpse \cite{HughesAndersenHywel-JonesEtAl2011BMCEcol}. Similar behavior is seen in lepidopteran larvae infected with baculovirus where infected individuals climb upwards prior to death to facilitate the distribution of viral particles in the environment \cite{HooverGroveGardnerEtAl2011Science, Goulson1997Oecologia, GasqueVanOersRos2019CurrOpinInsectSci}. Baculovirus infections can also increase the dispersal distance of infected larvae \cite{Goulson1997Oecologia}. However, the distance and speed of dispersal can be dependent on larval stage as well as the time since becoming infected \cite{VasconcelosCoryWilsonEtAl1996BiolControl}. Less well-known is how infection status \textit{and} resource quality of the host plant affect dispersal.

\subsection{Experimental Methods}\label{sec:experimental_methods}

%\subsubsection{Data Collection}

One of the many agricultural crops that the fall armyworm feeds on is soybean \cite{PerucaCoelhoDaSilvaEtAl2018Arthropod-PlantInteract}. Soybeans come in numerous genotypes/varieties and these varieties differ in their chemical and physical defenses that they employ against herbivores, thus having different effects on larval leaf consumption and virus-induced mortality \cite{ShikanoShumakerPeifferEtAl2017Oecologiaa}. Specifically, differences in the chemical constituency of the plant defense may affect infection rates and the production of viral particles by an infected larva.  These defenses against herbivory also affect the quality of the leaf tissue and can negatively impact growth rates in the fall armyworm \cite{ShikanoShumakerPeifferEtAl2017Oecologiaa}. Consequently, this may lead to changes in dispersal rates amongst individual larvae.

\newpage

To directly quantify how infection status and resource quality alter movement dynamics, we conducted a series of eight experiments where we measured the movement of fall armyworm larvae across an artificial landscape in the lab. The landscape consisted of four 175 cm $\times$ 175 cm plots with 45 evenly-spaced mature soybean plants with at least five tri-foliate leaves.  In order to simulate common farming practices, the plants were organized into five rows of nine plants in each plot.  We varied resource quality by using two  varieties of soybean that differed in their constitutive anti-herbivore defenses \cite{UnderwoodRausherCook2002Oecologia, ShikanoShumakerPeifferEtAl2017Oecologiaa}. These varieties were \textit{Stonewall}, which we considered as having a relatively high constitutive defense, and \textit{Gasoy}, which we considered as having a relatively low constitutive defense \cite{UnderwoodMorrisGrossEtAl2000Oecologia,UnderwoodRausherCook2002Oecologia}. The Stonewall variety could thus be considered a poor-quality resource as compared to the Gasoy variety.  

At the start of the experiment, we placed 20 fourth-instar larvae at the center of each of the four plots, on a single soybean plant. Each plot was planted with either the  Stonewall or Gasoy variety, and received either infected or uninfected larvae. After the start of the experiment, we measured the location of individual larvae along $x$, $y$, and $z$-axes at eight non-uniformly spaced times (i.e., $0, \, 1, \, 2, \, 4, \, 8, \, 16, \, 24,$ and $48$ hours). The $(x,y)$ measurements correspond to the location of the larvae in the plot, while the $z$-axis measurement indicates the height of the larva, with zero corresponding to the soil-level and any point above zero being the location of the larvae on a soybean plant. For each combination of plant variety and infection status, we conducted the experiment two times. % for a total of 40 larvae measured over the duration of the experiment. 
The empirical distributions are visualized for each observation time in Figure~\ref{fig:caterpillars_3d_density} (black dots) and in Figure~\ref{fig:caterpillars_violin_r}. Further details of the experimental setup can be found in the Appendix; see \S\ref{App:experiment_details} in particular.

\subsection{Training Dataset}

Although three-dimensional $(x,y,z)$ position measurements were obtained experimentally, due to the inherent sparsity of the data we focus on effective surface dispersal models by neglecting the vertical ($z$) components. Our training data thus consist of the set of two-dimensional position measurements, \begin{align*}
    \mathbf{X}_t := \big\{\mathbf{x}_t^{i}\big\}_{i=1}^{N_t}
    \quad \text{where} \quad
    \mathbf{x}_t^{i} := \big(x_t^i, \, y_t^i\big) \in \mathbb{R}^2,
\end{align*} of the $N_t$ larvae taken at times $t \in \{t_0 = 0 , \dots, t_{\textsc{f}} = 48\}$. All time measurements are recorded in hours and all length measurements in centimeters. We define a \lq{super-imposed}' empirical position distribution, \begin{align}\label{eq:empirical_distribution}
    \mu(\boldsymbol{x}; \mathbf{X}_t) := \frac{1}{N_t} \sum_{i = 1}^{N_t} \delta\big( \boldsymbol{x} - \mathbf{x}_t^{i} \big),
\end{align} where $\delta(\boldsymbol{x}) := \delta(x)\delta(y)$ denotes a Dirac delta distribution centered at the origin. 

The larvae are separated into four distinct and isolated planter domains $\Omega_1$, $\Omega_2$, $\Omega_3$, and $\Omega_4$, where each spatial domain $\Omega_j = [0,175]^2$ has identical dimensions and each domain contains plant resources evenly spaced into five rows and nine columns. To assess population movement dynamics in varied environmental conditions and infection regimes, each distinct planter $\Omega_j$ represents a separate experimental setting, containing  a unique combination of resource genotype (\textit{Stonewall} or \textit{Gasoy}) and larval infection status (\textit{infected} or \textit{not infected}). To distinguish between control population and experiment replicate number, we define analogous empirical measures $\mu(\boldsymbol{x}; \mathbf{X}_t^{j,k})$ for each plot index $j = 1, 2, 3, 4$ and replicate index $k = 1, 2$, where $\mathbf{X}_t^{j,k} := \{\mathbf{x}_t \in \Omega_j\}$. The super-imposed distribution in eq.~(\ref{eq:empirical_distribution}) is recovered by computing \begin{align*}
    \mu(\boldsymbol{x}; \mathbf{X}_t) = \sum_{k=1}^{2}
    \sum_{j=1}^{4} \mu\big(\boldsymbol{x}; \mathbf{X}_t^{j,k}\big).
\end{align*} We order the cases as follows: \textit{not infected} with \textit{Stonewall} ($j=1$); \textit{not infected} with \textit{Gasoy} ($j=2$); \textit{infected} with \textit{Stonewall} ($j=3$); and \textit{infected} with \textit{Gasoy} ($j = 4$). We again note that the position distributions $\mathbf{X}_{t_0}, \dots, \mathbf{X}_{t_{\textsc{f}}}$ are recorded using non-uniform temporal increments $\Delta t_n$, with $t_n \in \{0, \, 1, \, 2, \, 4, \, 8, \, 16, \, 24, \, 48\},$ measured relative to the beginning of each experiment.

\subsection{Mathematical Methods}\label{sec:mathematical_methods}
Here, we present and formalize the mathematical modeling methodology that will be used throughout. Our primary interest will be to develop an accurate partial differential equation (PDE) model for larval dispersal, by means of an evolution equation for the probability density (i.e., a \textit{Fokker-Planck equation}), with the secondary aim of understanding the influence of plant genotype and infection status on movement dynamics. Our underlying assumption is that each individual disperses according to an overdamped and biased random walk $\mathbf{x}^i_t$, where the drift $\mathbb{E}[\mathbf{x}^i_t]$ can be attributed to repulsive or attractive interactions between individuals and reactions to environmental features (e.g., plant resources). Under this assumption, the corresponding `coarsed-grained' model for the probability distribution obeys analogous advection-diffusion dynamics, which we learn in further sections using a weak-form data-driven approach.

Our approach is motivated by a broad tradition of mathematical methods for dispersal modeling in ecology. Interested readers are referred to, e.g., the reviews given in \cite{HolmesLewisBanksEtAl1994Ecology} and \cite{OthmerDunbarAlt1988JMathBiology} for more information. We also note that diffusion coefficients for such models have been experimentally measured for various insect species in \cite{Kareiva1983Oecologia}.

% \begin{figure*}
%     \centering
%     \fbox{\includegraphics[width=0.38\linewidth]{pictures/random_walk_potential.png}}
%     \qquad
%     \raisebox{-1.5mm}{\includegraphics[width=0.28\linewidth]{pictures/FAWStageStructureCanni.png}}
%     \caption{(Left) Illustrating the forces at play in eq.~(\ref{eq:SDE}). (Right) A fall armyworm larva (photo credit: Ben Van Allen).}
%     \label{fig:FAW_picture}
% \end{figure*}

\subsubsection{Governing Equations}
Mathematically, we treat the ensemble of larval positions $\mu(\boldsymbol{x};\mathbf{X}_t)$ as the empirical distribution of a stochastic interacting particle system $\mathbf{X}_t$, and use a sparse regression approach inspired by \cite{MessengerBortz2022PhysicaD, MessengerWheelerLiuEtAl2022JRSocInterface} to discover a governing equation for the probability density function $u(\boldsymbol{x},t)$. This probability density function can be approximated as a histogram of positions over $N_{\mathcal{B}}$ disjoint, equal-area bins, $\mathcal{B}_k = [\tilde{x}_k, \tilde{x}_k + \Delta{\tilde{x}}] \times [\tilde{y}_k, \tilde{y}_k + \Delta{\tilde{y}}] \subset \mathbb{R}^2$, \begin{align*}
    \hat{u}(\boldsymbol{x},t)
    :=
    \big(G * \mu(\,\cdot\,, \,\mathbf{X}_t)\big)(\boldsymbol{x}),
    \quad \text{with} \quad
    G(\boldsymbol{x}) := 
\sum_{k=1}^{N_{\mathcal{B}}} \frac{\mathbf{1}_{\mathcal{B}_k}\!(\boldsymbol{x})}{|\mathcal{B}|}.
\end{align*} Following \cite{MessengerBortz2022PhysicaD, HolmesLewisBanksEtAl1994Ecology}, we assume that each trajectory $\mathbf{x}_t^i  \in \mathbf{X}_t$ is a  random variable governed by a McKean-Vlasov stochastic differential equation (SDE) of the form \begin{align}\label{eq:SDE}
    d\mathbf{x}_t^i = -\!\Big(\nabla \mathcal{V}\big(\mathbf{x}_t^i\big) + \nabla\mathcal{K}*\mu(\boldsymbol{x};\mathbf{X}_t)\Big) dt
    %\\
    %&\quad
    + \boldsymbol{\sigma} \, d\mathbf{B}_t^i,
\end{align} where each vector $d\mathbf{B}_t^i \sim \mathcal{N}(0, dt\mathbf{I})$ is a Wiener process, the matrix $\boldsymbol{\sigma} \in \mathbb{R}^{2 \times 2}$ governs the diffusivity of the process, and $\mathcal{V}$ and $\mathcal{K}$ are effective scalar-valued \textit{environmental} and \textit{interaction potentials}, respectively. Conceptually, our underlying assumption is that each individual $\mathbf{x}^i$ responds to \lq{forces}' $-\nabla \mathcal{K}$ and $-\nabla \mathcal{V}$ exerted by other individuals and by the environment, respectively. In the absence of these forces, such trajectories reduce to purely random walks, with $d\mathbf{x}_t^i = \boldsymbol{\sigma} d\mathbf{B}_t^i$.

\newpage

We now consider the high resolution limit of the empirical distribution $\mu(\boldsymbol{x}; \mathbf{X}_t)$ of trajectories $\mathbf{X}_t$ governed by the SDE in eq.~(\ref{eq:SDE}). As the number of particles $N_t$ increases and bin area $|\mathcal{B}|$ shrinks, the limiting probability density, 
\begin{align*}
    u(\boldsymbol{x},t) := \lim_{N_t \rightarrow \infty} \lim_{|\mathcal{B}| \rightarrow 0} \hat{u}(\boldsymbol{x},t),
\end{align*} obeys a nonlinear Fokker-Planck equation driven by analogous advective and diffusive mechanisms, \begin{align}\label{eq:PDE}
    u_t &= \nabla \cdot \Big( u \big(\nabla \mathcal{V} + \nabla\mathcal{K} \! * \! u\big) + \mathbf{D} \nabla u\Big).
\end{align} Here, the \textit{diffusion matrix} is defined as $\mathbf{D} := \frac{1}{2} \boldsymbol{\sigma} \boldsymbol{\sigma}^T$, implying that $\mathbf{D}$ is a symmetric matrix, and the interaction term involves a spatial convolution given explicitly by \begin{align*}
    \big(\nabla\mathcal{K} * u\big)(\boldsymbol{x},t)
    :=
    \int\!\!\!\!\int_{\Omega} \nabla\mathcal{K}\big(\left\|\boldsymbol{x} - \boldsymbol{x}'\right\|_2\big) \, u\big(\boldsymbol{x}',t\big) \, dx' dy'.
\end{align*} Formally, eq.~(\ref{eq:PDE}) is to be understood in a weak sense, i.e., in terms of $\mu(\boldsymbol{x}; \mathbf{X}_t)$. For a discussion of how and under what conditions the SDE eq.~(\ref{eq:SDE}) converges to the PDE eq.~(\ref{eq:PDE}), we refer the reader to \cite{MessengerBortz2022PhysicaD}.

\subsubsection{Structural Assumptions}
Beyond our fundamental assumption that the larvae follow biased random walks according to eq.~(\ref{eq:SDE}), we further assume that:
\begin{enumerate}
    \item diffusion is  \textit{homogeneous} but potentially \textit{anisotropic}; i.e., each element $D_{ij}$ is a distinct constant that does not depend on space or time;

    \item biases in empirical diffusion coefficient estimates $\hat{D}_{ij}$ resulting from larvae spreading to the edge of the experimental plots $\Omega_j$ at later times ($t \geq 24$) are sufficiently small to be ignored;
    
    \item the environmental potential term $-\nabla\mathcal{V}$ accounts for all dynamics resulting from a non-homogeneous environment (e.g., attraction to plant resources);

    \item the interaction potential term $-\nabla\mathcal{K}$ accounts for all `social' interactions (e.g., repulsion or attraction due to cannibalism \cite{VanAllenDillemuthDukicEtAl2023Oecologia} or clumping, respectively), thus representing an effective  \lq{pressure}' mechanism;

    \item the (time-dependent) number of larvae in each plot, $N_t^j$, is sufficiently large that the dynamics of the aggregate model can be reasonably expected to approximate the true aggregate dynamics.
\end{enumerate} Note that in the experimental data, the separate control populations $\mathbf{X}_t^{j,k}$ cannot physically interact with each other (e.g., the \textit{infected} class is always separated from \textit{non-infected}); thus, we do not learn effective interaction potentials $\mathcal{K}$ for any of the cases in which we combine training data from several experiments (for more information, see Table~\ref{table:full_model} and Table~\ref{table:results_by_run_number}).

Finally, we pause to mention several features of the empirical data which particularly influence our data-driven modeling methodology. Unlike in \cite{MessengerWheelerLiuEtAl2022JRSocInterface}, only the ensemble of positions $\mathbf{X}_t$ is known, as there is no information about how the individual trajectories $\mathbf{x}^i_t$ persist over time. In addition, in our work $N_t$ is not constant, as larvae can be lost or may simply not be found within the 15-minute search window (see \S\ref{App:experiment_details} for more details). Furthermore, our data are significantly sparser, both in total count $N_t$ and in number of time snapshots $t_n$, than the minimum of $\mathcal{O}(10^3)$ samples assumed in \cite{MessengerBortz2022PhysicaD}.

\subsubsection{Nondimensionalization}
To rewrite the PDE in eq.~(\ref{eq:PDE}) in a unit-independent format in which the relative magnitudes of the various contributions to the dynamics can be sensibly compared, we consider a symmetric and positive-definite change of coordinates $\mathbf{A} = \mathbf{A}^T$ of the form \begin{align*}
    \boldsymbol{x} = \mathbf{A}\boldsymbol{\xi},
    \quad \text{along with} \quad
    t = \tau t_c,
\end{align*} where the $A_{ij}$ and $\tau_c$ are constant characteristics scales resulting in dimensionless coordinates $(\boldsymbol{\xi}, \tau)$. Similarly, we consider rescaled dimensionless variables $U$, $V$, and $K$ defined by \begin{align*}
    U(\boldsymbol{\xi}, \tau) := U_c^{-1} \, u\big(\boldsymbol{x}(\boldsymbol{\xi}), \, t(\tau)\big),
    \quad \text{with} \quad
    \begin{cases}
        V(\boldsymbol{\xi}) := V_c^{-1} \, \mathcal{V}\big(\boldsymbol{x}(\boldsymbol{\xi})\big),
        \\
        K(\boldsymbol{\xi}; \boldsymbol{\xi}') := K_c^{-1} \, \mathcal{K}\big(\boldsymbol{x}(\boldsymbol{\xi}); \, \boldsymbol{x}'(\boldsymbol{\xi}')\big).
    \end{cases}
\end{align*} We assume that the dimensional constants $U_c$, $V_c$, and $K_c$ are chosen such that the corresponding dimensionless gradients are of size $\mathcal{O}(1)$. A calculation included in \S\ref{sec:nondimensionalization_details} shows that substitution of the rescaled quantities into eq.~(\ref{eq:PDE}) then yields a nondimensionalized PDE of the form \begin{align}\label{eq:nondimmed_pde}
    U_{\tau} &= \bar{\nabla} \cdot \Big( U \big(\boldsymbol{\Pi}_V\bar{\nabla} V + \, \boldsymbol{\Pi}_K\bar{\nabla} K \star U\big) + \, \boldsymbol{\Pi}_D \bar{\nabla} U\Big).
\end{align} where here the operators $\bar{\nabla}$ and $\star$ are taken with respect to rescaled variables. The $\boldsymbol{\Pi}_i$ matrices in eq.~(\ref{eq:nondimmed_pde}) above represent dimensionless transformations defined by \begin{align}\label{eq:nondim_groups}
    \boldsymbol{\Pi}_V = t_c V_c \, \mathbf{\Lambda}^{-1},
    \quad
    \boldsymbol{\Pi}_K = t_c K_c U_c \, |\mathbf{\Lambda}|^{\frac{1}{2}} \mathbf{\Lambda}^{-1},
    \quad \text{and} \quad
    \boldsymbol{\Pi}_D = t_c \, \mathbf{A}^{-1} \mathbf{D} \mathbf{A}^{-1},
\end{align} where we've defined the Gram matrix $\mathbf{\Lambda} := \mathbf{A}^T \mathbf{A}$.

\subsubsection{Mathematical Theory}
\begin{figure*}
    \centering
    \includegraphics[width=\linewidth]{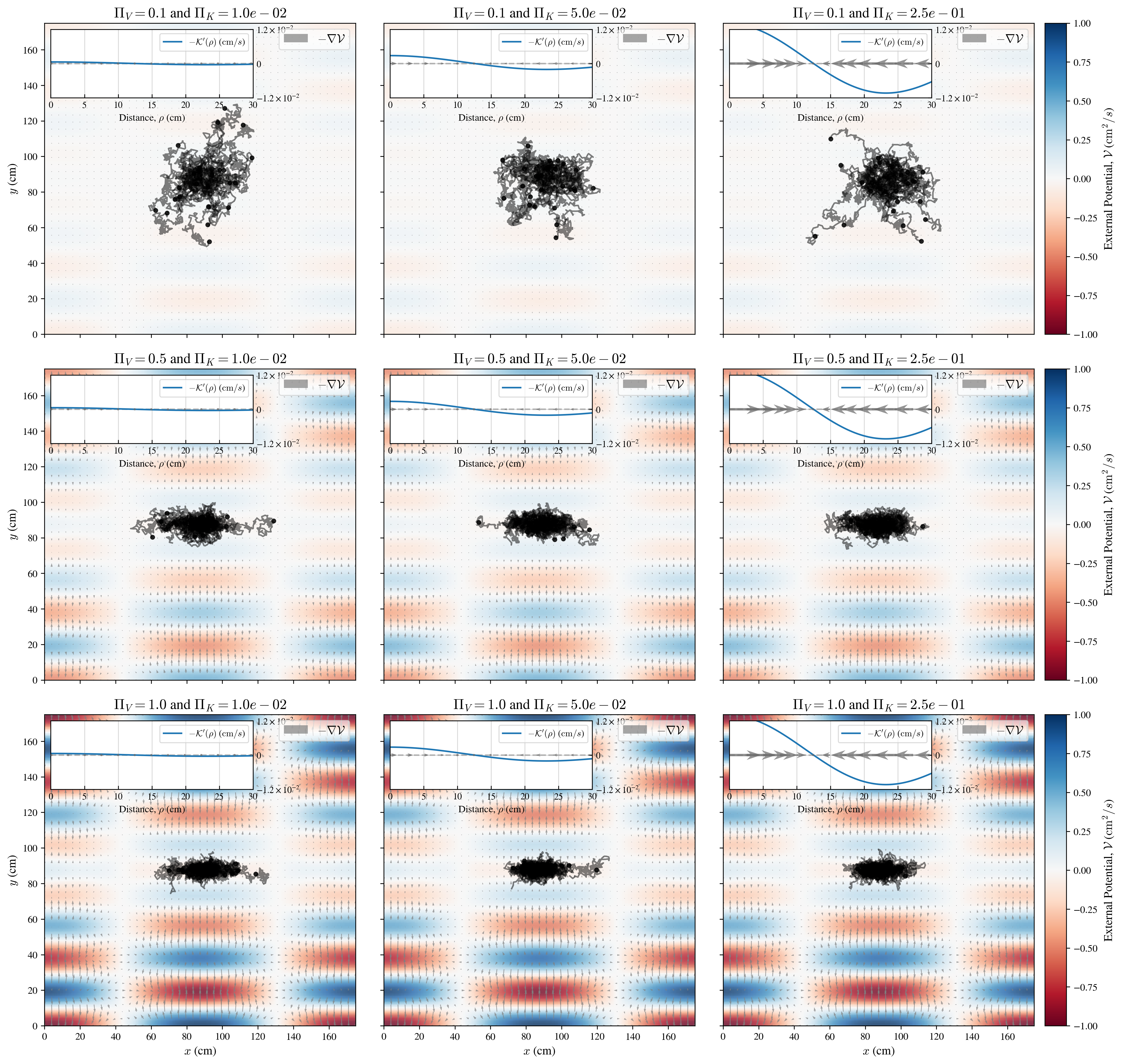}
    \caption{Illustrating the dynamics that are possible under an SDE consistent with eq.~(\ref{eq:simplified_diffusion_focused_pde}) in various parameter regimes. Here, we fix the diffusion strength $\Pi_D = 1$ and incrementally increase the potential strengths $\Pi_V$ and $\Pi_K$; see \S\ref{sec:additional_implementation_details}.}
    \label{fig:caterpillar_trajectories}
\end{figure*}

Analytical results about the rescaled PDE in eq.~(\ref{eq:nondimmed_pde}) become tractable in several parameter regimes. In this section, we discuss two illustrative examples of such regimes: (1) $\|\boldsymbol{\Pi}_V\|, \|\boldsymbol{\Pi}_K\| \approx 0$ and (2) $\|\boldsymbol{\Pi}_K\| \approx 0$ with $\mathbf{D} = D\mathbf{I}$. In any case, we note that a natural choice of diffusion-centric coordinates is given by $\mathbf{A} = (\mathbf{D} t_c)^{\frac{1}{2}} = (\frac{1}{2} t_c)^{\frac{1}{2}} \, \boldsymbol{\sigma}^{\star}$, where the matrix $\boldsymbol{\sigma}^{\star}$ represents the unique symmetric-positive-definite square root of the diffusion matrix $\mathbf{D}$, which in physically realistic cases is also symmetric-positive-definite. In this system of coordinates, the dimensionless groups in eq.~(\ref{eq:nondim_groups}) simplify to \begin{align*}
    \boldsymbol{\Pi}_V = V_c \, \mathbf{D}^{-1},
    \quad
    \boldsymbol{\Pi}_K = t_c K_c U_c \, |\mathbf{D}|^{\frac{1}{2}} \mathbf{D}^{-1},
    \quad \text{and} \quad
    \boldsymbol{\Pi}_D = \mathbf{I},
\end{align*} producing a non-dimensionalized PDE of the form \begin{align}\label{eq:diffusion_focused_pde}
    U_{\tau} &= \bar{\nabla} \cdot U \big(\boldsymbol{\Pi}_V\bar{\nabla} V + \, \boldsymbol{\Pi}_K\bar{\nabla} K \star U\big) \, + \, \bar{\Delta}U.
\end{align} Since one intuitively expects overdamped dynamics in the context of insect dispersal, the above formulation of the dynamics is `natural' in the sense that it gives unit weight to the diffusion term $\bar{\Delta}U$.\footnote{Note that the mean square displacement of an isotropic two-dimensional Brownian particle grows like $\mathbb{E}[\|\mathbf{x}_t - \mathbf{x}_0\|^2_2] = 4Dt$, with the mean displacement growing like $\mathbb{E}[\|\mathbf{x}_t-\mathbf{x}_0\|_2] = \sqrt{\pi Dt}$.} In this coordinate system, the dynamics are then characterized by the relative strengths of the remaining dimensionless groups $\boldsymbol{\Pi}_V$ and $\boldsymbol{\Pi}_K$; see Figure~\ref{fig:caterpillar_trajectories} for a comparison of the dynamics in various parameter regimes.

\newpage

We begin by considering a regime where the exogenous forces acting on individuals are negligible in comparison to diffusive forces (i.e., with $\|\boldsymbol{\Pi}_V\|, \, \|\boldsymbol{\Pi}_K\| \ll 1$), so that the non-dimensionalized SDE (cf. eq.~(\ref{eq:SDE})) and PDE in eq.~(\ref{eq:diffusion_focused_pde}) are, respectively, well-approximated by \begin{align*}
    d\boldsymbol{\xi}_{\tau}^i \approx \sqrt{2} \, d\mathbf{B}_{\tau}^i,
    \quad \text{and} \quad
    U_{\tau} \approx \bar{\Delta}U.
\end{align*} In this case, a general solution to the rescaled PDE can be approximated by convolving the initial distribution $U_0(\boldsymbol{\xi})$ against a heat kernel, $U(\boldsymbol{\xi},\tau) \approx (U_0 * H_{\mathbf{I}})(\boldsymbol{\xi},\tau)$, where \begin{align*}
    H_{\mathbf{M}}(\boldsymbol{x},t) = \frac{1}{4\pi t |\mathbf{M}|^{\frac{1}{2}}} \exp\!\left(-\frac{\boldsymbol{x}^T \mathbf{M}^{-1} \boldsymbol{x}}{4t}\right).
\end{align*} Analogously, the solution of original PDE in eq.~(\ref{eq:PDE}) satisfies $u(\boldsymbol{x},t) \approx (u_0 * H_{\mathbf{D}})(\boldsymbol{x},t)$. In this parameter regime, the diffusion and covariance matrices $\mathbf{D}$ and $\mathbf{C}$ are related via an ordinary differential equation, \begin{align}\label{eq:covariance_and_diffusion}
    \frac{d\mathbf{C}}{dt} = 2\mathbf{D},
    \quad \text{where} \quad
    \mathbf{C}_{ij}(t)
    :=
    \text{cov}(x_i, x_j)(t).
\end{align} To take a slightly different perspective, this means that each component $D_{ij}$ of the diffusion matrix can be related to an analogous mean-squared displacement via \begin{align*}
    D_{ij} = \frac{1}{2} \frac{d}{dt} \, \mathbb{E}[(x_i - \mu_i)(x_j - \mu_j)],
\end{align*} implying that each length scale $\ell^2 \sim D_{ij} t_c$ is physically meaningful. In particular, one has $\mathbb{E}[|x_j - \mu_j|]^2 = (4/\pi)D_{jj}t$ for the marginal distribution of $x_j$ with mean $\mu_j$.

As a brief aside, we note that for direct estimates $\hat{D}_{ij}$ from empirical data, where the covariance structure of the dynamics may not be as simple as in eq.~(\ref{eq:covariance_and_diffusion}), one can use $\mu(\boldsymbol{x}; \mathbf{X}_t)$ instead of $u(\boldsymbol{x},t)$ within the corresponding expected value operators to obtain an effective formula: \begin{align*}
    \hat{\mathbf{D}}_t = \frac{1}{2t}\hat{\mathbf{C}}_t,
    \quad \text{with} \quad
    \hat{\mathbf{C}}_t := \frac{1}{N_t-1} \sum_{i=1}^{N_t} \big(\mathbf{x}^i_t - \hat{\boldsymbol{\mu}}_t\big) \otimes \big(\mathbf{x}^i_t - \hat{\boldsymbol{\mu}}_t\big),
\end{align*} where $\hat{\mathbf{C}_t} \approx \text{cov}(\mathbf{x}_t, \mathbf{x}_t)$ is an estimator of $\mathbf{C}(t)$, $\hat{\boldsymbol{\mu}}_t$ is a sample mean, and $\otimes$ is the dyadic outer product. With this in mind, we define the empirical estimates \begin{align*}
    \hat{D}_{\rm{eff}} := \text{arg}\!\min_{\!\!\!D} \sum_{n} \Big|\left\langle\Delta\mathbf{x}^i_{t_n}\right\rangle - \sqrt{\pi Dt_n}\Big|^2\!,
    \ \
    \hat{D}_{jj} := \text{arg}\!\min_{\!\!\!D} \sum_{n} \bigg|\left\langle\Delta{x}^i_{j,t_n}\right\rangle - \sqrt{\tfrac{4}{\pi} Dt_n} \bigg|^2,
\end{align*} where $\Delta\mathbf{x}^i_t := \|\mathbf{x}^i_t - \bar{\mathbf{x}}^i_0\|_2 - \|\mathbf{x}^i_0 - \bar{\mathbf{x}}^i_0\|_2$. We report uncertainties $\hat{D}_{jj} \pm \delta\hat{D}_{jj}$ in these estimates by propagating the standard error of the sample mean $\hat{\mu}_j$ throughout these computations within a $2\hat{\sigma}$ confidence interval, $\hat{\mu}_j \pm 2\hat{\sigma}(\hat{\mu}_j)$. To compute the standard errors $\hat{\sigma}(\hat{\mu}_j)$, we use a bootstrapping method with 1000 samples; see Figure~\ref{fig:caterpillars_violin_r} and Figure~\ref{fig:cats_violin_xy} in the appendix for an illustration.

We now consider a second case in which the diffusion matrix reduces to $\mathbf{D} = D\mathbf{I}$ for a positive scalar $D > 0$, which suggests a natural change of coordinates given by $\mathbf{A} = \ell\mathbf{I}$ for a diffusive length scale $\ell^2 = D t_c$. The nondimensionalized PDE in eq.~(\ref{eq:diffusion_focused_pde}) then takes the form \begin{align}\label{eq:simplified_diffusion_focused_pde}
    U_{\tau} &= \bar{\nabla} \cdot U \! \left({\Pi_V}\bar{\nabla} V + \, {\Pi_K}\bar{\nabla} K \! \star \! U\right) + \bar{\Delta}U,
\end{align} where, in this case, $\Pi_V = V_c/D$ and $\Pi_K = t_c K_c U_c$ are dimensionless scalar parameters. Suppose that the external potential strength $\Pi_V$ is non-negligible with a simultaneously small interaction term $\Pi_K \approx 0$ (i.e., $\Pi_K/\Pi_V \ll 1$), so that first-order approximations to the non-dimensionalized SDE and PDE are \begin{align*}
    d\boldsymbol{\xi}_{\tau}^i \, \approx \, -\Pi_V \bar{\nabla} V\!\big(\boldsymbol{\xi}_{\tau}^i\big) d\tau
    + \sqrt{2} \, d\mathbf{B}_{\tau}^i,
    \quad \text{and} \quad
    U_{\tau} \approx \Pi_V \bar{\nabla} \cdot \! \left(U \bar{\nabla} V\right) + \bar{\Delta}U.
\end{align*} Results from the theory of Langevin equations allow one to characterize the stationary Boltzmann distribution $U^{\star}$ that the solution $U$ converges to in the long-time limit: \begin{align*}
    U^{\star}(\boldsymbol{\xi}) \, := \, \lim_{\tau \rightarrow \infty} U(\boldsymbol{\xi},\tau) \, = \, \exp\!\big(\!-\!\Pi_V V(\boldsymbol{\xi})\big).
\end{align*} Analogously, in the original state variable $u(\boldsymbol{x},t)$, one has $u^{\star}(\boldsymbol{x}) = \exp(-V_c/D)$. In cases where the profile of the external potential $\mathcal{V}(\boldsymbol{x})$ reflects the underlying crop spacing by peaking near plant sites, this result intuitively implies that the population density tends to accumulate near plant resources in the long-time limit. %{\color{red}(References for this?)}

%\newpage

%A natural question follows: under what conditions is it reasonable to assume that the diffusion matrix $\mathbf{D}$ can be written as $D\mathbf{I}$ for a single constant $D$? In our case, this assumption is tantamount to stating that in the absence of environmental pressures and caterpillar-to-caterpillar interactions (i.e., $\nabla\mathcal{V} = \nabla\mathcal{K} = 0$), an individual will not take a biased random walk -- that is, an individual will not \textit{inherently} prefer dispersing in one direction (e.g., $x$) faster than another (e.g., $y$). Note that with a homogeneous diffusion constant $D$, one can still accurately account for the effects of an inhomogeneous environment (e.g., mixed terrain or non-uniformly placed plant resources) by subsuming these mechanisms under either the effective environmental potential $\mathcal{V} = \mathcal{V}(x,y)$.

\newpage

\subsubsection{Weak Formulation}
We now consider multiplying each side of the PDE in eq.~(\ref{eq:nondimmed_pde}) by a collection $\{\psi_k\}_{k=1}^{\kappa}$ of translations of a symmetric and compactly-supported test function, \begin{align*}
    \psi_k(\boldsymbol{x},t) := \psi(\boldsymbol{x}_k - \boldsymbol{x}, t_k - t) \in C_c^p(\Omega_T),
\end{align*} where $p \geq 2$ and $\Omega_T := \Omega \times [0, T]$. In turn, we integrate over the space-time domain $\Omega_T$ to obtain \begin{align*}
    \left\langle \psi, \, u_t \right\rangle &= \left\langle \psi, \, \nabla \! \cdot \! \Big( u \big(\nabla \mathcal{V} + \nabla\mathcal{K} \! * \! u\big) + \mathbf{D} \nabla u\Big) \! \right\rangle,
\end{align*} where $\langle\cdot,\cdot\rangle$ denotes the $L^2$ inner product.\footnote{Note that for vector-valued functions, we integrate the dot-product, i.e., $\langle \vec{\boldsymbol{v}}, \vec{\boldsymbol{w}}\rangle := \sum_i \langle v_i, w_i\rangle$.} An application of Green's identities (i.e., integration by parts), exploiting the compact support of $\psi$ and the symmetry of $\mathbf{D}$, then yields the \textit{weak formulation} of eq.~(\ref{eq:nondimmed_pde}): \begin{align}\label{eq:weak_form}
    \left\langle \psi_t, \, u \right\rangle
    &=
    \left\langle \nabla\psi, \, u\big(\nabla\mathcal{V} +\nabla \mathcal{K} \! * \! u\big) \right\rangle
    +
    \left\langle \nabla\!\cdot\!\big(\mathbf{D}\nabla \psi\big), \, u \right\rangle.
\end{align} This weak formulation will serve as a foundation for our model discovery methodology, which is formally a Petrov-Galerkin approach.

Normally, the weak formulation in eq.~(\ref{eq:weak_form}) is viewed as a variational constraint on the solution $u$ of the PDE in eq.~(\ref{eq:PDE}). Here, however, we take an inverse perspective, viewing eq.~(\ref{eq:weak_form}) as a constraint on the $\mathcal{K}, \, \mathcal{V}$, and $\mathbf{D}$ terms, evaluated on the data $u$. That is, if $u(\boldsymbol{x},t)$ satisfies eq.~(\ref{eq:PDE}) and in turn eq.~(\ref{eq:weak_form}), then we have \begin{align}\label{eq:petrov_galerkin}
    b(\psi_k) = \mathcal{G}_V(\mathcal{V},\psi_k) + \mathcal{G}_K(\mathcal{K}, \psi_k) + \mathcal{G}_D(\mathbf{D},\psi_k),
\end{align} for each test function $\psi_k \in \{\psi_k\}_{k=1}^{\kappa}$, where here the $\mathcal{G}_i$ are bilinear forms defined by \begin{align*}
    \begin{cases}
        \mathcal{G}_V(\mathcal{V}, \psi; u) := \left\langle \nabla\psi, \, u \nabla\mathcal{V} \right\rangle,
        \\
        \mathcal{G}_K(\mathcal{K}, \psi; u) := \left\langle \nabla\psi, \, u\big(\nabla \mathcal{K} \! * \! u\big) \right\rangle,
        \\
        \mathcal{G}_D(\mathbf{D}, \psi; u) := \left\langle \nabla\!\cdot\!\big(\mathbf{D}\nabla \psi\big), \, u \right\rangle,
    \end{cases}
\end{align*} and $b$ is a linear functional defined by $b(\psi; u) := \langle \psi_t, \, u \rangle$. Correspondingly, we propose the finite basis expansions \begin{align*}
    \mathcal{V}_{\mathbf{w}}(x,y) := \sum_{n=1}^{J_V}\sum_{m=1}^{J_V} w^{(V)}_{nm} \, \mathcal{V}_{nm}(x,y),
    \quad \text{and} \quad
    \mathcal{K}_{\mathbf{w}}\big(\boldsymbol{x};\boldsymbol{x}'\big) := \sum_{j=1}^{J_K} w_j^{(K)} \mathcal{K}_j\big(\boldsymbol{x};\boldsymbol{x}'\big),
\end{align*} which can, in turn, be substituted into the linear expansion in eq.~(\ref{eq:petrov_galerkin}) to yield \begin{align*}
    b(\psi_k)
    =
    \!\! \Bigg[\sum_{n,m} w^{(V)}_{nm} \, \mathcal{G}_V(\mathcal{V}_{nm}, \psi_k) \Bigg]
    \! + \! \Bigg[\sum_{j} w^{(K)}_j \mathcal{G}_K(\mathcal{K}_j, \psi_k) \Bigg]
    \! + \! \Bigg[\sum_{i,j} w^{(D)}_{ij} \, \mathcal{G}_D\big(\boldsymbol{\delta}_{ij}, \psi_k\big) \Bigg],
\end{align*} where $w^{(D)}_{ij} := D_{ij}$. Note that we use `$\mathbf{w}$' to denote the $(J_V\!+\!J_K\!+\!3)$-element column vector obtained by `stacking' each set of parameters.

%\vfill\null
%\columnbreak
\newpage

The variational problem can be recast as a regression problem by, e.g., using the $\mathbf{w}$-parameterization described above to identify model terms $\mathcal{V}_{\mathbf{w}^{\star}}$, $\mathcal{K}_{\mathbf{w}^{\star}}$, and $\mathbf{D}^{\star}$ that minimize the weak-form equation residual, solving \begin{align*}
    \mathbf{w}^{\star}
    \, = \,
    \text{arg}\!\min_{\!\!\!\!\!\mathbf{w}}
    \, \sum_{k=1}^{\kappa} \big| r(\mathbf{w};\psi_k) \big|^2\!\!,
\end{align*} which is implicitly evaluated on the density estimate $\hat{u}(\boldsymbol{x},t)$, where \begin{align*}
    r(\mathbf{w};\psi_k) := b(\psi_k) - (\mathcal{G}_V + \mathcal{G}_K + \mathcal{G}_D)(\mathbf{w},\psi_k).
\end{align*} Since in our case we expect the environmental potential to reflect the structure of the regularly-spaced crops with negligible boundary effects, we express $\mathcal{V}(x,y) = \mathcal{V}_{\mathbf{w}}(x,y)$ in a cosine series basis, setting \begin{align*}
    \mathcal{V}_{nm}(x,y) := \cos\!\left(\frac{2\pi n x}{L}\right) \cos\!\left(\frac{2\pi m y}{W}\right).
\end{align*} where we use equivalent lengths and widths $L, W = 175$. Similarly, we search for a radially-symmetric\footnote{The gradient of a radially-symmetric function reduces to $\nabla\mathcal{K}(\rho) = \left(\boldsymbol{x}/\rho\right) \mathcal{K}'(\rho)$.} interaction potential $\mathcal{K}(\rho) = \mathcal{K}_{\mathbf{w}}(\rho)$ by setting \begin{align*}
\mathcal{K}_n(\rho) := j_{n-1}\!\left( \frac{\rho}{\rho_0}\right),
\end{align*} where $j_n$ denotes the degree-$n$ spherical Bessel function of the first kind and $\rho_0$ is a scaling factor we provisionally set to $\rho_0 = 6$ throughout. Note that the potentials can be offset by arbitrary constants $\mathcal{V}_0$ and $\mathcal{K}_0$ to yield the same results under the gradients $\nabla\mathcal{V}$ and $\nabla\mathcal{K}$; for simplicity, we choose gauge constants $\mathcal{V}_0$ and $\mathcal{K}_0$ such that the resulting potentials have zero mean.

\subsection{Numerical Methods}\label{sec:numerical_methods}
\begin{figure*}
    \centering
    \includegraphics[width=\linewidth]{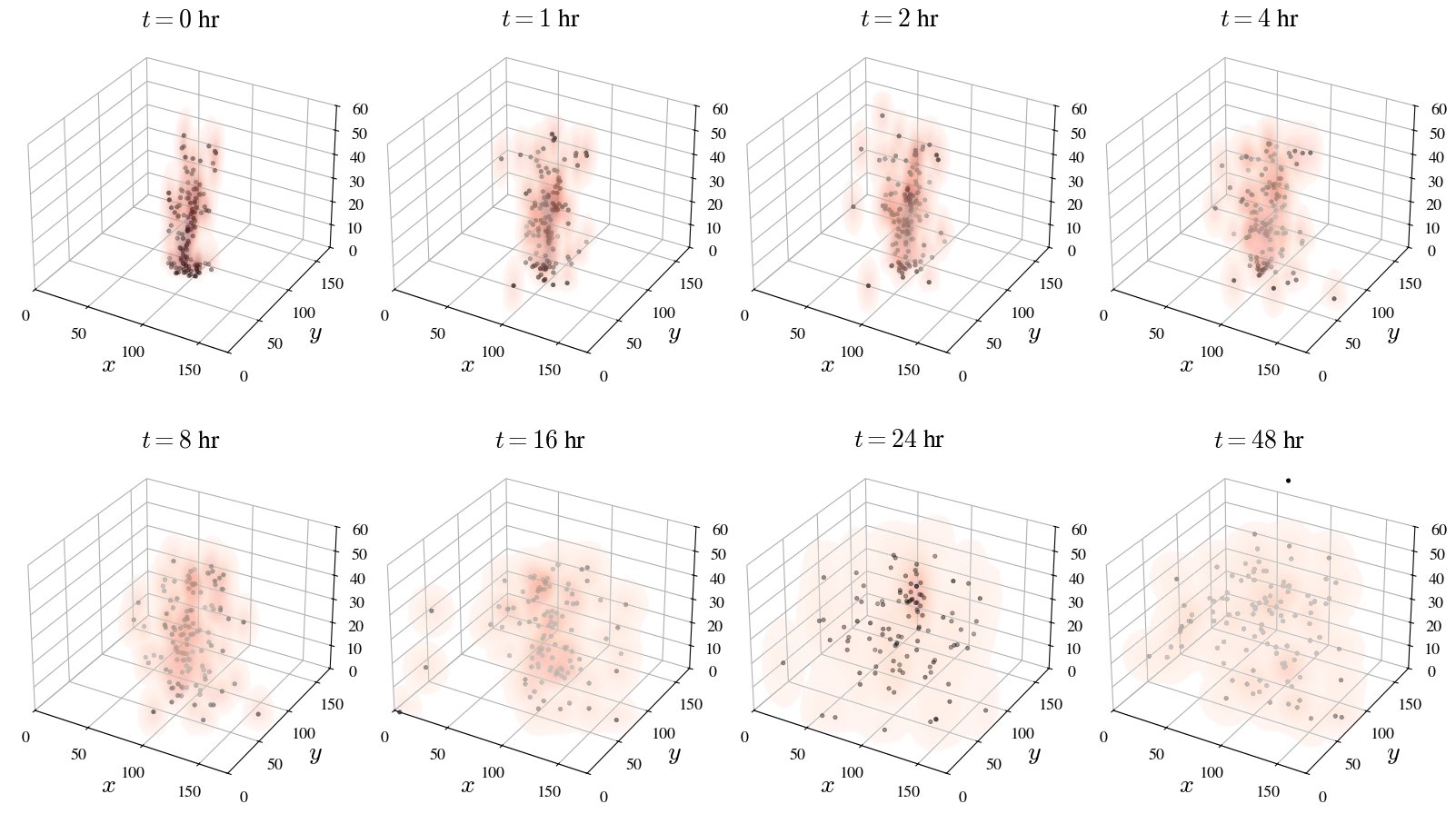}
    \caption{Visualizing the combined armyworm positions $\mathbf{X}_t$ from each experiment (black dots) and the resulting KDEs $\hat{u}_h(\boldsymbol{x},t)$ (red volume), plotted at eight times $t \in \{t_0, \dots, t_{\textsc{f}}\}$. Note that we neglect the $z$-component in our models.}
    \label{fig:caterpillars_3d_density}
\end{figure*}

To formulate a coarse-grained model with the finite number of samples $\mathbf{X}_t$ given in eq.~(\ref{eq:empirical_distribution}), where $N_t < \infty$, we estimate a density $\hat{u}_h(\boldsymbol{x},t)$ by smoothing the empirical data using \begin{align}\label{eq:gaussian}
    \hat{u}_h(\boldsymbol{x},t) := \frac{1}{N_t} \int\!\!\!\!\int_{\Omega} G_h\big(\boldsymbol{x} - \boldsymbol{x}'; t\big) \, \mu\big(\boldsymbol{x}'; \mathbf{X}'_t\big) \, dx' dy'.
\end{align} Here, $G_h$ is a Gaussian kernel of bandwidth $h$, defined by \begin{align*}
    G_h(\boldsymbol{x};t) := \frac{1}{2\pi h|\hat{\mathbf{C}}_t|^{\frac{1}{2}}} \exp\!\left(-\frac{\boldsymbol{x}^T\hat{\mathbf{C}}_t^{-1}\boldsymbol{x}}{2h^2} \right)\!,
\end{align*} where $\hat{\mathbf{C}}_t$ represents the sample estimate of the covariance matrix of the data $\mathbf{X}_t$, as before, and the (time-dependent) bandwidth $h = 1/\sqrt[6]{N_t}$ is chosen according to \textit{Silverman's rule of thumb} \cite{silverman_density_2018}. The resulting kernel density estimate (KDE) of the empirical distribution $\mu(\boldsymbol{x}; \boldsymbol{X}_t)$ is shown in Figure~\ref{fig:caterpillars_3d_density} (red volume). Note that the level of smoothing may impact the model discovery results; see the sensitivity analysis detailed in \S\ref{sensitivity_analysis} below.

\newpage

% \begin{figure*}
%     \includegraphics[width=\linewidth]{pictures/cats_violin_plots.png}
%     \caption{Illustrating the average radial displacement $\langle\rho\rangle$ at each snapshot $t_n$ (see inset panels for the full distributions). For the raw data, $2.5\%$ to $97.5\%$ confidence intervals were computed using a bootstrapping method with $1000$ samples from each distribution. For the empirical and PDE models, we plot $\rho(t) = \sqrt{\pi (D_{\rm{eff}} \pm 2\hat{\sigma}) t}$, where $D_{\rm{eff}}$ is the corresponding parameter estimate with standard deviation $\hat{\sigma}$.}
%     \label{fig:caterpillars_violin_r}
% \end{figure*}

\newpage

\subsubsection{Weak SINDy}
A popular paradigm for data-driven PDE discovery is that of dictionary learning, which broadly attempts to equate an \textit{evolution operator} (e.g., $\partial_t u$) with a closed-form expression consisting of functions taken from a \textit{library} $\boldsymbol{\Theta}(\mathcal{U})$ of candidate terms, \begin{align*} 
    \boldsymbol{\Theta}(\mathcal{U})
    =
    \Big\{ \mathcal{D}^j\!f_j(u_m) \, : \, u_m \in \mathcal{U} \ \text{and} \ j=1,\dots,J \Big\}.
\end{align*} Here, $\mathcal{U}$ represents a set of empirical observations of a state variable $u_m := u(\boldsymbol{x}_m,t_m)$; in our case, we use the set of density estimates obtained over a discretized spatiotemporal grid $\Omega^{\Delta}_T$, with \begin{align*}
    \mathcal{U} = \Big\{\hat{u}(\boldsymbol{x}_m, t_m) \, : \, (\boldsymbol{x}_m,t_m) \in \Omega_T^{\Delta}\Big\}.
\end{align*} In the above formulation, each $\mathcal{D}^j$ denotes a distinct differential operator while each $f_j$ represents a distinct scalar-valued functions of the state variable $u$.

In the Sparse Identification of Nonlinear Dynamics (SINDy) algorithm \cite{BruntonProctorKutz2016ProcNatlAcadSci}, the model discovery problem is structured as a regression problem posed over a sparse vector of coefficients which weight candidate basis functions in the library, \begin{align*}
    \mathbf{w} \, = \, [w_1, \, \dots, \, w_{J}]^T,
    \quad \text{with} \ \quad
    \lvert\!\lvert \mathbf{w} \rvert\!\rvert_{0} = J' \leq J.
\end{align*} Here, $\lvert\!\lvert \, \cdot \, \rvert\!\rvert_{\text{0}}$ denotes the $\ell_0$ \lq\lq{norm},'' which returns the number of non-zero elements of a vector. Although SINDy originally addressed ordinary differential equations, subsequent work by \cite{RudyBruntonProctorEtAl2017SciAdv,Schaeffer2017ProcRSocMathPhysEngSci} has extended it to the context of PDEs, where the central problem is to find sparse $\mathbf{w}$ such that: \begin{align} \label{eq:sindy}
    \partial_t u_m \approx
    \sum_{j=1}^{J} w_j \, \mathcal{D}^j \! f_j (u_m),
\end{align} for each empirical observation $u_m \in \mathcal{U}$. Numerically, we restructure eq.~(\ref{eq:sindy}) as an equivalent linear system \begin{align*}
    \partial_t \mathbf{u} = \boldsymbol{\Theta}(\mathbf{u}) \mathbf{w},
\end{align*} by vectorizing the data via $\mathbf{u} := \texttt{vec}\big\{ \hat{u}_m \big\} \in \mathbb{R}^{M}$. In turn, one uses a matrix-valued library $\boldsymbol{\Theta}(\mathbf{u}) \in \mathbb{R}^{M \times J}$ whose columns $\vec{\Theta}_j$ are given by \begin{align*}
    \mathcal{D}^j\!f_j(\mathbf{u}) := \texttt{vec}\big\{ \mathcal{D}^j\!f_j(u_m) \big\} \in \mathbb{R}^M.
\end{align*} The terms in eq.~(\ref{eq:sindy}) then take the form of data matrices, which can schematically represented in the form \begin{align*}
    \begin{bmatrix}
        \vline
        \\
        \partial_t\mathbf{u}
        \\
        \vline
    \end{bmatrix}
    =
    \begin{bmatrix}
        \vline & & \vline \\
        \mathcal{D}^1\!f_1\!\left(\mathbf{u}\right) & \cdots & \mathcal{D}^J\!f_J\!\left(\mathbf{u}\right) \\
        \vline & & \vline
    \end{bmatrix}
    \begin{bmatrix}
        \vline 
        \\
        \mathbf{w}
        \\
        \vline
    \end{bmatrix}.
\end{align*} Note that when applying operators to the data $\mathbf{u}$, such as $\partial_t\mathbf{u}$ and $f_j(\mathbf{u})$, we perform element-wise computations.

%\vfill\null
%\columnbreak

Weak SINDy (WSINDy) \cite{ MessengerBortz2021JComputPhys,MessengerBortz2021MultiscaleModelSimul} generalizes the SINDy algorithm by converting it to an integral formulation which alleviates the need to approximate derivatives on potentially ill-behaved data $\mathbf{u}$. In particular, WSINDy extends the original work by converting sparse parameter-estimation problems of the form of eq.~(\ref{eq:sindy}) into a weak, integral-based formulation: \begin{align*}
    \left\langle \partial_t\psi_k, \, u \right\rangle
    \approx
    \sum_{j=1}^{J} w_j \left\langle \mathcal{D}^j\psi_k, \, f_j (u)\right\rangle.
\end{align*} A key benefit of the weak formulation is that derivative approximations of the data are avoided by transferring the differential operators $\mathcal{D}^j$ from nonlinear observations of the data $f_j(u)$ to the test functions $\psi_k$ by repeated integration by parts, exploiting the compact support of the test functions.\footnote{The sign convention in the argument of each test function eliminates any resulting alternating factors of $(-1)^{\alpha_j}$, where $\alpha_j$ is the order of $\mathcal{D}^j$.} This integral formulation has been shown to exhibit substantially higher-fidelity results than SINDy in the presence of noisy data; see, e.g., Table 6 in \cite{MessengerBortz2021JComputPhys}.

One can discretize the variational problem in eq.~(\ref{eq:petrov_galerkin}) in the form of an equivalent linear system $\mathbf{b} = \mathbf{Gw}$, where the response vector $\mathbf{b} \in \mathbb{R}^{\kappa}$ and weak-form library $\mathbf{G} \in \mathbb{R}^{\kappa\times\!{J}}$, with $J:=J_V\!+\!J_K\!+\!3$, are defined by \begin{align}\label{eq:wsindy}
    \begin{cases}
        \ \mathbf{b}[k] := \left(\psi_t \star \hat{u}_h\right)(\boldsymbol{x}_k, t_k),
        \\
        \mathbf{G}[j,k] := \big(\mathcal{D}^j\psi \star f_j(\hat{u}_h)\big)(\boldsymbol{x}_k, t_k),
    \end{cases}
\end{align} for the appropriate differential operator $\mathcal{D}^j$ and function $f_j$. Here, $\star$ denotes the discrete convolution operator, computed using the trapezoidal rule on the discrete grid $\Omega_T^{\Delta}$.\footnote{As outlined in detail in \cite{MessengerBortz2021JComputPhys}, we note that the discrete convolution in eq.~(\ref{eq:wsindy}) can be computed using the FFT in $\mathcal{O}(\kappa \log\kappa)$ time.} The `optimal' sparse vector of coefficients $\mathbf{w}^{\star}$ is found by minimizing a regularized loss function $\mathcal{L}$, leading to an optimization problem given by
\begin{align}\label{eq:wsindyLoss}
    \mathbf{w}^{\star} = \text{arg}\!\min_{\!\!\!\!\!\mathbf{w}} \ \mathcal{L}\left(\mathbf{w}; \mathbf{b}, \mathbf{G}\right),
\end{align} where $\mathcal{L}$ has the form (see \S\ref{sec:additional_implementation_details} in the appendix): \begin{align*}
    \mathcal{L}\left(\mathbf{x}; \mathbf{b},\mathbf{A}\right) := \lvert\!\lvert \mathbf{b} - \mathbf{A}\mathbf{x} \rvert\!\rvert_2^2 + \eta\lvert\!\lvert \mathbf{x} \rvert\!\rvert_0.
\end{align*} The regularization term  $\eta\lvert\!\lvert \mathbf{w} \rvert\!\rvert_{0}$ promotes the selection of a sparse model by penalizing models with a large number of terms. In practice, this is achieved by using iterative thresholding optimization schemes which progressively restrict the number of terms available to the model; see \cite{BruntonProctorKutz2016ProcNatlAcadSci, MessengerBortz2021JComputPhys}.

% \begin{figure*}
%     \centering
%     %\includegraphics[width=0.965\linewidth]{pictures/cats_simulated_trajectories_SDE.png}
%     \includegraphics[width=\linewidth]{pictures/cats_simulated_trajectories_SDE_2.png}
%     \caption{Illustrating the dynamics that are possible under an SDE consistent with eq.~(\ref{eq:simplified_diffusion_focused_pde}) in various parameter regimes. Here, we fix the diffusion strength $\Pi_D = 1$ and incrementally increase the potential strengths $\Pi_V$ and $\Pi_K$; see \S\ref{sec:additional_implementation_details}.}
%     \label{fig:caterpillar_trajectories}
% \end{figure*}

We follow \cite{MessengerBortz2021JComputPhys} in using localized test functions $\psi_k$ with compact support given by \begin{align*}%\label{eq:testFcnSupport}
    \text{supp}(\psi) = \big[-m_{x}\Delta{x}, \ m_{x}\Delta{x}\big]
    \times \big[-m_{y}\Delta{y}, \ m_{y}\Delta{y}\big]
    \times \big[-m_{t}\Delta{t}, \ m_{t}\Delta{t}\big],
\end{align*} where the tuple $\boldsymbol{m} = (m_{x}, m_{y}, m_t)$ then becomes a tunable hyperparameter; see \S\ref{sec:additional_implementation_details} in the appendix for more details on our choice of hyperparameters. We note that as the support radii $m_i \rightarrow 0$, the WSINDy algorithm collapses to the SINDy algorithm; in particular, the test functions $\psi_k$ converge to Dirac delta functions $\delta(\boldsymbol{x}_k,t_k)$ while $\mathcal{D}^i \psi(\Omega^{\Delta}_T)$ converge to kernels resembling difference operators.

\subsubsection{Discretization}
In our numerical implementation, we discretize the data by subsampling the KDE given in eq.~(\ref{eq:gaussian}), $\hat{u}_h(x,y,t)$, over a discrete and equi-spaced grid, \begin{align*}
    \Omega^{\Delta}_T := \mathbf{x} \otimes \mathbf{y} \otimes \mathbf{t},
\end{align*} of size $80\!\times\!80\!\times\!98$, producing a tensor $\mathbf{u}[i,j,n]$ of the same shape.\footnote{We find that an $80 \times 80$ spatial resolution is sufficient to avoid aliasing artifacts from the sinusoidal $\mathcal{V}_{nm}$ terms up to degree $J_V \leq 9$, which corresponds to the number of crops along the $x$-axis. Note that evaluation over $\mathbf{t}$ corresponds to linear interpolation over the snapshots $t_n$; see Figure~\ref{fig:temporal_interpolation} in the appendix.} Similarly, we discretize the external potential into a matrix $\mathbf{V}[i,j]$ by subsampling $\mathcal{V}(x,y)$ over $\Omega^{\Delta} = \mathbf{x} \otimes \mathbf{y}$ (see Figure~\ref{fig:learned_potentials}, left panel), where we set $J_V = 9$. Because the interaction potential $\mathcal{K}(\boldsymbol{x}; \boldsymbol{x}')$ represents a \textit{local} convolution kernel, we represent it as a matrix $\mathbf{K}[i-i',j-j']$ computed over a symmetric grid $(\mathbf{x}-\mathbf{x}') \otimes (\mathbf{y}-\mathbf{y}')$ of radius $30\Delta{x}$ (Figure~\ref{fig:learned_potentials}, right panel), modeling interactions over length scales of $\leq 65$ cm. In all cases, we set $J_K = 5$.

We discretize the variational problem as in eq.~(\ref{eq:wsindy}) above, using a set of separable test functions of the form \begin{align*}
    \psi(\boldsymbol{x},t) = \phi_x(x) \phi_y(y) \phi_t(t),
\end{align*} where each $\phi_i$ is given by \begin{align*}
    \phi_i(x) := \left[1-(x/m_{i}\Delta_i)^2 \right]^{p_i}\!\!\!,
    \quad \text{for} \quad
    x \in [-m_i\Delta_i, \, m_i\Delta_i],
\end{align*} and the test function degrees $p_i$ are defined for a highest degree $\bar{\alpha}_i$ and support tolerance $\tau_0 = 1\texttt{e}-10$ via \begin{align*}
    p_i = \max\left\{ \left\lceil
    \frac{\ln(\tau_0)}{\ln\!\big((2\ell_i-1)/\ell_i^2\big)}
    \right\rceil\!, \ \bar{\alpha}_i + 1 \right\}.
\end{align*} For additional information about our hyperparameter selection and numerical implementation, we refer the reader to \S\ref{sec:additional_implementation_details} in the appendix.

\newpage

\begin{figure*}
    \includegraphics[width=\linewidth]{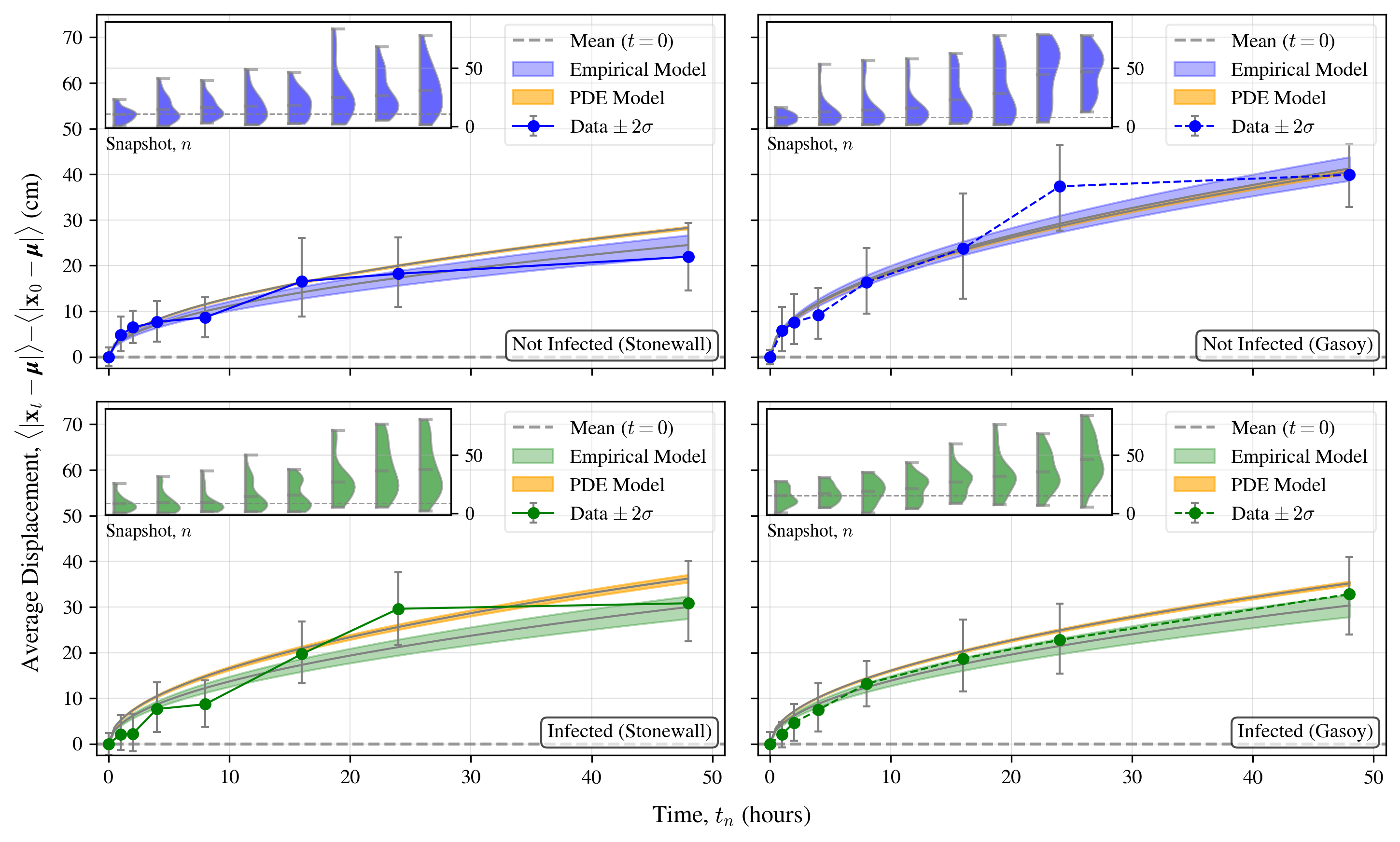}
    \caption{Illustrating the average radial displacement $\langle\rho\rangle$ at each snapshot $t_n$ (see inset panels for the full distributions). For the raw data, $2.5\%$ to $97.5\%$ confidence intervals were computed using a bootstrapping method with $1000$ samples from each distribution. For the empirical and PDE models, we plot $\rho(t) = \sqrt{\pi (D_{\rm{eff}} \pm 2\hat{\sigma}) t}$, where $D_{\rm{eff}}$ is the corresponding parameter estimate with standard deviation $\hat{\sigma}$.}
    \label{fig:caterpillars_violin_r}
\end{figure*}

\section{Results}
\label{Sec:results}
To illustrate a trade-off between model complexity and goodness of fit, we obtain results using a hierarchy of PDE models, respectively referenced in Tables~\ref{table:full_model}, \ref{table:purely_diffusive_model}, and \ref{table:Deff_model}: \\
(1) a complete McKean-Vlasov model of the form \begin{align*}
    u_t = \nabla \cdot \Big(u \big(\nabla \mathcal{V} + \nabla\mathcal{K} \! * \! u\big) + \mathbf{D} \nabla u\Big),
\end{align*} (2) a partially-idealized and purely-diffusive, but \textit{anisotropic}, model of the form \begin{align*}
    u_t = \nabla\cdot(\mathbf{D}\nabla u),
\end{align*} and, lastly, (3) a highly-idealized and isotropic \textit{effective} diffusion model of the form \begin{align*}
    u_t = D_{\rm{eff}} \Delta{u}.
\end{align*} To help gauge the quality of the results, we report the coefficient of determination $R^2$ corresponding to each WSINDy regression, which is defined by \begin{align*}%\label{eq:rsquare}
    R^2 = 1 - \frac{\| \, \mathbf{r} \, \|_2^2}{\big|\!\big| \, \mathbf{b} - \overline{\mathbf{b}} \, \big|\!\big|_2^2},
    \quad \text{with} \quad
    \overline{\mathbf{b}} := \left(\frac{1}{\kappa}\sum_{k=1}^{\kappa} b_k\right) \vec{\mathbf{1}},
\end{align*} where $\mathbf{r} := \mathbf{b} - \mathbf{G}\mathbf{w}^{\star}$ is the query-pointwise weak-form equation residual. This metric, which equals the proportion of the variance of $\mathbf{b}$ that is explained by the discovered sparse model $\mathbf{G}\mathbf{w}^{\star}$, satisfies $R^2 \leq 1$, with the values closer to 1 indicating a better performing model. In turn, we assess the balance between goodness of fit and model complexity by reporting the comparative \textit{Akaike information criterion} (AIC) for each regression,\footnote{Note that the WSINDy loss function in eq.~(\ref{eq:wsindyLoss}) is equivalent to ${\rm{AIC}}$ under a logarithmic rescaling of $\mathbf{r}$ and choice of $\eta=2$.} defined by \begin{align*}
    {\rm{AIC}}(\mathbf{u}, \mathbf{w}) := 2\lvert\!\lvert \mathbf{w} \rvert\!\rvert_{0} - 2\ell(\mathbf{w}; \mathbf{u}),
\end{align*} where $\ell(\mathbf{w}; \mathbf{u})$ denotes the maximized log-likelihood of the model with weights $\mathbf{w}$, given data $\mathbf{u}$. Note that when reporting $\Delta{\rm{AIC}}(\cdot, \mathbf{w}_1, \mathbf{w}_2) := {\rm{AIC}}(\cdot, \mathbf{w}_1) - {\rm{AIC}}(\cdot, \mathbf{w}_2)$, we estimate log-likelihood values ($\ell$-values) using the ordinary least squares (OLS) estimator, neglecting the arbitrary normalization constant: \begin{align*}
    \ell(\mathbf{w}; \mathbf{u})
    \approx
    - \frac{N}{2} \ln\!\Big(\big|\!\big|\mathbf{r}(\mathbf{u})\big|\!\big|^2_2\Big),
    \ \ \ \text{where} \ \ \
    N = N_{t_0} + \cdots + N_{t_{\textsc{f}}}.
\end{align*} The standard error estimates $\hat{\sigma}(w_j)$ for the learned model weights, reported in Tables~\ref{table:purely_diffusive_model} and \ref{table:Deff_model}, are computed via \begin{align*}
    \hat{\sigma}(w_{j})^2
    = \, \hat{\mathbf{S}}_{jj},
\end{align*} where $\hat{\mathbf{S}} \approx \text{var}(\mathbf{w}-\mathbf{w}^{\star})$, see eq.~(\ref{eq:standard_errors}), is the `robust' estimate of the parameter covariance matrix derived in \S\ref{sec:residuals_and_parameter_error}.

Overall, the WSINDy (and OLS) models are found to be in good qualitative agreement with the empirical results, both in terms of dynamical consistency (see Figure~\ref{fig:caterpillars_violin_r}) and in relation to the empirical diffusion coefficients $\hat{D}_{ij}$. Unsurprisingly, the ensemble models consistently obtain a better fit. In the remainder of this section, we detail these relationships as well as comment on relevant differences between the various experimental control groups. See also the supplemental results in the appendix (Figures~\ref{fig:HISTOGRAMS}-\ref{fig:empirical_Dx_vs_Dy}).

%\newpage

\subsection{Raw Data}\label{sec:raw_data}
In Figure~\ref{fig:caterpillars_violin_r}, we plot the average radial displacement $\langle\rho\rangle$ of the individual displacements $\{\rho^i_t\}_{i=1}^{N_t}$ evolving over each temporal snapshot $t_n$, where $\rho^i_t := \|\mathbf{x}^i_t - \langle\mathbf{X}_0\rangle\|_2$. To give a sense of the variance in these measurements, we overlay the KDEs corresponding to each empirical distribution of $\{\rho^i_t\}$ values; see also Figure~\ref{fig:cats_violin_xy} and Figure~\ref{fig:cats_violin_z} in the appendix, which illustrate the $\{x^i_t\}$, $\{y^i_t\}$ and $\{z^i_t$\} distributions and averaged $\langle{x}\rangle$, $\langle{y}\rangle$, and $\langle{z}\rangle$ displacements, respectively. Most importantly, these plots illustrate that the movement dynamics are indeed dominantly diffusive, with displacements growing on the order $\mathcal{O}(\sqrt{D_{ij}t})$ in time. Although our experiment simulated realistic farm practices by featuring anisotropic crop-spacing along the $x$ and $y$ axes, the data do not clearly indicate that the diffusion constants $D_x$ and $D_y$ along these axes differ in a systematic way; see also Figure~\ref{fig:empirical_Dx_vs_Dy} in the appendix, which displays the superimposed $\langle{x}\rangle$ and $\langle{y}\rangle$ averages.\footnote{A potential exception to this result are the uninfected larvae on the Stonewall variety, which appear to disperse faster along the $y$-direction.} Moreover, the comparatively small $\langle z \rangle$-displacements (see Figure~\ref{fig:cats_violin_z}) indicate that, while the individuals do tend to ascend the plant a vertical distance of roughly $10 \pm 5$ cm over the course of the two-day experiment, diffusion rates along the vertical $z$-axis are substantially weaker than those along either the $x$ or $y$ axes. This empirical result further motivates our choice to use only $\mathbf{x}^i_t = (x^i_t, y^i_t)$ observations in our data-driven models.

\newpage

\begin{landscape}
\begin{table}[htb]
    \centering
    \begin{tabular}{||c c c c c c c||}
     \hline
     & & & & & & \\[-2.0 ex]
     \textbf{Plant} & \textbf{Virus} & \ \ \ $\boldsymbol{V_c \pm 2\hat{\sigma}}$ & \ \ \ $\boldsymbol{K_c \pm 2\hat{\sigma}}$ & $\boldsymbol{[D_{x}, \, D_{xy}, \, D_{y}] \pm 2\hat{\sigma}}$ & $\boldsymbol{R^2}$ & $\boldsymbol{\Delta{\rm{AIC}}}$ \\ [0.5ex]
     \hline
     \hline
     & & & & & & \\[-1.5 ex]
     %\hline
     $\dagger$ & $\dagger$ & \ \ \ $\boldsymbol{1.8} \, | \, 3.6$ & \ \ \ \textsc{n/a} & \ \ \ $\boldsymbol{[8.6, 0.0, 9.1]} \, | \, [7.4, 1.0, 8.4]$ & $\boldsymbol{0.67} \, | \, 0.70$ & -59.64 \\[0.5 ex]
     & & {\color{gray}$\boldsymbol{\pm 0.1} \, | \, 1.0$} & & {\color{gray}$\boldsymbol{\pm [0.2, 0.3, 0.2]} \, | \, [0.3, 0.3, 0.2]$} & & \\[0.75 ex]
     \hline
     & & & & & & \\[-1.5 ex]
     Stonewall & $\dagger$ & \ \ \ $\boldsymbol{2.0} \, | \, 3.1$ & \ \ \ \textsc{n/a} & \ \ \ $\boldsymbol{[6.1,1.9,7.1]} \, | \, [6.1,2.0,7.1]$ & $\boldsymbol{0.59} \, | \, 0.60$ & -148.41 \\[0.5 ex]
     & & {\color{gray}$\boldsymbol{\pm 0.2} \, | \, 1.5$} & & {\color{gray}$\boldsymbol{\pm [0.3, 0.3, 0.3]} \, | \, [0.4, 0.3, 0.3]$} & & \\[0.5 ex]
     %\hline
     Gasoy & $\dagger$ & \ \ \ $\boldsymbol{1.7} \, | \, 2.8$ & \ \ \ \textsc{n/a} & \ \ \ $\boldsymbol{[7.0,{\text{-}}2.2,9.6]} \, | \, [6.9,{\text{-}}2.0,9.7]$ & $\boldsymbol{0.51} \, | \, 0.51$ & -135.90 \\[0.5 ex]
     & & {\color{gray}$\boldsymbol{\pm 0.0} \, | \, 1.1$} & & {\color{gray}$\boldsymbol{\pm [0.3, 0.5, 0.4]} \, | \, [0.3, 0.5, 0.4]$} & & \\[0.75 ex] 
     \hline
     & & & & & & \\[-1.5 ex]
     $\dagger$ & No & \ \ \ $\boldsymbol{1.5} \, | \, 2.0$ & \ \ \ \textsc{n/a} & \ \ \ $\boldsymbol{[5.5, 0.9, 7.8]} \, | \, [5.6,0.9,7.8]$ & $\boldsymbol{0.64} \, | \, 0.65$ & -136.71 \\[0.5 ex]
     & & {\color{gray}$\boldsymbol{\pm 0.0} \, | \, 0.9$} & & {\color{gray}$\boldsymbol{\pm [0.2, 0.2, 0.2]} \, | \, [0.2, 0.2, 0.2]$} & & \\[0.5 ex]
     %\hline
     $\dagger$ & Yes & \ \ \ $\boldsymbol{1.4} \, | \, 4.6$ & \ \ \ \textsc{n/a} & \ \ \ $\boldsymbol{[11.6, 0.0, 8.3]} \, | \, [11.9, 0.2, 8.7]$ & $\boldsymbol{0.58} \, | \, 0.59$ & -141.17 \\[0.5 ex]
     & & {\color{gray}$\boldsymbol{\pm 0.1} \, | \, 1.4$} & & {\color{gray}$\boldsymbol{\pm [0.6, 1.0, 0.5]} \, | \, [0.6, 0.6, 0.5]$} & & \\[0.75 ex]
     \hline
     & & & & & & \\[-1.5 ex]
     Stonewall & No &
     \ \ \ $\boldsymbol{0.9} \, | \, 2.2$ & \ \ \ $\boldsymbol{0.0}^* \, | \, 0.6^*$ & \ \ \ $\boldsymbol{[3.6, 2.5, 6.7]} \, | \, [3.8, 2.5, 6.7]$ & $\boldsymbol{0.36} \, | \, 0.36$ & -148.35 \\[0.5 ex]
     & & {\color{gray}$\boldsymbol{\pm 0.1} \, | \, 1.4$} & {\color{gray}$\boldsymbol{\pm 0.0^*} \, | \, 2.3^*$} & {\color{gray}$\boldsymbol{\pm [0.3, 0.3, 0.2]} \, | \, [0.3, 0.3, 0.2]$} & & \\[0.5 ex]
     %\hline
     Gasoy & No &
     \ \ \ $\boldsymbol{1.2} \, | \, 2.3$ & \ \ \ $\boldsymbol{0.0}^* \, | \, 5.0^*$ & \ \ \ $\boldsymbol{[7.7, {\text{-}}1.7, 8.3]} \, | \, [7.9,{\text{-}}1.7, 8.0]$ & $\boldsymbol{0.54} \, | \, 0.54$ & -148.95 \\[0.5 ex]
     & & {\color{gray}$\boldsymbol{\pm 0.0} \, | \, 1.1$} & {\color{gray}$\boldsymbol{\pm 0.0^*} \, | \, 3.2^*$} & {\color{gray}$\boldsymbol{\pm [0.3, 0.4, 0.3]} \, | \, [0.3, 0.4, 0.3]$} & & \\[0.5 ex]
     %\hline
     Stonewall & Yes &
     \ \ \ $\boldsymbol{0.8} \, | \, 3.5$ & \ \ \ $\boldsymbol{0.1}^* \, | \, 2.6^*$ & \ \ \ $\boldsymbol{[11.5, 0.0, 6.1]} \, | \, [11.8, \text{-}0.7, 6.3]$ & $\boldsymbol{0.53} \, | \, 0.53$ & -139.96 \\[0.5 ex]
     & & {\color{gray}$\boldsymbol{\pm 0.1} \, | \, 2.2$} & {\color{gray}$\boldsymbol{\pm 0.0^*} \, | \, 4.4^*$} & {\color{gray}$\boldsymbol{\pm [0.7, 0.7, 0.4]} \, | \, [0.7, 0.6, 0.4]$} & & \\[0.5 ex]
     %\hline
     Gasoy & Yes &
     \ \ \ $\boldsymbol{1.7} \, | \, 2.0$ & \ \ \ $\boldsymbol{0.0}^* \, | \, 2.8^*$ & \ \ \ $\boldsymbol{[6.0, 0.0, 7.2]} \, | \, [6.2, {\text{-}}0.6, 7.7]$ & $\boldsymbol{0.31} \, | \, 0.32$ & -135.20 \\[0.5 ex]
     & & {\color{gray}$\boldsymbol{\pm 0.1} \, | \, 1.1$} & {\color{gray}$\boldsymbol{\pm 0.0^*} \, | \, 3.9^*$} & {\color{gray}$\boldsymbol{\pm [0.3, 1.3, 0.4]} \, | \, [0.3, 0.5, 0.4]$} & & \\[0.75 ex]
     \hline
    \end{tabular}
    \caption{Relating the magnitudes of the various terms in the learned PDE model, $u_t = \nabla \cdot [u (\nabla \mathcal{V} + \nabla\mathcal{K} \! * \! u) + \mathbf{D} \nabla u]$, nondimensionalized via eq.~(\ref{eq:nondimmed_pde}). All results were obtained using test function support radii $\boldsymbol{m} = (10, 10, 6)$. Entries with a dagger ($\dagger$) indicate that synthetically-combined experimental training data from each test case were used, while entries listed in ($\boldsymbol{\cdot} \, | \, \cdot$) order denote the parameters obtained via WSINDy and ordinary least squares, respectively. The (grayed out) value below each parameter  is the standard error. We report AIC scores relative to the least squares solution; i.e., $\Delta{\rm{AIC}} = \Delta{\rm{AIC}}(\mathbf{u}, \mathbf{w}_{\textsc{ws}}, \mathbf{w}_{\textsc{ls}})$. Because it only makes physical sense to learn interaction potentials $\mathcal{K}$ for each experimental run \textit{separately} (two runs were performed for each case), the results reported here neglect this term; for reference, we list the average of the two $K_c$ values (denoted by an asterisk $*$) listed in Table~\ref{table:results_by_run_number}. Note that the learned $\mathcal{K}$ potentials corresponding to these $K_c$ values do not contribute to the reported $R^2$ or $\Delta{\rm{AIC}}$ values.}
    \label{table:full_model}
\end{table}
\end{landscape}

\newpage

%\begin{landscape}
\begin{table}[htb]
    \centering
    \begin{tabular}{||c c c c c c c||}
     \hline
     & & & & & & \\[-2.0 ex]
     \textbf{Plant} & \textbf{Virus} & $\boldsymbol{D_{x} \pm 2\hat{\sigma}}$ & $\boldsymbol{D_{xy} \pm 2\hat{\sigma}}$ & $\boldsymbol{D_{y} \pm 2\hat{\sigma}}$ & $\boldsymbol{R^2}$ & $\boldsymbol{\Delta{\rm{AIC}}}$ \\ [0.5ex]
     \hline
     \hline
     & & & & & & \\[-1.5 ex]
     %\hline
     $\dagger$ & $\dagger$ &
     $8.0 {\color{gray}\pm 0.2}$ & $1.0 {\color{gray}\pm 0.3}$ & $9.0 {\color{gray}\pm 0.2}$ & $0.66$ & \textbf{+15.3} \big| -41.6 \\[0.75 ex] 
     \hline
     & & & & & & \\[-1.5 ex]
     S & $\dagger$ &
     $6.3 {\color{gray}\pm 0.2}$ & $1.9 {\color{gray}\pm 0.3}$ & $6.9 {\color{gray}\pm 0.3}$ & $0.58$ & \textbf{+10.5} \big| -137.9 \\[0.5 ex] 
     %\hline
     G & $\dagger$ &
     $10.6 {\color{gray}\pm 0.2}$ & $-2.6 {\color{gray}\pm 0.5}$ & $11.1 {\color{gray}\pm 0.4}$ & $0.46$ & \textbf{+23.1} \big| -112.8 \\[0.75 ex]
     \hline
     & & & & & & \\[-1.5 ex]
     $\dagger$ & No &
     $6.8 {\color{gray}\pm 0.1}$ & $1.1 {\color{gray}\pm 0.2}$ & $9.1 {\color{gray}\pm 0.2}$ & $0.60$ & \textbf{+30.1} \big| -106.6 \\[0.5 ex] 
     %\hline
     $\dagger$ & Yes &
     $12.3 {\color{gray}\pm 0.4}$ & $0.1 {\color{gray}\pm 0.6}$ & $8.5 {\color{gray}\pm 0.5}$ & $0.56$ & \textbf{+6.3} \big| -134.8 \\[0.75 ex]
     \hline
     & & & & & & \\[-1.5 ex]
     S & No &
     $4.0 {\color{gray}\pm 0.2}$ & $2.5 {\color{gray}\pm 0.3}$ & $7.2 {\color{gray}\pm 0.2}$ & $0.34$ & \textbf{-7.0} \big| -155.4 \\[0.5 ex]
     %\hline
     G & No &
     $11.9 {\color{gray}\pm 0.2}$ & $-1.6 {\color{gray}\pm 0.4}$ & $9.5 {\color{gray}\pm 0.3}$ & $0.50$ & \textbf{+7.1} \big| -141.9 \\[0.5 ex]
     %\hline
     S & Yes &
     $11.1 {\color{gray}\pm 0.5}$ & $-0.9 {\color{gray}\pm 0.6}$ & $6.0 {\color{gray}\pm 0.4}$ & $0.51$ & \textbf{-9.7} \big| -149.7 \\[0.5 ex]
     %\hline
     G & Yes &
     $9.1 {\color{gray}\pm 0.3}$ & $-0.6 {\color{gray}\pm 0.5}$ & $6.8 {\color{gray}\pm 0.4}$ & $0.26$ & \textbf{-9.0} \big| -144.2 \\[0.75 ex]
     \hline
    \end{tabular}
    \caption{Identified diffusion constants for the \textit{purely diffusive} PDE model, $u_t = \nabla\cdot(\mathbf{D} \nabla u)$. Because the proposed model is already sparse, only the values obtained via ordinary least squares are listed. In this case, the reported $\Delta{\rm{AIC}}$ metrics, listed in ($\boldsymbol{\cdot} \, | \, \cdot$) order, are computed relative to the corresponding WSINDy and ordinary least squares models from Table~\ref{table:full_model}, respectively. See Figure~\ref{fig:comparing_D_params} (as well as Figures~\ref{fig:cats_violin_xy}-\ref{fig:cats_violin_cross_term}) in the appendix for the corresponding empirical estimates $\hat{D}_{ij}$.}
    \label{table:purely_diffusive_model}
\end{table}
%\end{landscape}

\begin{table}[htb]
    \centering
    \begin{tabular}{||c c | c | c c c||}
     \hline
     & & & & & \\[-2.0 ex]
     \textbf{Plant} & \textbf{Virus} & $\boldsymbol{\hat{D}_{\rm{eff}} \pm \delta\hat{\mathbf{D}}_{\rm{eff}}}$ & $\boldsymbol{D_{\rm{eff}} \pm 2\hat{\sigma}}$ & $\boldsymbol{R^2}$ & $\boldsymbol{\Delta{\rm{AIC}}}$ \\ [0.5ex]
     \hline
     \hline
     & & & & & \\[-1.5 ex]
     %\hline
     $\dagger$ & $\dagger$ & $6.5 {\color{gray}\pm 1.1}$ &
     $8.3 {\color{gray}\pm 0.1}$ & $0.66$ & +11.4 \\[0.75 ex] 
     \hline
     & & & & & \\[-1.5 ex]
     Stonewall & $\dagger$ & $4.9 {\color{gray}\pm 1.2}$ &
     $6.5 {\color{gray}\pm 0.2}$ & $0.56$ & +21.1 \\[0.5 ex] 
     %\hline
     Gasoy & $\dagger$ & $8.5 {\color{gray}\pm 1.8}$ &
     $10.2 {\color{gray}\pm 0.2}$ & $0.45$ & +5.8 \\[0.75 ex]
     \hline
     & & & & & \\[-1.5 ex]
     $\dagger$ & No & $7.1 {\color{gray}\pm 1.6}$ &
     $7.5 {\color{gray}\pm 0.1}$ & $0.59$ & +10.0 \\[0.5 ex] 
     %\hline
     $\dagger$ & Yes & $5.9 {\color{gray}\pm 1.4}$ &
     $11.0 {\color{gray}\pm 0.3}$ & $0.56$ & +4.9 \\[0.75 ex]
     \hline
     & & & & & \\[-1.5 ex]
     Stonewall & No & $4.0 {\color{gray}\pm 1.5}$ &
     $5.3 {\color{gray}\pm 0.1}$ & $0.30$ & +11.8 \\[0.5 ex]
     %\hline
     Gasoy & No & $11.4 {\color{gray}\pm 2.9}$ &
     $10.9 {\color{gray}\pm 0.2}$ & $0.49$ & -0.1 \\[0.5 ex]
     %\hline
     Stonewall & Yes & $5.9 {\color{gray}\pm 1.9}$ &
     $8.7 {\color{gray}\pm 0.4}$ & $0.48$ & +8.8 \\[0.5 ex]
     %\hline
     Gasoy & Yes & $6.1 {\color{gray}\pm 2.0}$ &
     $8.2 {\color{gray}\pm 0.2}$ & $0.26$ & -3.2 \\[0.75 ex]
     \hline
    \end{tabular}
    \caption{Learned constants for the \textit{isotropic} and \textit{purely diffusive} PDE model $u_t = D_{\rm{eff}} \Delta{u}$. Because the proposed model is already sparse (i.e., it has a single parameter), only the values obtained via ordinary least squares are listed. Here, each $\Delta{\rm{AIC}}$ metric is computed relative to the corresponding anisotropic model from Table~\ref{table:purely_diffusive_model}. For a comparison of the corresponding direct empirical estimates $\hat{D}_{\rm{eff}}$, also see Figures~\ref{fig:caterpillars_violin_r} and \ref{fig:comparing_D_params}.}
    \label{table:Deff_model}
\end{table}

\newpage

In terms of the influence of infection status and plant resource quality on population dispersal rates, the empirical results listed in Figure~\ref{fig:caterpillars_violin_r} and Table~\ref{table:Deff_model} indicate that: \begin{enumerate}
    \item[(i)] infected larvae \textit{are not inherently} slower or faster than uninfected larvae -- the relationship between dispersal rates and infection is complex (cf. \cite{OsnasHurtadoDobson2015AmNat, Goulson1997Oecologia});
    %infected larvae \textit{do not} systemically disperse faster or slower than uninfected larvae (cf. \cite{OsnasHurtadoDobson2015AmNat, Goulson1997Oecologia});

    \item[(ii)] in general, larvae \textit{do} tend to disperse systematically faster on the high-quality resource, Gasoy, than on the low-quality variety, Stonewall (cf. \cite{ShikanoShumakerPeifferEtAl2017Oecologiaa}).
\end{enumerate} Interestingly, while in general (ii) holds with little variance, the dynamics of uninfected larvae in particular appear to have a sensitive dependence on resource quality, i.e., \begin{enumerate}
    \item[(iii)] a change in resource quality elicits a dramatic response from uninfected larvae, with individuals dispersing appreciably faster in an environment featuring the high-quality resource (Gasoy), rather than low-quality resource (Stonewall), variety (see also Figure~\ref{fig:comparing_D_params} in the appendix).
\end{enumerate} In summary, infected individuals are not found to disperse faster or slower than uninfected individuals uniformly. Rather, this relationship depends on other environmental factors such as resource quality, which primarily affect the dispersal rates of the uninfected larvae. However, more data are be required to make a conclusive claim about the nature of this mechanism.

\subsection{Model Assessment and Comparison}
\begin{figure*}
    \centering
    \includegraphics[width=\linewidth]{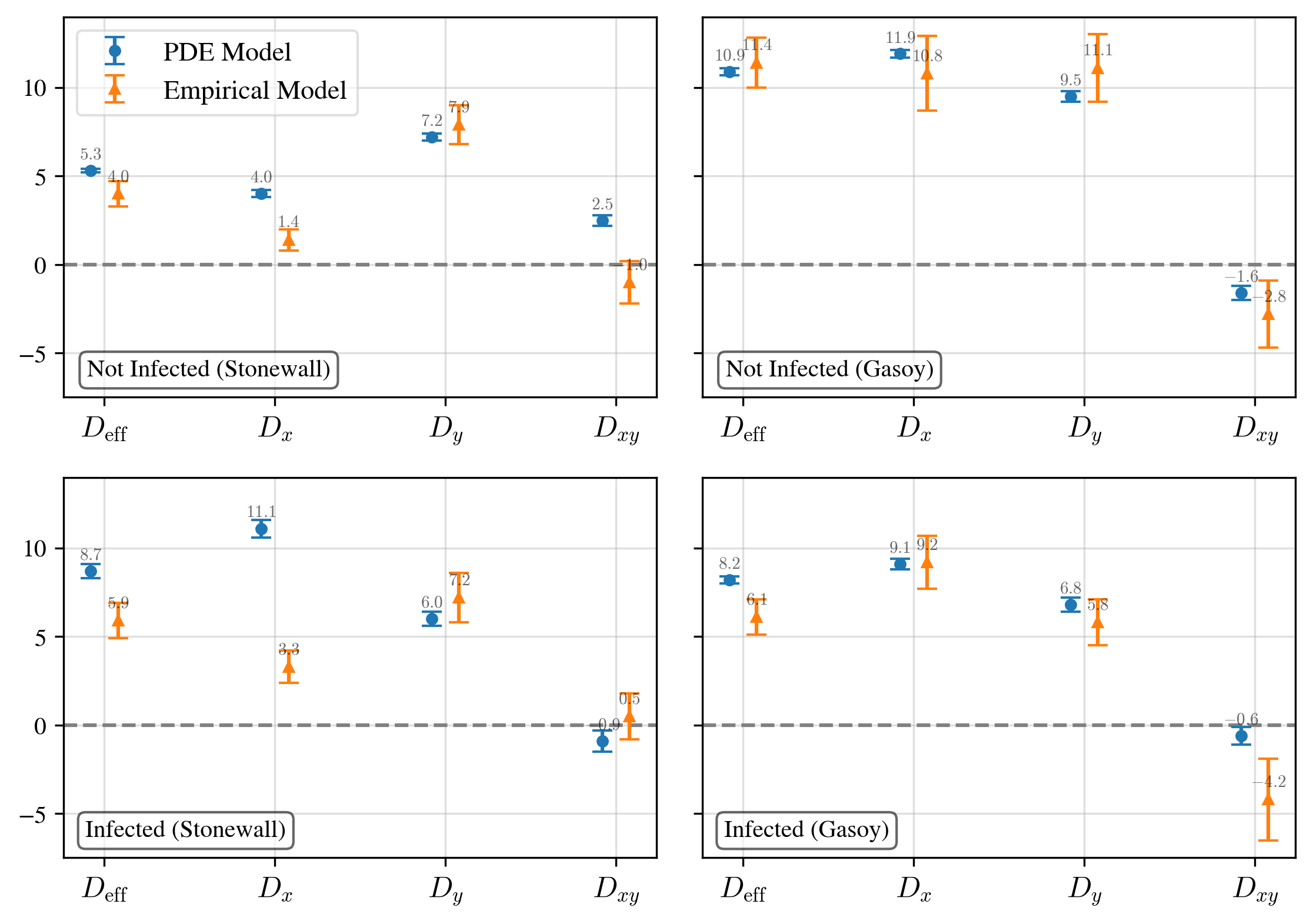}
    \caption{Comparing empirical and weak-form PDE estimates of the diffusion coefficients $D_{ij}$ for each control population in Tables~\ref{table:purely_diffusive_model} and \ref{table:Deff_model}.}
    \label{fig:comparing_D_params}
\end{figure*}

The major qualitative results latent in the empirical data, discussed in \S\ref{sec:raw_data} above, are largely in agreement with the data-driven PDE model results listed in Tables~\ref{table:full_model}, \ref{table:purely_diffusive_model}, and \ref{table:Deff_model}. Namely, the identified PDE models reaffirm that:\begin{enumerate}
    \item[(i)] infected larvae \textit{are not inherently} slower or faster than uninfected larvae vis-\`a-vis dispersal,

    \item[(ii)] larvae tend to disperse faster on a higher-quality plant resource (Gasoy) than on a lower-quality resource (Stonewall),

    \item[(iii)] uninfected larvae elicit more dramatic response to a change in resource quality than infected larvae.
\end{enumerate} Although the forms of the PDE models in Tables~\ref{table:full_model}-\ref{table:Deff_model} vary significantly, the resulting diffusion constant estimates remain remarkably consistent (i.e., distinct PDE models produced similar $D_{ij}$ estimates on the same training data). Moreover, Figure~\ref{fig:comparing_D_params} indicates that these PDE estimates are consistent with the trends exhibited by the empirical data, excluding the $D_x$ parameter in the infected, Stonewall case.\footnote{Note, however, that the identified PDEs tend to identify larger effective diffusion constants $D_{\text{eff}}$ than the direct empirical estimates $\hat{D}_{\text{eff}}$; see Table~\ref{table:Deff_model}.}

Comparing the McKean-Vlasov models listed in Table~\ref{table:full_model} with the idealized and purely-diffusive models of Tables~\ref{table:purely_diffusive_model} and \ref{table:Deff_model}, we observe that the addition of parameterized environmental and interaction potentials $\mathcal{V}_{\mathbf{w}}$ and $\mathcal{K}_{\mathbf{w}}$ into the data-driven model increase the corresponding $R^2$ values by roughly $5\%$ to $10\%$, relative to the idealized models. Since these increases are relatively small compared to increase in model complexity, this result indicates that the \textit{anisotropic} or \textit{effective} diffusion models are sufficient to capture the majority of the variance of the data in most cases. Still, our results indicate that the sparsely-weighted McKean-Vlasov PDE models are the `AIC preferred' models in each case of synthetically-combined training data featuring mixed control populations. When separating the training data by control population (inducing large variance via the fewest number of samples), the AIC-preferred model instead becomes either the idealized \textit{anisoptropic} or \textit{effective} model (see Tables~\ref{table:purely_diffusive_model}-\ref{table:Deff_model}).

\newpage

\begin{figure*}
    \centering
    \includegraphics[width=0.49\linewidth]{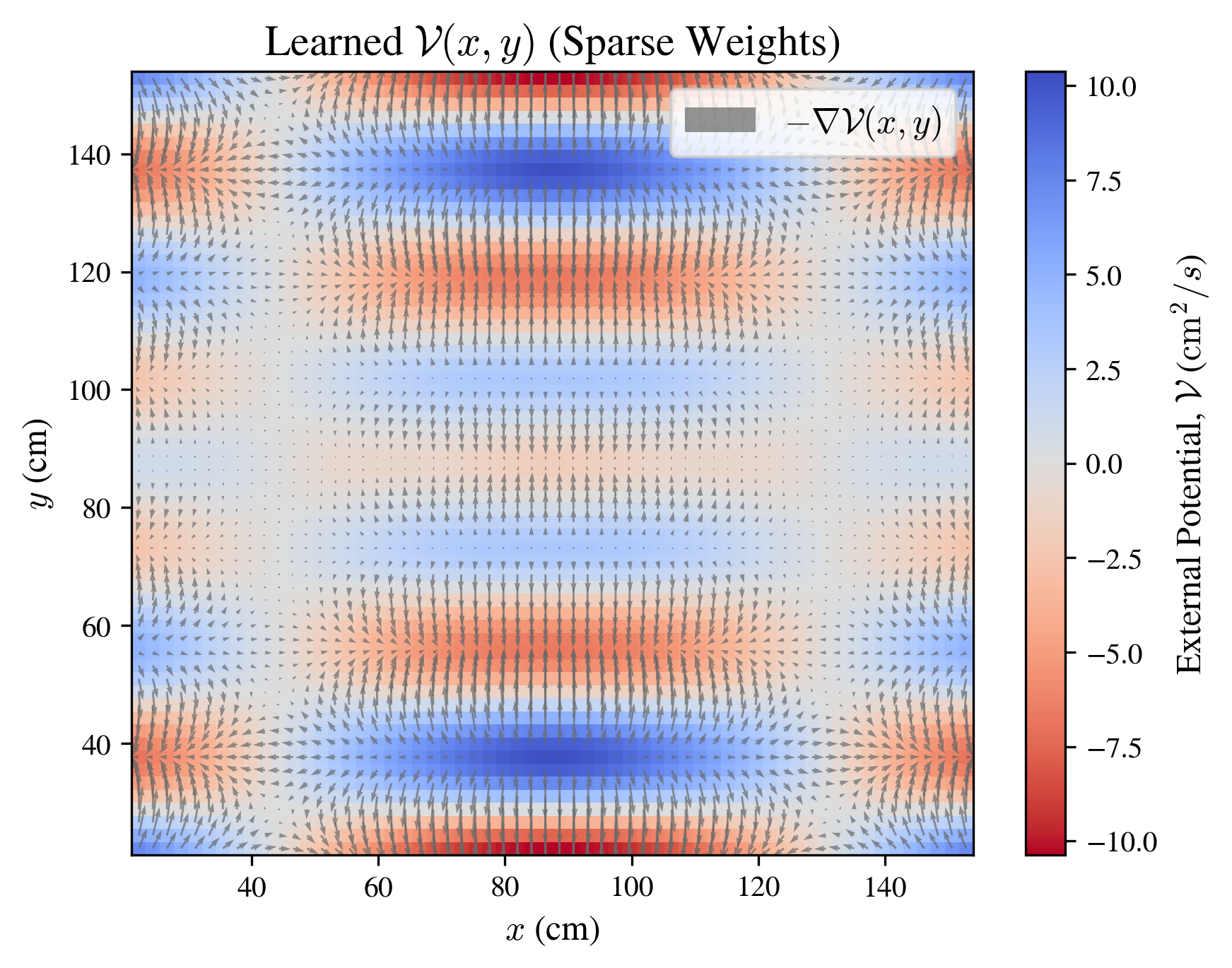}
    %\quad
    \includegraphics[width=0.49\linewidth]{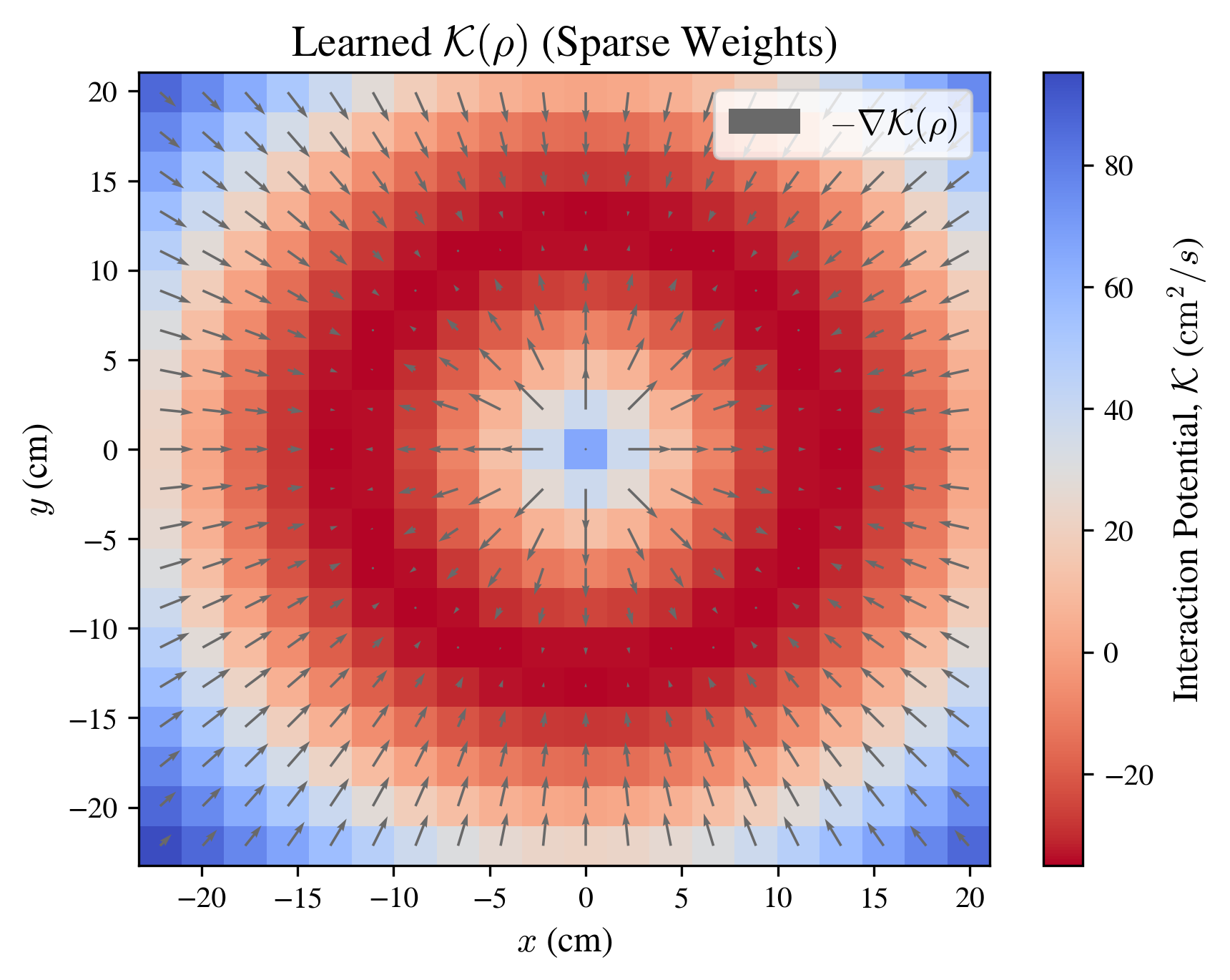}
    \caption{Visualizing the learned environmental potential $\mathcal{V}$ and interaction potential $\mathcal{K}$ for the empirical distributions $\mu(\boldsymbol{x}; \mathbf{X}_t)$ and $\mu(\boldsymbol{x}; \mathbf{X}_t^{3,1})$, respectively. Note that the learned $\mathcal{V}$ resembles the soybean plant spacing in each domain.}
    \label{fig:learned_potentials}
\end{figure*}

Interestingly, of the two categories of `force' potentials represented in eq.~(\ref{eq:SDE}), the environmental potential $\mathcal{V}$ appears to have the largest influence on the dispersal dynamics (see Table~\ref{table:full_model}). As one might intuitively expect, the learned parameterized expansions $\mathcal{V}_{\mathbf{w}}$ tend to reflect the underlying spatial distribution of plant resources; see Figure~\ref{fig:learned_potentials}, left panel. Although the interaction potential $\mathcal{K}$ has a weaker effect on the dynamics in terms of a dominant balance, the learned $\mathcal{K}_{\mathbf{w}}$ indicate that the larvae are weakly attracted to each other at large distances but extremely repulsive at close distances; see Figure~\ref{fig:learned_potentials}, right panel.

\subsection{Sensitivity and Error Analysis}\label{sensitivity_analysis}
\begin{figure*}
    \centering
    %\includegraphics[width=0.3\linewidth]{pictures/cats_wsindy_weak_lhs.png}
    %\quad
    %\includegraphics[width=0.3\linewidth]{pictures/cats_wsindy_weak_rhs.png}
    %\quad
    %\includegraphics[width=0.3\linewidth]{pictures/cats_wsindy_absolute_error.png}
    \includegraphics[width=\linewidth]{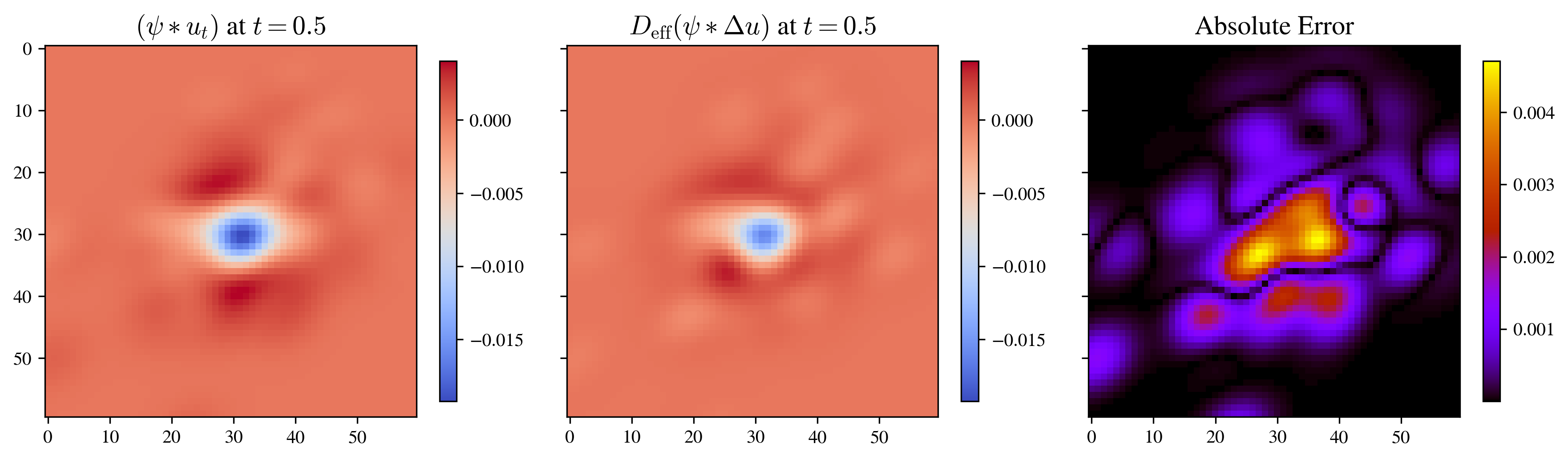}
    \\
    \hspace*{-1 cm}
    \includegraphics[width=0.47\linewidth]{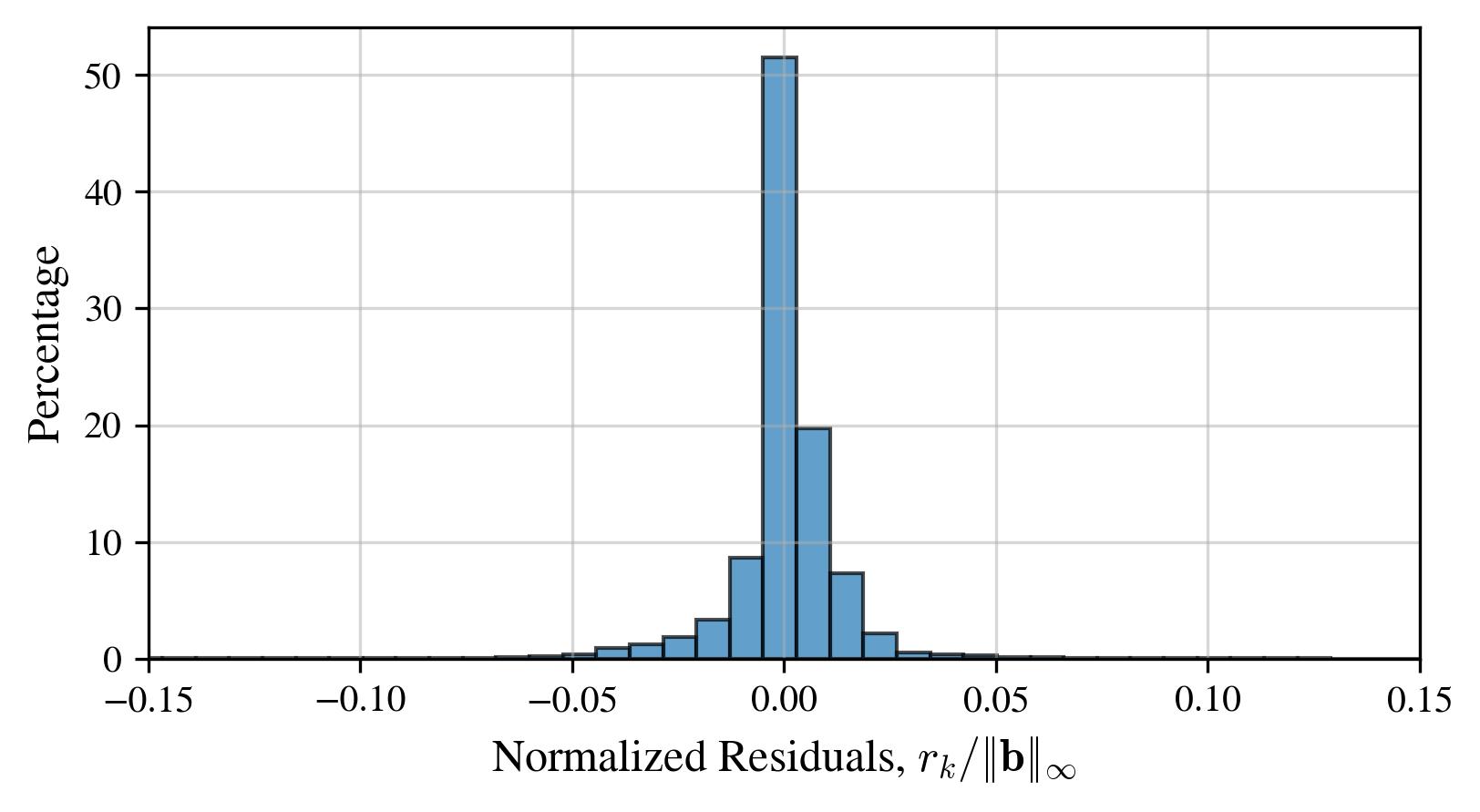}
    \includegraphics[width=0.45\linewidth]{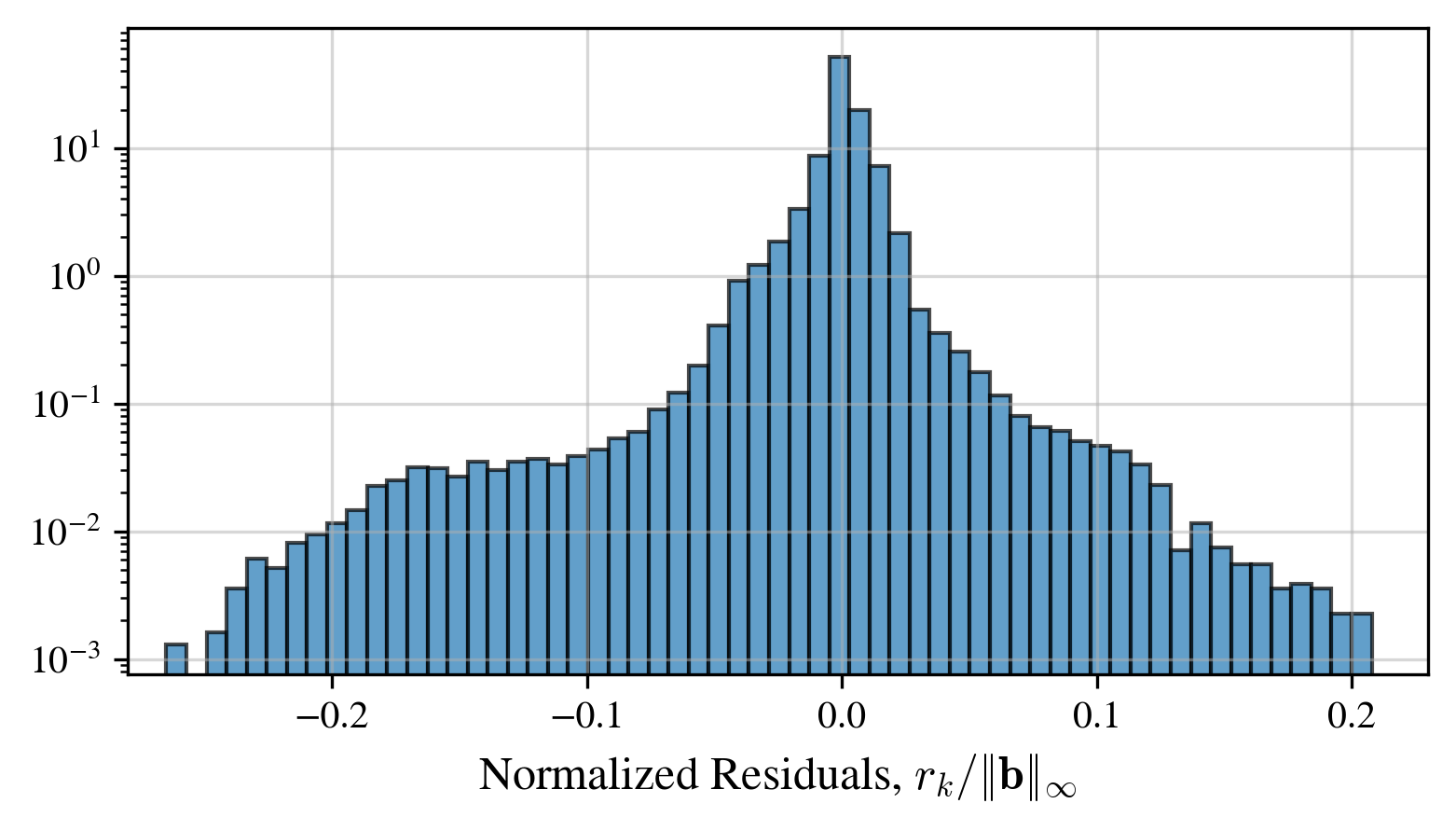}
    \caption{(Top) An example of pointwise error between $(\psi * u_t)$ and the weak-form feature $D_{\rm{eff}}(\psi * \Delta u)$, in this case for the first entry of Table~\ref{table:Deff_model}. (Bottom) Histogram of the corresponding fit residuals $\mathbf{r}$, exhibiting a typical peaked distribution.}
    \label{fig:wsindy_residual_plots}
\end{figure*}

In \S\ref{sec:error_analysis} of the appendix, we include a brief error analysis vis-\`a-vis the Gaussian KDE process described in \S\ref{sec:numerical_methods}; in particular, we show that the expected bias induced by this process is $\mathcal{O}(\sigma/h)$. Moreover, \S\ref{sec:residuals_and_parameter_error} includes histograms of the fitted residual vectors $\mathbf{r} = \mathbf{b} - \mathbf{G}\mathbf{w}$ (see Figure~\ref{fig:wsindy_residual_plots}), where the vector of weights $\mathbf{w}$ is either: computed via sparse regression as per eq.~(\ref{eq:wsindyLoss}),\footnote{In practice, we use a normalized version of the loss function $\mathcal{L} = \mathcal{L}(\mathbf{w}; \mathbf{b}, \mathbf{G})$ given in eq.~(\ref{eq:wsindyLoss}); see eq.~(\ref{eq:wsindy_loss}) in the appendix for more information.} or given by the OLS estimator. As is typical of errors-in-variables regression in the context of PDEs, the fitted residuals $\{r_k\}$ appear to be drawn from product-like (e.g., Bessel-function type) distributions, suggesting that an iteratively-reweighted least squares optimization approach may improve the parameter estimates; see, e.g., the `WENDy' algorithm \cite{BortzMessengerDukic2023BullMathBiol}. Finally, Figure~\ref{fig:hyperparam_sweep} indicates the level of sensitivity of the $D_{\text{eff}}$ parameter estimates to the support radii $\boldsymbol{m} = (m_x, m_y, m_t)$.
%WSINDy's primary algorithmic hyperparameter, the test function support radii $\boldsymbol{m} = (m_x, m_y, m_t)$.

%\newpage

\section{Discussion}
\label{Sec:disc}
In this paper, we have adapted the weak form modeling framework of WSINDy in the context of lepidopteran larval dispersal. The data-driven methodology used here builds off of the mean-field approach presented in \cite{MessengerBortz2022PhysicaD}, extending it to accommodate model terms describing larval dispersal, larva-to-larva interactions, and interactions of larvae with their environment. Besides illustrating the promise of the modeling technique, the ecological purpose of this study was to make quantitative estimates of the larval diffusion constants $D_{ij}$, as well as to determine how infection status and resource quality affect movement dynamics.

\newpage

A primary benefit of using a symbolic and PDE-based modeling approach in the context of insect dispersal is the ability to quantitatively characterize the dominant balance of various mechanisms in the dynamics. In particular, our results suggest that the dominant contributions to the dispersal dynamics listed in eq.~(\ref{eq:PDE}) are: (1) the diffusion term $\nabla\cdot(\mathbf{D}\nabla{u})$ (associated with random movement), followed in importance by (2) the environmental potential term $\nabla\cdot(u\nabla\mathcal{V})$ (associated with non-homogeneous terrain and plant resource distribution), and most weakly (3) by the non-linear interaction `force' $\nabla\cdot{u}(\nabla\mathcal{K} * u)$ between individuals (associated with social repulsion or attraction). As might be intuitively expected, the parameterized external potentials $\mathcal{V}_{\mathbf{w}}(x,y)$ identified by the data mimic the underlying plant crop spacing. Moreover, in cases where the interaction potential force is relevant, the identified parameterized kernels $\nabla \mathcal{K}_{\mathbf{w}}(\boldsymbol{x};\boldsymbol{x}')$ indicate the existence of a preferred inter-larva spacing. We note that the relatively small interaction force observed between individuals may be the result of an abundance of plant resources precluding overcrowding, as one would normally expect a non-negligible contribution due to the larvae's predilection for cannibalism \cite{VanAllenDillemuthDukicEtAl2023Oecologia}. Lastly, we emphasize that the PDE models using sparse weights and OLS weights are both internally consistent and qualitatively consistent with the raw experimental data.

We have found that idealized and spatially-uncorrelated surrogate models of the form $u_t \approx D_{\rm{eff}}\Delta{u}$ are effective approximations of the dynamics; i.e., these idealized models are sufficient to capture the majority of the variance of the dynamics in many instances. Of the tested PDE models, the idealized models tend to be `AIC optimal' whenever the corresponding training data consists only of the separate control populations. However, in cases with synthetically combined training data, the information criterion favors the full McKean-Vlasov models, suggesting that non-random mechanisms become statistically relevant with sufficient data. Furthermore, while both the identified PDE models and experimental data indicate that: (1) infected larvae are \textit{not} systematically slower or faster than uninfected larvae, and (2) larvae tend to disperse faster on high-quality plant resources than on low-quality varieties, a more nuanced interaction is observed between infection status and resource quality. In particular, the uninfected larvae are observed to elicit more dramatic response to a change in resource quality than the infected larvae.

Finally, we conclude with a brief survey of natural extensions of this work. Our general approach using data-driven PDE modeling frameworks such as WSINDy could be used to inform agricultural pest management strategies (e.g., \textit{trap-cropping} or \textit{inter-cropping}) by quantifying how environmental changes are expected to alter pest dispersal. From a methodological perspective, future work might also consider improving the realism of the candidate models by, e.g., incorporating compartmental models of disease and/or population dynamics, accounting for the effect of predators, or by incorporating dynamics along the $z$-axis. Lastly,  the precision of the identified  dynamics is undoubtedly limited by the sparsity of the current experimental datasets, and we expect that parameter estimates and model identification results could be substantially improved by an expanded store of experimental and field data, an area which we regard as a fruitful avenue for future ecological research.

\newpage

\section*{Acknowledgments}
The authors wish to thank Prof.~Greg Dwyer and Dr.~Katie Dixon (University of Chicago, Department of Ecology \& Evolution) for helpful discussions  regarding ecological applications and Dr.~Daniel Messenger (Los Alamos National Lab) for insight regarding weak form scientific machine learning methods.

\section*{Data Access}
All data and software used to generate the results in this work are listed on Zenodo: \url{https://zenodo.org/records/17156064}. Also see the following GitHub repository: \url{https://github.com/MathBioCU/WSINDy4Dispersal}.

\section*{Competing Interests}
The authors declare no competing interests.

\section*{Disclaimer}
Any opinions, findings, and conclusions or recommendations expressed in this material are those of the author(s) and do not necessarily reflect the views of the National Institutes of Food and Agriculture, Health, or the National Science Foundation.

\section*{Funding}
This research was supported in part by the NIFA Biological Sciences Grant 2019-67014-29919, in part by the NSF Division Of Environmental Biology Grant 2109774, and in part by the NIGMS Division of Biophysics, Biomedical Technology and Computational Biosciences grant R35GM149335. This study was also funded in part by USDA grant 2019-67014-29919 and NSF grant 1316334 as part of the joint NSF–NIH–USDA Ecology and Evolution of Infectious Diseases program. This work utilized the Blanca condo computing resource at the University of Colorado Boulder. Blanca is jointly funded by computing users and the University of Colorado Boulder.

% \aucontribute{
% S.M.: conceptualization, formal analysis, investigation, methodology, software, validation, visualization, writing—original draft, writing—review and editing;
% B.E.: conceptualization, funding acquisition, investigation, methodology, project administration, supervision, visualization, writing—original draft, writing—review and editing;
% B.V-A.: conceptualization, data curation, investigation;
% D.B.: investigation, methodology, supervision, visualization, writing—review and editing;
% V.D.: conceptualization, funding acquisition, investigation, methodology, project administration, supervision, visualization, writing—review and editing.
% }

%\newpage

\bibliographystyle{siamplain}
\bibliography{references}

\newpage

%%%%%%%%%% APPENDIX %%%%%%%%%%%%%%

\section{Appendix}

\subsection{Experimental Setup}\label{App:experiment_details}
One of the many agricultural crops that the fall armyworm feeds on is soybean \cite{PerucaCoelhoDaSilvaEtAl2018Arthropod-PlantInteract}. Soybeans come in numerous genotypes/varieties and these varieties differ in their chemical and physical defenses that they employ against herbivores. Some of the varieties have strong constitutive defenses that interfere with larval consumption of the plant, while other varieties have strong induced defenses \cite{UnderwoodRausherCook2002Oecologia}. As compared to constitutive defenses that are continually present in the plant, induced defenses are only produced after the plant has experienced some herbivory.  Different varieties can thus have differing effects on consumption and virus-induced mortality \cite{ShikanoShumakerPeifferEtAl2017Oecologiaa}; specifically, difference in the chemical constituency of the defense may affect infection rates and the production of viral particles by an infected larva.  These defenses against herbivory also affect the quality of the leaf tissue and can negatively impact growth rates in the fall armyworm \cite{ShikanoShumakerPeifferEtAl2017Oecologiaa}. Consequently, this may lead to changes in dispersal rates amongst individual larvae. 

To directly quantify how infection status and resource quality alter movement dynamics, we conducted a series of four experiments where we measured the movement of fall armyworm larvae across an artificial landscape in the lab. The landscape consisted of four 175 cm $\times$ 175 cm plots, constructed from wood and filled with a standard soil mixture (Sunshine Grow Mix, Agawam, MA). %{\color{red}City, State}).  
Inside of the plot, we placed
45 evenly-spaced mature soybean plants with at least five tri-foliate leaves.  In order to simulate common farming practices, the plants were organized into five rows of nine plants in each plot. Each of the plants had at least five tri-foliate leaves.  We varied resource quality by using two  varieties of soybean that differed in their constitutive anti-herbivore defenses \cite{UnderwoodRausherCook2002Oecologia, ShikanoShumakerPeifferEtAl2017Oecologiaa}. These varieties were \textit{Stonewall}, which we considered as having a relatively high constitutive defense, and \textit{Gasoy}, which we considered as having a relatively low constitutive defense \cite{UnderwoodMorrisGrossEtAl2000Oecologia,UnderwoodRausherCook2002Oecologia}. The Stonewall variety could thus be considered a poor-quality resource as compared to the Gasoy variety.  

To examine the effect of infection status, we fed recently molted fourth-instar larvae a small diet cube (Southland Products, Conway Lake, Arkansas) inoculated with 3 $\mu$l of DI water.  The droplet either contained no virus or $3 \cdot 10^5$ viral particles, which is a dose that would cause the larvae to die of infection at least 95\% of the time (Elderd, \textit{unpublished data}). To ensure that the larvae ate the entire dose, all food was withheld for 24 hours prior to the experiment. 

At the start of the experiment, we placed 20 fourth-instar larvae at the center of each of the four plots, on a single soybean plant. Each plot was planted with either the  Stonewall or Gasoy variety, and received either infected or uninfected larvae. The larvae were contained on the center plant for two hours by placing a plastic tube made of Dura-Lar (Maple Heights, OH) over the plant. This allowed the larvae to settle on the plant after placement. After removing the tube, we measured the location of individual larvae along $x$, $y$, and $z$-axes. The $(x,y)$ measurements correspond to the location of the larvae in the plot, while the $z$-axis measurement indicates the height of the larva, with zero corresponding to the soil-level and any point above zero being the location of the larvae on a soybean plant. Each plot was searched for 15 minutes at eight non-uniformly spaced times ($0, \, 1, \, 2, \, 4, \, 8, \, 16, \, 24,$ and $48$ hours) after the start of the experiment. The positions of all larvae found were recorded. For each combination of plant variety and infection status, we conducted the experiment twice.

\newpage

\subsection{Nondimensionalization Details}\label{sec:nondimensionalization_details}
Consider a symmetric rescaling of the form $\boldsymbol{x} = \mathbf{A}\boldsymbol{\xi}$, with $\mathbf{A} = \mathbf{A}^T$, and define $\bar{\nabla} := \nabla_{\!\boldsymbol{\xi}}$ with $\nabla = \nabla_{\boldsymbol{x}}$. For scalar-valued functions $f(\boldsymbol{x}(\boldsymbol{\xi})) = f(\mathbf{A}\boldsymbol{\xi})$, we have \begin{align*}
    \bar{\nabla} f
    =
    \mathbf{A} \nabla f,
    \quad \text{so that} \quad
    \nabla f = \mathbf{A}^{-1} \bar{\nabla} f.
\end{align*} Similarly, for vector-valued functions $\vec{\boldsymbol{f}}(\boldsymbol{x}(\boldsymbol{\xi})) = \vec{\boldsymbol{f}}(\mathbf{A}\boldsymbol{\xi})$, we have \begin{align*}
    \bar{\nabla} \cdot \vec{\boldsymbol{f}}
    =
    \nabla \cdot \mathbf{A}\vec{\boldsymbol{f}},
    \quad \text{so that} \quad
    \nabla \cdot \vec{\boldsymbol{f}} = \bar{\nabla} \cdot \mathbf{A}^{-1} \vec{\boldsymbol{f}}.
\end{align*} Note also that under the transformation $\boldsymbol{x} \mapsto \boldsymbol{\xi}$, the Jacobian determinant becomes $dx \, dy \, \mapsto \, |\mathbf{A}| \,  d\xi \, d\eta$. Introducing a temporal rescaling $t = \tau t_c$ for dynamics quantities of the form $u(\boldsymbol{x}(\boldsymbol{\xi}), t(\tau)) = u(\mathbf{A}\boldsymbol{\xi}, \tau t_c)$, we find that \begin{align*}
    \frac{\partial{u}}{\partial\tau} = t_c \cdot \frac{\partial{u}}{\partial{t}}.
\end{align*} Applying the coordinate transformation to the PDE in eq.~(\ref{eq:PDE}), we find that \begin{align*}
    \frac{u_{\tau}}{t_c} &= \bar{\nabla} \cdot \mathbf{A}^{-1} \Big( u \mathbf{A}^{-1} \big(\bar{\nabla} \mathcal{V} + |\mathbf{A}|\bar{\nabla}\mathcal{K} \! \star \! u\big) + \mathbf{D} \mathbf{A}^{-1}\bar{\nabla} u\Big),
\end{align*} where \begin{align*}
    \big(\bar{\nabla}\mathcal{K} \star u\big)(\mathbf{A}\boldsymbol{\xi}, \tau t_c)
    :=
    \int\!\!\!\!\int_{\bar{\Omega}} \bar{\nabla}\mathcal{K}\big(\big|\mathbf{A}(\boldsymbol{\xi} - \boldsymbol{\xi}')\big|\big) \, u\big(\mathbf{A}\boldsymbol{\xi}', \tau t_c \big) \, d\xi' \, d\eta'.
\end{align*} We now introduce the dimensionless quantities \begin{align*}
    U(\xi, \eta, \tau) := U_c^{-1} \, u(\mathbf{A}\boldsymbol{\xi}, \, \tau t_c),
    \quad \text{with} \quad
    \begin{cases}
        %U(\xi, \eta, \tau) := U_c^{-1} u(\mathbf{A}\boldsymbol{\xi}, \, \tau t_c),
        %\\
        V(\xi, \eta) := V_c^{-1} \, \mathcal{V}(\mathbf{A}\boldsymbol{\xi}),
        \\
        K(\xi, \eta) := K_c^{-1} \, \mathcal{K}(\mathbf{A}\boldsymbol{\xi}),
    \end{cases}
\end{align*} where substitution into the rescaled PDE above, and a bit of subsequent simplification, then yields \begin{align*}
    U_{\tau}
    &=
    \bar{\nabla} \cdot t_c\mathbf{A}^{-1} \Big( U \mathbf{A}^{-1} \big(V_c\bar{\nabla}V + K_cU_c|\mathbf{A}| \bar{\nabla}K \! \star \! U\big) + \mathbf{D} \mathbf{A}^{-1}\bar{\nabla} U\Big)
    \\
    &=
    \bar{\nabla} \cdot \Big[ \Big(t_cV_c\mathbf{\Lambda}^{-1}\Big) U\bar{\nabla}V + \Big(t_cK_cU_c |\mathbf{\Lambda}|^{\frac{1}{2}} \mathbf{\Lambda}^{-1}\Big) U\big(\bar{\nabla}K \! \star \! U\big) + \Big(t_c \, \mathbf{A}^{-1} \mathbf{D} \mathbf{A}^{-1}\Big) \bar{\nabla} U\Big].
\end{align*} Here, we've used the fact that $\mathbf{D} := \frac{1}{2} \boldsymbol{\sigma} \boldsymbol{\sigma}^T$ and defined the Gram matrix $\mathbf{\Lambda} := \mathbf{A}^T\mathbf{A}$ for notational convenience.

\begin{table}[htb]
    \centering
    \begin{tabular}{||c c c c||}
         \hline
         \textbf{Variable} & \textbf{Definition} & \textbf{Dimensions} & \textbf{Units} \\ %[0.1ex] 
         \hline%\hline
         & & & \\[-1.75 ex]
         $\big(x^i_t, \, y^i_t\big)$ & Position measurements & $\mathbf{L}$ & cm \\[0.5 ex] 
         %\hline
         $u(x,y)$ & Probability density & $\mathbf{L}^{-2}$ & ${\rm{cm}}^{-2}$ \\[0.5 ex] 
         %\hline
         $\mathcal{V}(x,y)$ & Environmental potential & $\mathbf{L}^{2}\mathbf{T}^{-1}$ & ${\rm{cm}}^{2} s^{-1}$ \\ [0.5ex] 
         %\hline
         $\mathcal{K}(\rho)$ & Interaction potential & $\mathbf{L}^{2}\mathbf{T}^{-1}$ & ${\rm{cm}}^{2} s^{-1}$  \\ [0.5ex] 
         %\hline
         $D_{ij}$ & Diffusion constant & $\mathbf{L}^{2}\mathbf{T}^{-1}$ & ${\rm{cm}}^{2} s^{-1}$  \\ [1ex] 
         \hline
    \end{tabular}
    \label{table:physical_dimensions}
    \caption{Physical dimensions of the quantities involved in the SDE of eq.~(\ref{eq:SDE}) and PDE of eq.~(\ref{eq:PDE}).}
\end{table}

\newpage

\begin{figure*}
    \centering
    \includegraphics[width=\linewidth]{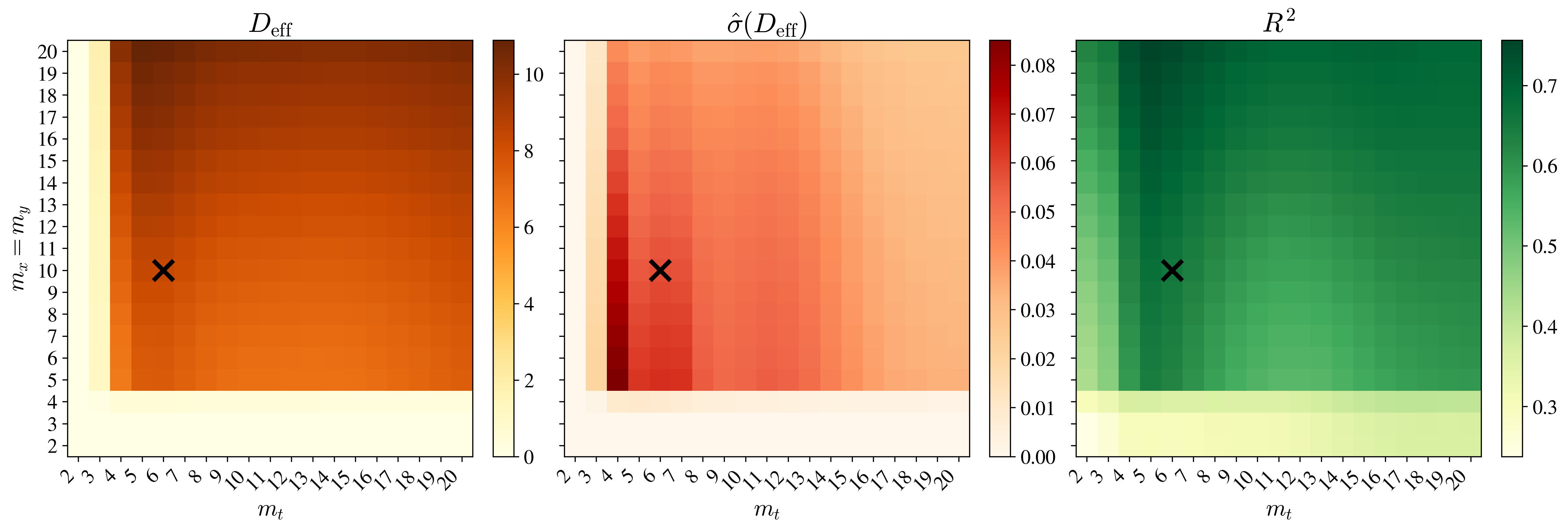}
    \caption{A hyperparameter sweep illustrating the sensitivity of the effective diffusion constants $D_{\text{eff}}$ predicted by WSINDy to changes in the test function support radii $\boldsymbol{m} = (m_x, m_y, m_t)$. Here, we use $m_x = m_y$ and plot an `$\boldsymbol{\times}$' at $(m_x,m_t) = (10,6)$.}
    \label{fig:hyperparam_sweep}
\end{figure*}

\subsection{Additional Implementation Details}\label{sec:additional_implementation_details}
\begin{figure*}
    \centering
    % \includegraphics[width=0.32\linewidth]{pictures/cats_wsindy_u.png}
    % \quad
    % \includegraphics[width=0.32\linewidth]{pictures/cats_wsindy_weak_u.png}
    \includegraphics[width=\linewidth]{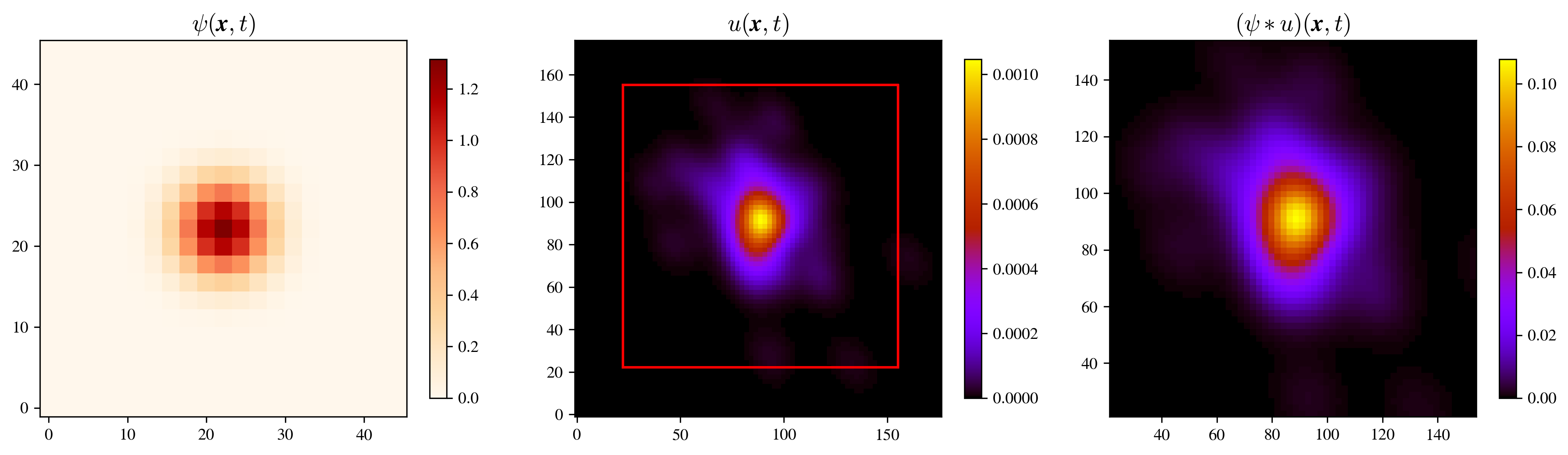}
    \caption{Illustrating the weak-form feature $(\psi * u)(\boldsymbol{x},t)$ at time $t=1$ for the ensemble distribution $\mu(\mathbf{X}_t)$, given the chosen test function support radii $\boldsymbol{m}=(10, 10, 6)$. We manually select test function spectra $|\hat{\psi}|$ that induce minimal smoothing.}
    \label{fig:test_fcn_filter}
\end{figure*}

As mentioned in \S\ref{sec:numerical_methods}, the primary set of WSINDy hyperparameters are the test function support radii, \begin{align*}
    \boldsymbol{m} = (m_x, m_y, m_t),
\end{align*} which determine the amount of `smoothing' that is applied to $\mathbf{u}$, i.e., determining the bandwidth of the kernel $\psi$. In our specific case, we find that that na\"ive methods for selecting $\boldsymbol{m}$ lead to over-smoothed data $\psi * \mathbf{u}$ and, in turn, learned models with spuriously large $R^2$ values which over-emphasize the diffusion term $\Delta{u}$; see the hyperparameter sweep in Figure~\ref{fig:hyperparam_sweep}. To prevent this, we select the radii \begin{align*}
    \boldsymbol{m} = (10, 10, 6)
\end{align*} by manually matching Fourier spectra such that \begin{align*}
    \mathcal{F}[\mathbf{u}] \approx \mathcal{F}[\psi * \mathbf{u}].
\end{align*} We plot the resulting weak-form features $\psi * \mathbf{u}$ in Figure~\ref{fig:test_fcn_filter}. Correspondingly, we use test function degrees given by \begin{align*}
    \boldsymbol{p} = (14, 14, 20).
\end{align*} We use a uniformly-spaced grid of 309,600 query points $\{(\boldsymbol{x}_k, t_k)\}^{\kappa}_{k=1}$ throughout; see Table~\ref{table:numerical_details}. Moreover, we respectively compute the characteristic dimensional constants $V_c$ and $K_c$ via \begin{align*}
    V_c := \|\nabla \mathcal{V}_{\mathbf{w}}\|_2
    \quad \text{and} \quad
    K_c := \|\nabla \mathcal{K}_{\mathbf{w}} * \hat{u}_h\|_2.
\end{align*} Lastly, we note that during the model discovery process, the discrete interaction potential $\mathbf{K}$ was pre-scaled by a factor of $U^{-1}_c$ (i.e., $\beta_n \mapsto \beta_n/U_c$) to avoid scaling issues, where we use \begin{align*}
    U_c := \| \hat{u}_h \|_{\infty} = \mathcal{O}\big(10^{-2}\big).
\end{align*}

\newpage

To solve the sparse regression problem posed in eq.~(\ref{eq:wsindyLoss}), we use the Modified Sequential Thresholding Least Squares (MSTLS) algorithm formulated in \cite{MessengerBortz2021JComputPhys}. In MSTLS, a sparse vector of model weights $\mathbf{w}^{\star}$ is obtained by minimizing a normalized version of the loss function $\mathcal{L}$ given in eq.~(\ref{eq:wsindyLoss}) over a set of increasing \textit{thresholding parameters} $\{\lambda_i\}^{N_{\lambda}}_{i=1} \subset (0,1)$,\footnote{We follow \cite{MessengerBortz2021JComputPhys} in scanning over a set of candidate values $\big\{\lambda_i\big\}^{50}_{i=1}$ defined by uniformly log-spaced increments $\log_{10}(\lambda_i) \in (-4,0)$.} \begin{align}\label{eq:wsindy_loss}
    \mathbf{w}^{\star} := \texttt{MSTLS}\Big(\mathbf{b}, \, \mathbf{G}, \, \text{arg}\!\!\!\!\min_{\lambda \in \{\lambda_i\}} \mathcal{L}_{\textsc{mstls}}(\lambda)\Big),
    % \quad \text{with} \quad
    % \mathcal{L}_{\textsc{mstls}}(\lambda) := \mathcal{L}\left(\mathbf{w}^{\lambda}; \, \frac{\mathbf{b}_{\textsc{ls}}}{\| \mathbf{b}_{\textsc{ls}} \|_2}, \, \frac{\mathbf{G}}{\| \mathbf{b}_{\textsc{ls}} \|_2}\right)
    % \quad \text{for} \quad
    % \eta = \frac{1}{J},
\end{align} where the loss function $\mathcal{L}_{\textsc{mstls}}$ is defined by \begin{align*}
    \mathcal{L}_{\textsc{mstls}}(\lambda) := \mathcal{L}\left(\mathbf{w}^{\lambda}; \, \frac{\mathbf{b}_{\textsc{ls}}}{\| \mathbf{b}_{\textsc{ls}} \|_2}, \, \frac{\mathbf{G}}{\| \mathbf{b}_{\textsc{ls}} \|_2}\right)
    \quad \text{for} \quad
    \eta = \frac{1}{J}.
\end{align*} In the above expression, $\mathbf{b}_{\textsc{ls}} := \mathbf{G}\mathbf{w}_{\textsc{ls}}$ is the projection of the ordinary least-squares estimate defined by \begin{align*}
    \mathbf{w}_{\textsc{ls}} := (\mathbf{G}^T\mathbf{G})^{-1}\mathbf{G}^T\mathbf{b}.
\end{align*} The $\texttt{MSTLS}$ routine returns the the vector of $\lambda$-thresholded weights, \begin{align*}
    \mathbf{w}^{\lambda} := \texttt{MSTLS}(\mathbf{b}, \mathbf{G}, \lambda),
\end{align*} and is defined as the result of the sequence \begin{align*}
    w^{\lambda}_{n+1} \, = \, \text{arg}\!\!\!\!\!\!\!\!\!\!\!\min_{\text{supp}(\mathbf{w}^{\lambda}_{n}) \subseteq \mathcal{I}_n} \|\mathbf{b} - \mathbf{Gw}\|^2_2,
    % \quad \text{for} \quad
    % \mathcal{I}_{n} := \left\{ 1 \leq j \leq J \, : \, \big(\mathbf{w}^{\lambda}_n\big)_j \in \left[\lambda \max\left(1, \frac{\|\mathbf{b}\|_2}{\|\mathbf{G}_j\|_2}\right), \, \lambda^{-1}\min\left(1, \frac{\|\mathbf{b}\|_2}{\|\mathbf{G}_j\|_2}\right)\right] \right\},
\end{align*} using the stopping criterion $\mathcal{I}_{n+1} = \mathcal{I}_n$, where $\mathcal{I}_n$ is the set of indices defined by \begin{align*}
    \mathcal{I}_{n} := \left\{ 1 \leq j \leq J \, : \, \big(\mathbf{w}^{\lambda}_n\big)_j \in \left[\lambda \max\left(1, \tfrac{\|\mathbf{b}\|_2}{\|\mathbf{G}_j\|_2}\right), \, \lambda^{-1}\min\left(1, \tfrac{\|\mathbf{b}\|_2}{\|\mathbf{G}_j\|_2}\right)\right] \right\},
\end{align*} Note that at each iteration, the MSTLS weights satisfy a dominant balance rule of the form $\|w_j\mathbf{G}_j\|_2 / \|b_j\| \in [\lambda, \lambda^{-1}]$.

\begin{table*}[htb]
    \centering
    \begin{tabular}{||c c c c c||} 
    \hline
    & & & & \\ [-1.75ex]
    \textbf{Model} & $\boldsymbol{\kappa(\mathbf{G})}$ & \textbf{Candidate Terms} & \textbf{Query Points} & \textbf{Time ($\boldsymbol{s}$)} \\ [0.5ex] 
    \hline\hline
    & & & & \\ [-1.75ex]
    Full & $1.6{\texttt{e}}4$ & $84$ & $309,\!600$ & $\sim 130$ \\ [1ex]
    Anisotropic & $2.3$ & $3$ & $309,\!600$ & $<1$ \\ [1ex]
    Effective & 1.0 & $1$ & $309,\!600$ & $<1$ \\ [1ex]
    \hline
    \end{tabular}
    \caption{Supplemental numerical details for each type of model used in this paper. Here, the reported results correspond to the models trained on the combined ensemble dataset (i.e., using all of the available data, $\mathbf{X}_t$, for training) from Tables~\ref{table:full_model}, \ref{table:purely_diffusive_model}, and \ref{table:Deff_model}, respectively. The `$\kappa(\mathbf{G})$' column lists the condition number of the weak library $\mathbf{G}$. The \lq{Time}' column lists the wall time in seconds required to run the MSTLS algorithm on a 2-core Intel Xeon 2.2GHz CPU with 13 GB of RAM.}
    \label{table:numerical_details}
\end{table*}

\newpage

\subsection{Errors in Kernel Density Estimation}\label{sec:error_analysis}
Observational errors, when present, would presumably enter our training data at the level of the experimental position measurements $\mathbf{x}_t = (x_t, y_t) \in \mathbf{X}_t$. To mathematically account for potential errors, we let $\mathbf{x}^{\star}_t \in \mathbf{X}^{\star}_t$ denote the `true' positions and write each measurement as $\mathbf{x}_t = \mathbf{x}^{\star}_t + \boldsymbol{\eta}_t$. In turn, we investigate the pointwise difference between the analogous kernel density estimates, $\varepsilon_h := \hat{u}_h - \hat{u}^{\star}_h$, computed as in eq.~(\ref{eq:gaussian}) but with a Gaussian kernel $G(\boldsymbol{x}; \mathbf{C}_h)$ defined by a fixed (i.e., sample-independent) covariance matrix $\mathbf{C}_h := h^2\mathbf{C}$, \begin{align*}
    \varepsilon(\boldsymbol{x},t; \mathbf{C}_h) 
    = \frac{1}{N_t} \sum_{i=1}^{N_t} \Big[G\big(\boldsymbol{x} - \mathbf{x}^i_t; \mathbf{C}_h\big) - G_h\big(\boldsymbol{x} - (\mathbf{x}^{\star})^{i}_t; \mathbf{C}_h\big) \Big].
\end{align*} We claim that no obvious systematic measurement errors were made during the experiment and instead suggest that the most appropriate error model comes in the form of normally-distributed and unbiased random noise, $\boldsymbol{\eta}_t \sim \mathcal{N}\big(0, \sigma^2\mathbf{I}\big)$. For a \textit{fixed} set of true positions $\mathbf{X}^{\star}_t$, the assumption of normality implies that $\mathbf{x}_t|_{\mathbf{x}^{\star}_t} \sim \mathcal{N}\big(\mathbf{x}^{\star}_t, \sigma^2\mathbf{I}\big)$, which in turn yields a conditional expectation $E_h := \mathbb{E}[\varepsilon_h \, | \, \mathbf{X}^{\star}_t]$ given by \begin{align}\label{eq:expected_KDE_error}
    E(\boldsymbol{x},t; \mathbf{C}_h)
    &= \frac{1}{N_t} {\sum}' \Big[ G\big(\boldsymbol{x} - \mathbf{x}^{\star}_t; \mathbf{C}_h + \sigma^2\mathbf{I}\big) -  G\big(\boldsymbol{x} - \mathbf{x}^{\star}_t; \mathbf{C}_h\big) \Big],
\end{align} where $\sum'$ denotes a sum over each position $\mathbf{x}^{\star}_t \in \mathbf{X}^{\star}_t$. If the standard deviation $\sigma$ of the noise term $\boldsymbol{\eta}_t$ is small in comparison to the bandwidth $h$ of the Gaussian kernel (i.e., $\sigma/h \ll 1$), then it becomes natural to expand eq.~(\ref{eq:expected_KDE_error}) via \begin{align*}
    G\big(\boldsymbol{y}; \, \mathbf{C}_h + \epsilon\mathbf{I}\big)
    -
    G\big(\boldsymbol{y}; \mathbf{C}_h\big)
    \, &= \,
    \epsilon\left[\frac{\partial}{\partial\epsilon} G\big(\boldsymbol{y}; \, \mathbf{C}_h + \epsilon\mathbf{I}\big) \, \Big|_{\epsilon=0}\right]
    + \mathcal{O}\big(\epsilon^2\big)
    %\\
    % &= \,
    % G\big(\boldsymbol{y}; \mathbf{C}_h\big)
    % +
    % \frac{\epsilon}{2} \left[ \boldsymbol{y}^T\mathbf{C}_h^{-2}\boldsymbol{y} - {\rm{tr}}\big(\mathbf{C}_h^{-1}\big) \right] G(\boldsymbol{y};\mathbf{C}_h)
    % +
    % \mathcal{O}\big(\epsilon^2\big),
    \\
    &= \,
    \epsilon \, (\Delta_{\boldsymbol{y}}G)(\boldsymbol{y};\mathbf{C}_h)
    + \mathcal{O}\big(\epsilon^2\big),
\end{align*} which can be substituted into eq.~(\ref{eq:expected_KDE_error}) and simplified to yield a leading-order approximation in the form of a convolution of $\mu(\mathbf{X}^{\star}_t)$ against a `Laplacian of Gaussian' (\textit{LoG}) filter: \begin{align}\label{eq:expected_KDE_error_approx}
    % {\color{red}E_h}(\boldsymbol{x},t; \mathbf{C}) = \frac{\sigma^2}{2h^2 |\mathbf{C}|} \left[ \boldsymbol{x}^T\!\mathbf{A}^2_h\boldsymbol{x} - {\rm{tr}}\big(\mathbf{C}\big) \right] \hat{u}^{\star}_h(\boldsymbol{x},t)
    % + \mathcal{O}\big(\sigma^4\big),
    % \quad \text{where} \quad
    % \mathbf{A}_{h} := \frac{\text{adj}\big(\mathbf{C}\big)}{h \sqrt{|\mathbf{C}|} }.
    E(\boldsymbol{x},t; \mathbf{C}_h)
    &= \frac{\sigma^2}{N_t} {\sum}' (\Delta{G})\big(\boldsymbol{x} - \mathbf{x}^{\star}_t; \mathbf{C}_h\big)
    + \mathcal{O}\big(\sigma^4\big).
\end{align}

\newpage

The approximation given in eq.~(\ref{eq:expected_KDE_error_approx}) above represents the influence of measurement noise $\boldsymbol{\eta}_t$ on the density estimation process (i.e., for a given $\mathbf{C}_h$). With this in mind, we note that it is also possible to estimate the error resulting from a finite number of samples. Assuming that $\mathbf{X}^{\star}_t$ represents $N_t$ samples drawn from an underlying distribution $\mathbf{x}^{\star}_t \sim u^{\star}(\boldsymbol{x},t)$, the density estimate $\hat{u}^{\star}_h$ is known to converge in probability to $u^{\star}$ in the limit of infinite data (i.e., as $N_t \rightarrow \infty$).\footnote{That is, under certain assumptions on the kernel $G_h$, the Gaussian kernel density estimate $\hat{u}^{\star}_h$ is an asymptotically-unbiased estimator of $u^{\star}$.} For a finite number of samples, the expected value of the induced $L^2$ truncation error is given at a time $t$ by \begin{align*}
    \mathbb{E}\left[ \big|\!\big|\big(u^{\star} - \hat{u}^{\star}_h\big)(\cdot,t) \big|\!\big|_2^2\right]
    =
    \frac{1}{4\pi h N_t |\mathbf{C}|^{\frac{1}{2}} } + H(\boldsymbol{x};t) + o\left(\frac{1}{N_t |\mathbf{C}|^{\frac{1}{2}} } + {\rm{tr}}\big(\mathbf{C}^2\big)\!\right),
\end{align*} where the $H$-term in the above expression is given explicitly by \begin{align*}
    H(\boldsymbol{x};t)
    :=
    {\rm{vec}}(\mathbf{C})^T
    \left[\, \frac{1}{4} \int\!\!\!\!\int_{\Omega} {\rm{vec}}\!\left(\nabla\nabla^Tu^{\star}(\boldsymbol{x},t)\right) \, {\rm{vec}}\!\left(\nabla\nabla^Tu^{\star}(\boldsymbol{x},t)\right)^{\!T} \! dx \, dy \,\right]
    {\rm{vec}}(\mathbf{C}),
\end{align*} with $\nabla\nabla^Tu^{\star}$ denoting a Hessian matrix taken with respect to space. %{\color{red}(Reference for this?)}

%\newpage

\subsection{Standard Error in Parameter Estimates}\label{sec:residuals_and_parameter_error}
Here, we derive an approximation of the parameter covariance matrix $\hat{\mathbf{S}} := \text{var}(\hat{\mathbf{w}} - \mathbf{w}^{\star})$. We begin by assuming that our model specification is correct; i.e., suppose that a vector of coefficients $\mathbf{w}^{\star}$ exists such that for error-less data $\mathbf{u}^{\star}$\!, we have the weak-form equality \begin{align*}
    \mathbf{G}^{\star}\mathbf{w}^{\star} - \mathbf{b}^{\star} = \mathbf{r}_{\rm{int}},
\end{align*} where $\|\mathbf{r}_{\rm{int}}\|_{\infty} = \mathcal{O}\big(\Delta x^{p+1}\big)$ represents the truncation error induced by numerical quadrature.

\newpage

Under the introduction of a perturbation $\mathbf{u} = \mathbf{u}^{\star} + \boldsymbol{\epsilon}$, leading to analogous perturbations $\mathbf{G}=\mathbf{G}^{\star}+\mathbf{G}^{\epsilon}$ and $\mathbf{b}=\mathbf{b}^{\star}+\mathbf{b}^{\epsilon}$, we follow an analysis similar to that of \cite{BortzMessengerDukic2023BullMathBiol} to obtain \begin{align}\label{eq:residual}
    \mathbf{r}(\mathbf{u}, \mathbf{w})
    \, &:= \, \mathbf{G}(\mathbf{u})\mathbf{w} - \mathbf{b}(\mathbf{u}) \nonumber
    % \, = \, \big(\mathbf{G}^{\epsilon}(\mathbf{u})\mathbf{w} - \mathbf{b}^{\epsilon}(\mathbf{u})\big) + \mathbf{G}^{\star}\big(\mathbf{w} - \mathbf{w}^{\star}\big) + \mathbf{r}_{\rm{int}}.
    \\
    &= \, \big(\mathbf{G}^{\epsilon}(\mathbf{u})\mathbf{w}^{\star} - \mathbf{b}^{\epsilon}(\mathbf{u})\big) + \mathbf{G}(\mathbf{u})\big(\mathbf{w} - \mathbf{w}^{\star}\big) + \mathbf{r}_{\rm{int}}.
\end{align} In the absence of 'noise` and parameter error, the residual in eq.~(\ref{eq:residual}) collapses to $\mathbf{r}(\mathbf{u}^{\star}\!,\mathbf{w}^{\star}) = \mathbf{r}_{\rm{int}}$. With this expansion in mind, we note that the true weights satisfy $\mathbf{b} = \mathbf{G}\mathbf{w}^{\star} \! - \mathbf{r}(\mathbf{u},\mathbf{w}^{\star})$, which means that we can in turn express the ordinary least-squares parameter estimates $\hat{\mathbf{w}}$ to the tune of \begin{align*}%\label{eq:what_minus_wstar}
    \hat{\mathbf{w}}(\mathbf{u})
    :=
    \mathbf{G}^{\dagger}(\mathbf{u})\mathbf{b}(\mathbf{u})
    =
    \mathbf{G}^{\dagger}(\mathbf{u})\left(\mathbf{G}(\mathbf{u})\mathbf{w}^{\star} \! - \mathbf{r}(\mathbf{u},\mathbf{w}^{\star})\right),
    % \quad \text{so that} \quad
    % \hat{\mathbf{w}}(\mathbf{u}) - \mathbf{w}^{\star}
    % =
    % - \mathbf{G}^{\dagger}(\mathbf{u})\mathbf{r}(\mathbf{u},\mathbf{w}^{\star}),
\end{align*} so that \begin{align}\label{eq:what_minus_wstar}
    \hat{\mathbf{w}}(\mathbf{u}) - \mathbf{w}^{\star}
    =
    - \mathbf{G}^{\dagger}(\mathbf{u})\mathbf{r}(\mathbf{u},\mathbf{w}^{\star}),
\end{align} where $\mathbf{G}^{\dagger} = \big(\mathbf{G}^T\mathbf{G}\big)^{-1} \mathbf{G}^T$ denotes the left pseudo-inverse of $\mathbf{G}$. To simplify this expression, we note that a Taylor series expansion of $\mathbf{r}(\mathbf{u},\mathbf{w}^{\star})$ and $\mathbf{G}^{\dagger}(\mathbf{u})$ around the error-less data,\footnote{Here, we have computed the series expansions in terms of Fr\'echet derivatives of the form $\mathbf{L}_{\mathbf{f}}(\boldsymbol{\xi}, \dots) := \big(\nabla_{\mathbf{u}}^T \otimes \mathbf{f}\big)(\boldsymbol{\xi},\dots)$, where $\otimes$ denotes the Kronecker product.} \begin{align*}
    \begin{cases}
        \mathbf{r}(\mathbf{u}^{\star}+\boldsymbol{\epsilon},\mathbf{w}^{\star})
        \, = \,
        \mathbf{r}_{\rm{int}}
        +
        \mathbf{L}_{\mathbf{r}}(\mathbf{u}^{\star}\!,\mathbf{w}^{\star}) \, \boldsymbol{\epsilon}
        + \mathcal{O}\big(|\boldsymbol{\epsilon}|^2\big),
        \\
        \mathbf{G}^{\dagger}(\mathbf{u}^{\star} + \boldsymbol{\epsilon})
        \, = \,
        (\mathbf{G}^{\star})^{\dagger}
        +
        \mathbf{L}_{\mathbf{G}^{\dagger}}(\mathbf{u}^{\star})\big(\boldsymbol{\epsilon} \otimes \mathbf{I}\big)
        + \mathcal{O}\big(|\boldsymbol{\epsilon}|^2\big),
    \end{cases}
\end{align*} can be substituted into eq.~(\ref{eq:what_minus_wstar}) to yield a helpful leading-order approximation, which, under the additional assumptions that the integration error is negligible (i.e., $\mathbf{r}_{\rm{int}} \approx 0$) and the perturbation is unbiased (i.e., $\mathbb{E}[\,\boldsymbol{\epsilon}\,] = 0$), takes the form \begin{align*}
    \hat{\mathbf{w}} - \mathbf{w}^{\star}
    \approx
    - (\mathbf{G}^{\star})^{\dagger} \mathbf{L}_{\mathbf{r}}(\mathbf{u}^{\star}\!,\mathbf{w}^{\star}) \, \boldsymbol{\epsilon},
    % \quad \text{so that} \quad
    % \mathbb{E}\!\left[\hat{\mathbf{w}} - \mathbf{w}^{\star}\right]
    % \approx
    % - (\mathbf{G}^{\star})^{\dagger} \mathbf{L}_{\mathbf{r}}(\mathbf{u}^{\star}\!,\mathbf{w}^{\star}) \, \mathbb{E}[\,\boldsymbol{\epsilon}\,] = 0.
\end{align*} so that \begin{align*}
    \mathbb{E}\!\left[\hat{\mathbf{w}} - \mathbf{w}^{\star}\right]
    \approx
    - (\mathbf{G}^{\star})^{\dagger} \mathbf{L}_{\mathbf{r}}(\mathbf{u}^{\star}\!,\mathbf{w}^{\star}) \, \mathbb{E}[\,\boldsymbol{\epsilon}\,] = 0.
\end{align*} To leading order in $\boldsymbol{\epsilon}$, the parameter covariance matrix $\mathbf{S} := \text{var}(\hat{\mathbf{w}} - \mathbf{w}^{\star})$ is thus given by \begin{align*}
    \mathbf{S}
    \approx
    \mathbb{E}\!\left[ (\hat{\mathbf{w}} - \mathbf{w}^{\star}) (\hat{\mathbf{w}} - \mathbf{w}^{\star})^T \right]
    \approx
    \left[
    \mathbf{G}^{\dagger}
    \mathbf{L}_{\mathbf{r}}
    \, \mathbb{E}\left[\boldsymbol{\epsilon} \otimes \boldsymbol{\epsilon}\right]
    \mathbf{L}_{\mathbf{r}}^T
    \big(\mathbf{G}^{\dagger}\big)^T
    \right]\!(\mathbf{u}^{\star}\!, \mathbf{w}^{\star}).
\end{align*} To obtain numerical practical estimates $\hat{\sigma}(w_{j})$ of the standard errors $\sigma(w_j) = \sqrt{\mathbf{S}_{jj}}$, we follow \cite{White1980Econometrica} in computing \begin{align}\label{eq:standard_errors}
    \hat{\sigma}(w_{j}) = 
    \sqrt{\hat{\mathbf{S}}_{jj}}, 
    \quad \text{where} \quad
    \hat{\mathbf{S}}
    := \left[
    \mathbf{G}^{\dagger}
    \, \text{diag}\big(r_1^2, \dots, r_{\kappa}^2\big)
    \big(\mathbf{G}^{\dagger})^T
    \right]\!(\mathbf{u}, \hat{\mathbf{w}}),
\end{align} which is based on the estimate $\mathbf{r} \approx \mathbf{L}_{\mathbf{r}} \boldsymbol{\epsilon}$ and uses a sample mean for the resulting expectation $\mathbb{E}[(\mathbf{L}_{\mathbf{r}}\boldsymbol{\epsilon}) \otimes (\mathbf{L}_{\mathbf{r}}\boldsymbol{\epsilon})] \approx \mathbb{E}[\mathbf{r} \otimes \mathbf{r}]$.

\newpage

\begin{landscape}
\begin{table*}
    \centering
    \begin{tabular}{||c c c c c c c c||}
     \hline
     & & & & & & & \\[-2.0 ex]
     \textbf{Run, $\boldsymbol{k}$} & \textbf{Plant} & \textbf{Virus} & \ $\boldsymbol{V_c \pm 2\hat{\sigma}}$ & \ \ $\boldsymbol{K_c \pm 2\hat{\sigma}}$ & \ \ $\boldsymbol{[D_{x}, \, D_{xy}, \, D_{y}] \pm 2\hat{\sigma}}$ & $\boldsymbol{R^2}$ & $\boldsymbol{\Delta{\rm{AIC}}}$ \\ [0.5ex]
     \hline
     \hline
     & & & & & & & \\[-1.5 ex]
     1 & Stonewall & No &
     \ \ \ $\boldsymbol{0.7} \, | \, 20.1$ & \ \ \ $\boldsymbol{0.0} \, | \, 0.7$ & \ \ \ $\boldsymbol{[3.4, 3.4, 4.4]} \, | \, [2.9, 3.2, 3.4]$ & $\boldsymbol{0.13} \, | \, 0.15$ & -157.5 \\[0.5 ex]
     & & & ${\color{gray}\boldsymbol{\pm 0.1} \, | \, 16.8}$ & ${\color{gray}\boldsymbol{\pm 0.0} \, | \, 2.4}$ & ${\color{gray}\boldsymbol{\pm [0.3, 0.5, 0.3]} \, | \, [0.4, 0.6, 0.3]}$ & & \\[0.5 ex]
     %\hline
     1 & Gasoy & No &
     \ \ \ $\boldsymbol{0.9} \, | \, 1.1$ & \ \ \ $\boldsymbol{0.0} \, | \, 6.8$ & \ \ \ \ $\boldsymbol{[3.3, 0.0, 7.3]} \, | \, [4.4, \text{-}0.2, \text{-}2.2]$ & $\boldsymbol{0.29} \, | \, 0.36$ & -149.0 \\[0.5 ex]
     & & & ${\color{gray}\boldsymbol{\pm 0.1} \, | \, 1.1}$ & ${\color{gray}\boldsymbol{\pm 0.0} \, | \, 3.6}$ & ${\color{gray}\boldsymbol{\pm [0.2, 0.4, 0.4]} \, | \, [0.4, 0.6, 0.6]}$ & & \\[0.5 ex]
     %\hline
     1 & Stonewall & Yes &
     \ \ \ $\boldsymbol{1.0} \, | \, 9.9$ & \ \ \ $\boldsymbol{0.1} \, | \, 1.9$ & \ \ \ $\boldsymbol{[12.0, 0.0, 7.2]} \, | \, [13.6, \text{-}1.8, 3.7]$ & $\boldsymbol{0.50} \, | \, 0.58$ & -143.8 \\[0.5 ex]
     & & & ${\color{gray}\boldsymbol{\pm 0.0} \, | \, 3.9}$ & ${\color{gray}\boldsymbol{\pm 0.0} \, | \, 6.5}$ & ${\color{gray}\boldsymbol{\pm [0.8, 0.2, 0.7]} \, | \, [1.0, 1.2, 0.6]}$ & & \\[0.5 ex]
     %\hline
     1 & Gasoy & Yes &
     \ \ \ $\boldsymbol{1.9} \, | \, 4.3$ & \ \ \ $\boldsymbol{0.0} \, | \, 2.2$ & \ \ \ $\boldsymbol{[7.6, \text{-}5.2, 6.3]} \, | \, [2.9, \text{-}5.2, 5.9]$ & $\boldsymbol{0.12} \, | \, 0.16$ & -151.5 \\[0.5 ex]
     & & & ${\color{gray}\boldsymbol{\pm 0.1} \, | \, 2.2}$ & ${\color{gray}\boldsymbol{\pm 0.0} \, | \, 6.7}$ & ${\color{gray}\boldsymbol{\pm [0.6, 1.3, 0.6]} \, | \, [0.9, 1.4, 1.0]}$ & & \\[0.75 ex]
     \hline
     & & & & & & & \\[-1.5 ex]
     2 & Stonewall & No &
     \ \ \ $\boldsymbol{2.4} \, | \, 1.7$ & \ \ \ $\boldsymbol{0.0} \, | \, 0.5$ & \ \ \ \ $\boldsymbol{[0.0, 0.0, 8.2]} \, | \, [0.3, \text{-}0.4, 9.2]$ & $\boldsymbol{0.36} \, | \, 0.38$ & -153.0 \\[0.5 ex]
     & & & ${\color{gray}\boldsymbol{\pm 0.1} \, | \, 1.8}$ & ${\color{gray}\boldsymbol{\pm 0.0} \, | \, 3.9}$ & ${\color{gray}\boldsymbol{\pm [0.6, 0.3, 1.0]} \, | \, [0.9, 0.7, 0.6]}$ & & \\[0.5 ex]
     %\hline
     2 & Gasoy & No &
     \ \ \ $\boldsymbol{2.6} \, | \, 18.4$ & \ \ \ $\boldsymbol{0.0} \, | \, 3.2$ & \ \ \ $\boldsymbol{[5.7, \text{-}2.2, 3.4]} \, | \, [8.0, \text{-}4.1, 4.6]$ & $\boldsymbol{0.43} \, | \, 0.53$ & -143.7 \\[0.5 ex]
     & & & ${\color{gray}\boldsymbol{\pm 0.1} \, | \, 1.8}$ & ${\color{gray}\boldsymbol{\pm 0.0} \, | \, 5.2}$ & ${\color{gray}\boldsymbol{\pm [0.7, 0.6, 0.3]} \, | \, [0.7, 0.5, 0.3]}$ & & \\[0.5 ex]
     %\hline
     2 & Stonewall & Yes &
     \ \ \ $\boldsymbol{1.1} \, | \, 2.1$ & \ \ \ $\boldsymbol{0.0} \, | \, 3.3$ & \ \ \ \ $\boldsymbol{[2.8, 0.0, 6.6]} \, | \, [4.7, \text{-}1.2, 6.0]$ & $\boldsymbol{0.27} \, | \, 0.31$ & -159.4 \\[0.5 ex]
     & & & ${\color{gray}\boldsymbol{\pm 0.0} \, | \, 1.9}$ & ${\color{gray}\boldsymbol{\pm 0.0} \, | \, 5.8}$ & ${\color{gray}\boldsymbol{\pm [0.4, 0.2, 0.5]} \, | \, [0.6, 0.5, 0.4]}$ & & \\[0.5 ex]
     %\hline
     2 & Gasoy & Yes &
     \ \ \ $\boldsymbol{1.6} \, | \, 2.4$ & \ \ \ $\boldsymbol{0.0} \, | \, 3.4$ & \ \ \ \ $\boldsymbol{[5.7, 0.0, 4.7]} \, | \, [7.6, \text{-}0.6, 3.2]$ & $\boldsymbol{0.44} \, | \, 0.49$ & -152.3 \\[0.5 ex]
     & & & ${\color{gray}\boldsymbol{\pm 0.0} \, | \, 1.0}$ & ${\color{gray}\boldsymbol{\pm 0.0} \, | \, 4.0}$ & ${\color{gray}\boldsymbol{\pm [0.4, 0.3, 0.3]} \, | \, [0.6, 0.6, 0.4]}$ & & \\[0.75 ex]
     \hline
    \end{tabular}
    \caption{Supplemental model discovery results for the control populations listed in Table~\ref{table:full_model}, here separated by `run number' (i.e., the unique ID referring to one of the two possible experiment dates for each case). By grouping the data according to their actual experiment date (instead of a synthetically-combined, ensemble dataset), we select for populations that were distributed across the same planter at the same times and thus had the opportunity to physically interact. This allows us to estimate the corresponding interaction potentials $\mathcal{K}(\boldsymbol{x}, \boldsymbol{x}')$, although this separation of the training data comes at the cost of inducing large variances due to small sample counts.}
    \label{table:results_by_run_number}
\end{table*}
\end{landscape}

\newpage

\begin{landscape}
\begin{figure*}
    \centering
    \includegraphics[width=\linewidth]{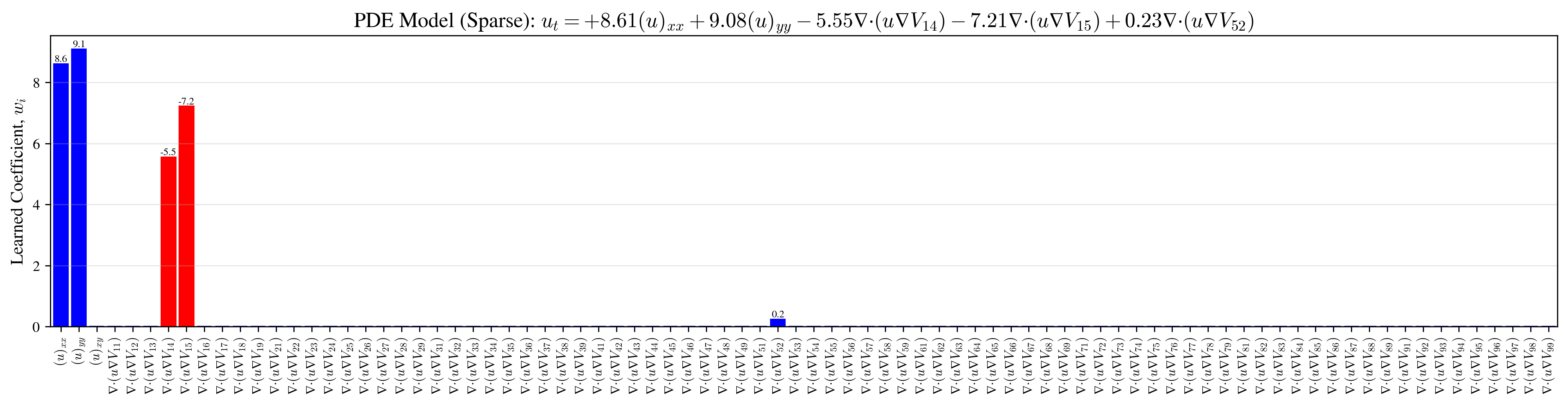}
    \includegraphics[width=\linewidth]{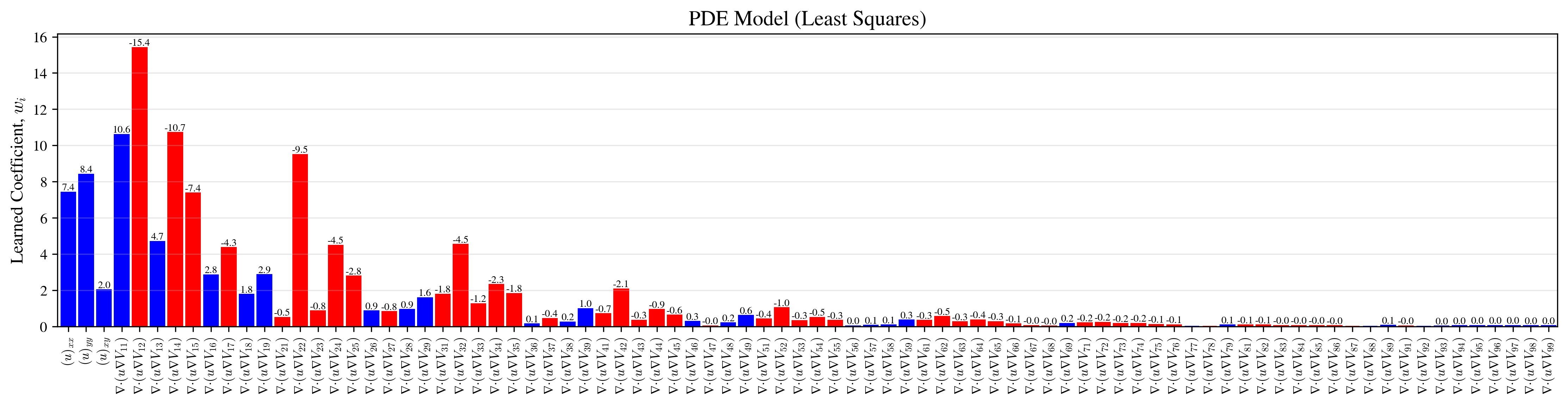}
    \caption{Illustrating the relative magnitudes of the learned sparse and least-squares models weights, respectively, for the combined training data $\mathbf{X}_t$ of Table~\ref{table:full_model}.}
\end{figure*}
\end{landscape}

\newpage

\begin{figure*}
    \centering
    \includegraphics[width=0.98\linewidth]{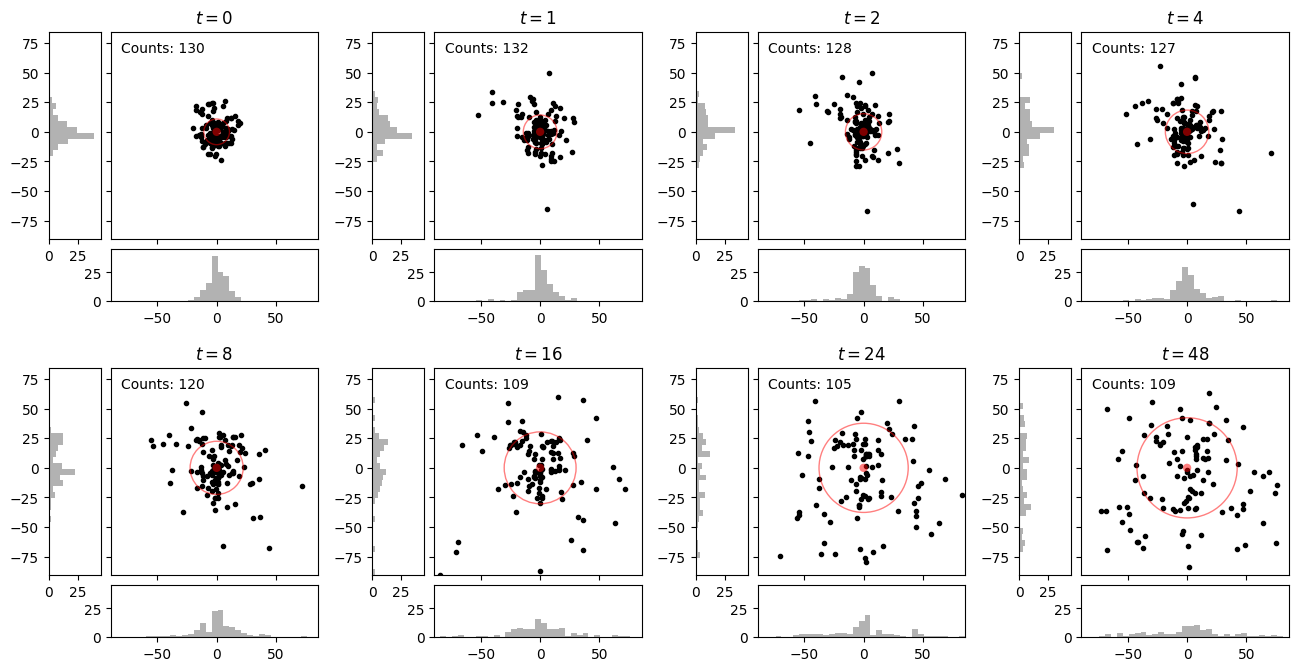}
    \includegraphics[width=0.98\linewidth]{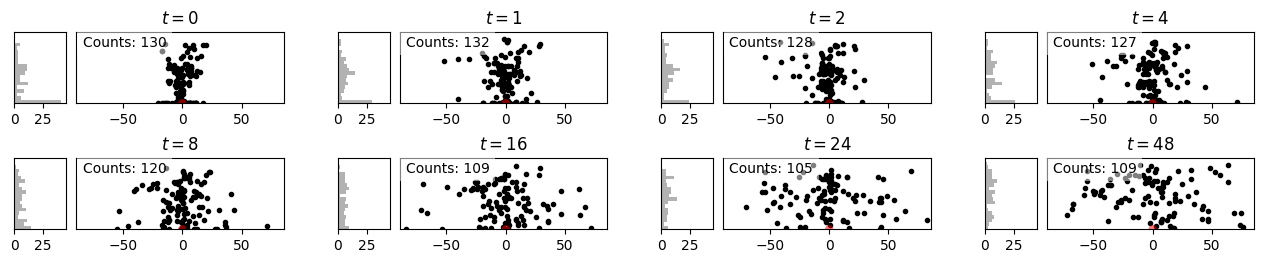}
    \includegraphics[width=0.98\linewidth]{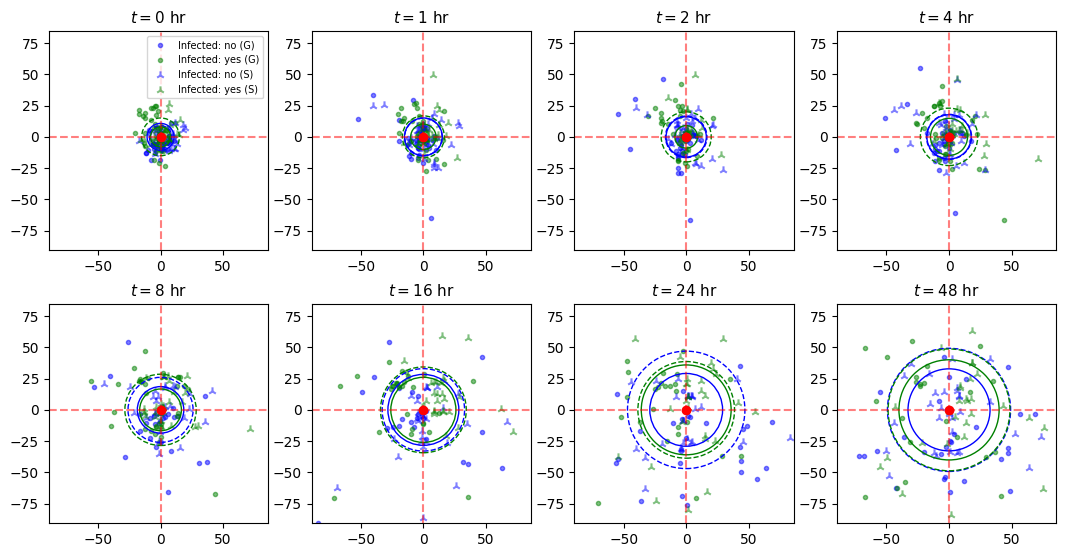}
    \includegraphics[width=0.98\linewidth]{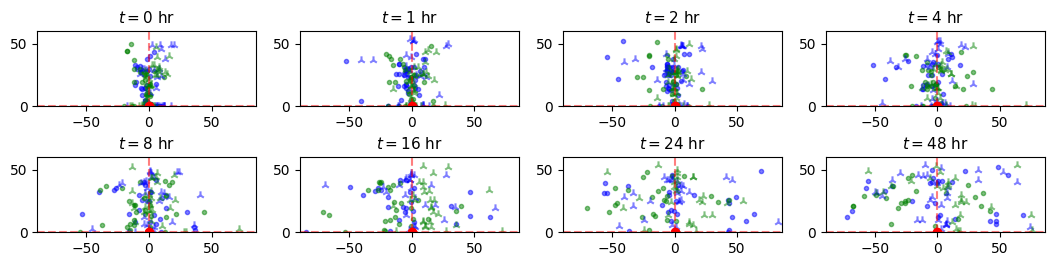}
    \caption{(Top panels) Individual positions projected into the $(y,x)$-plane at each snapshot $t_n$. The displacement radius $\langle \rho \rangle = \langle |\mathbf{x} - \langle{\mathbf{X}}_0\rangle| \rangle$ is also shown in red, where $\langle{\mathbf{X}}_0\rangle$ denotes the initial center of mass. (Bottom panels) Plotting the corresponding $z$-displacements evolving over time in the $(x,z)$ plane.}
    \label{fig:HISTOGRAMS}
\end{figure*}

\newpage

\begin{figure*}
    \centering
    \includegraphics[width=0.98\linewidth]{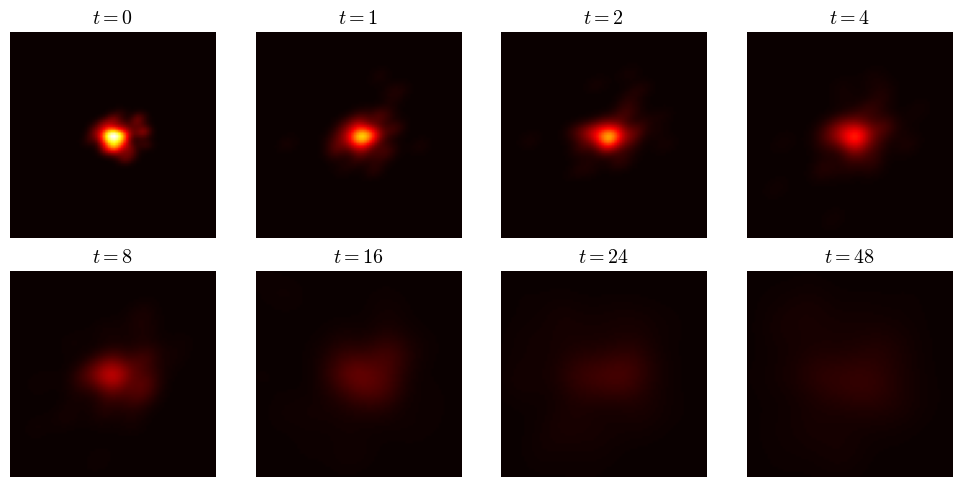}
    \includegraphics[width=0.98\linewidth]{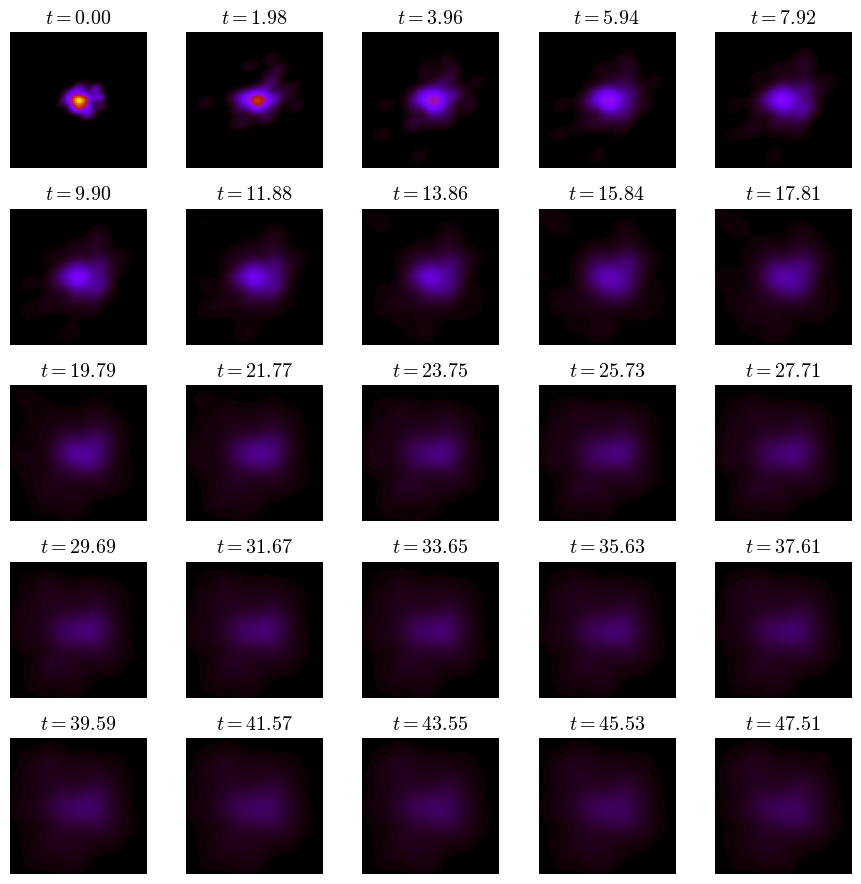}
    \caption{Illustrating the temporal interpolation between snapshots $t_n$. (Top) The density estimate $\hat{u}_h(\boldsymbol{x},t)$ obtained with the combined data $\mathbf{X}_t$ at the original snapshots $t_n \in \{t_0, \dots, t_{\textsc{f}}\}$. (Bottom) The interpolated density evolving in time.}
    \label{fig:temporal_interpolation}
\end{figure*}

\newpage

\begin{figure*}
    \centering
    \includegraphics[width=\linewidth]{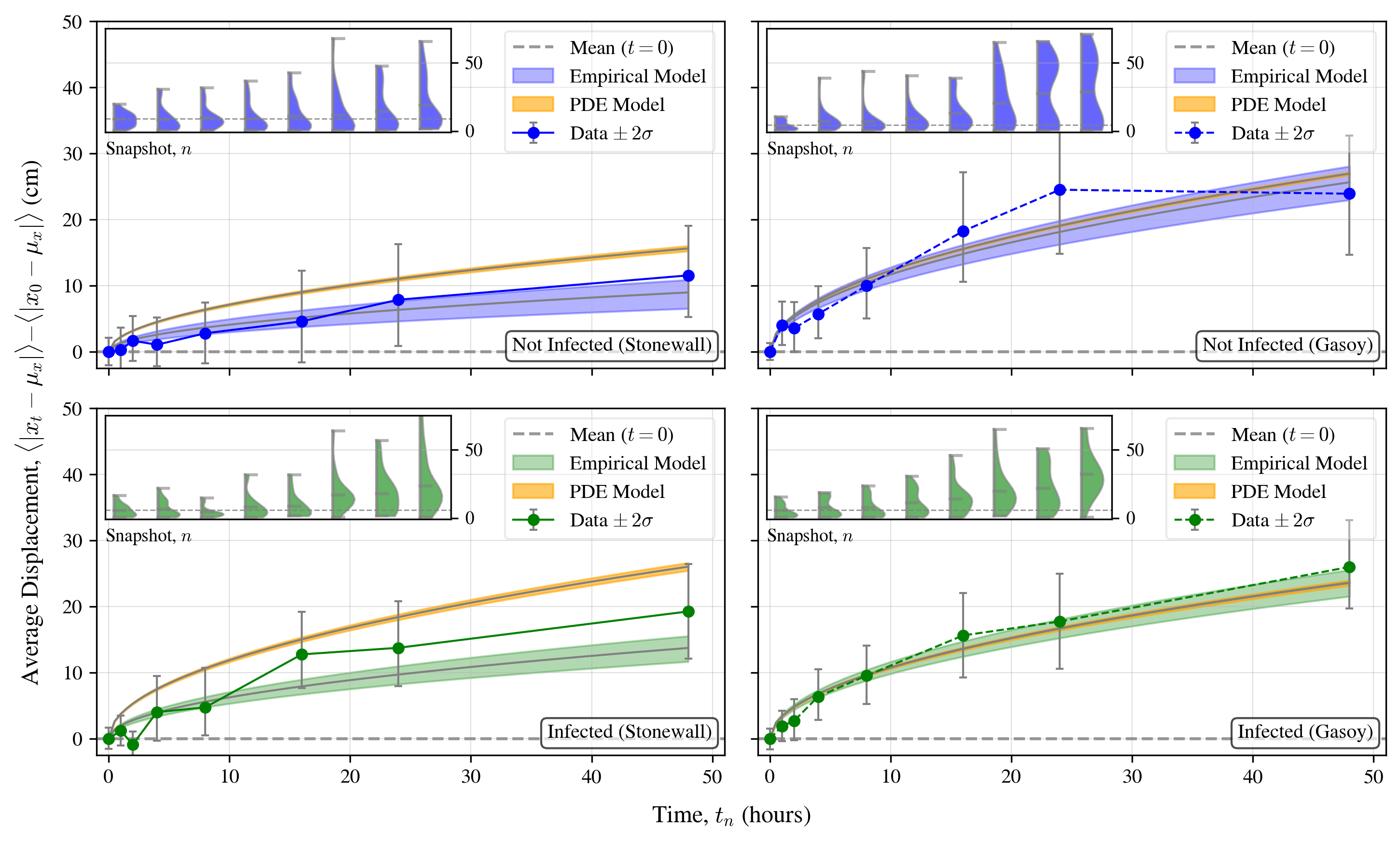}
    \\
    \bigskip
    \includegraphics[width=\linewidth]{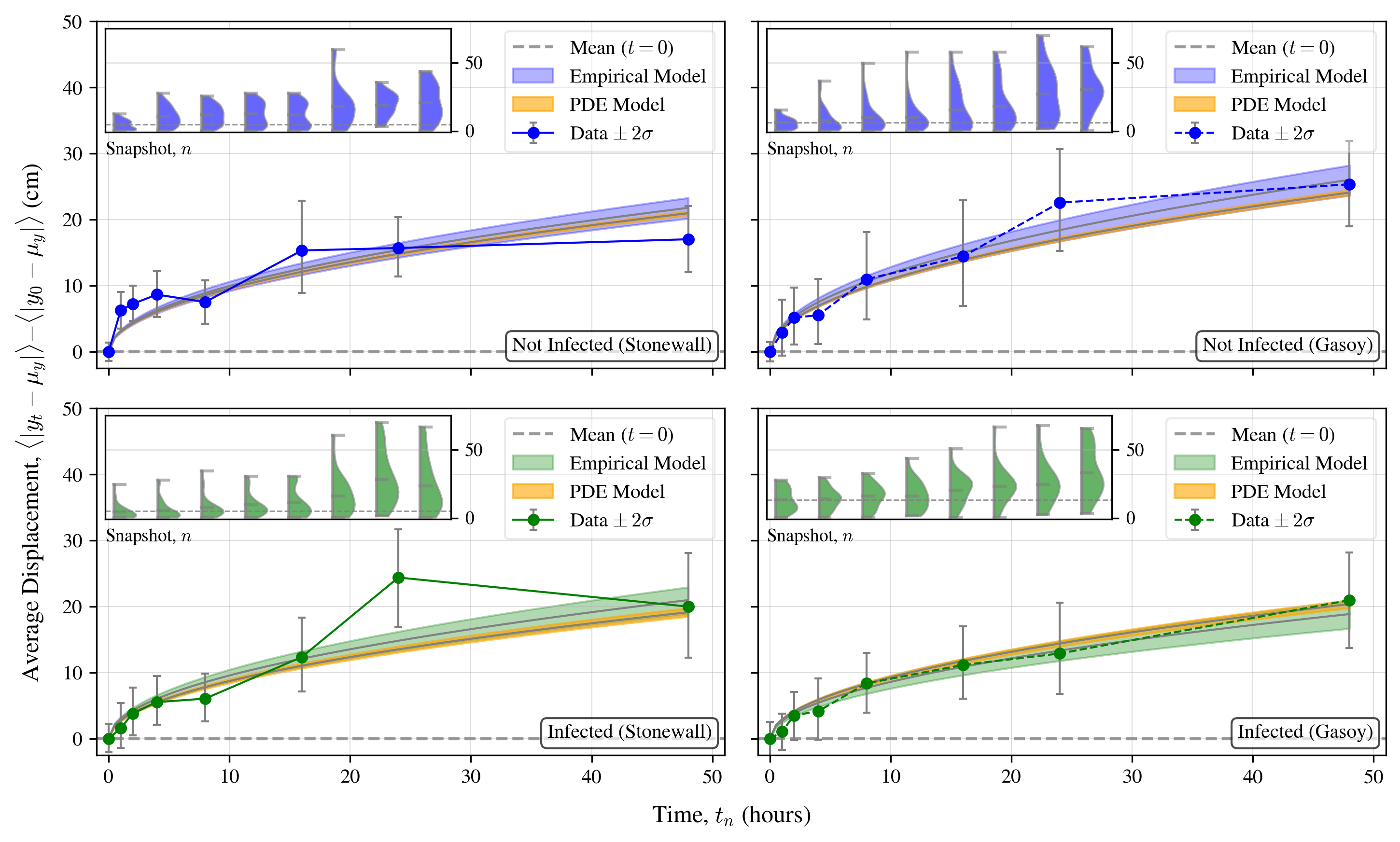}
    \caption{Similar to Figure~\ref{fig:caterpillars_violin_r} except using the $x$ and $y$-axes, respectively, instead of the radial displacement $\rho$. For the weak-form model, we plot $|x_i(t)-\mu_x| = \sqrt{(4/\pi)(D_{ii} \pm 2\hat{\sigma})t}$.}
    \label{fig:cats_violin_xy}
\end{figure*}

\newpage

\begin{figure*}
    \centering
    \includegraphics[width=\linewidth]{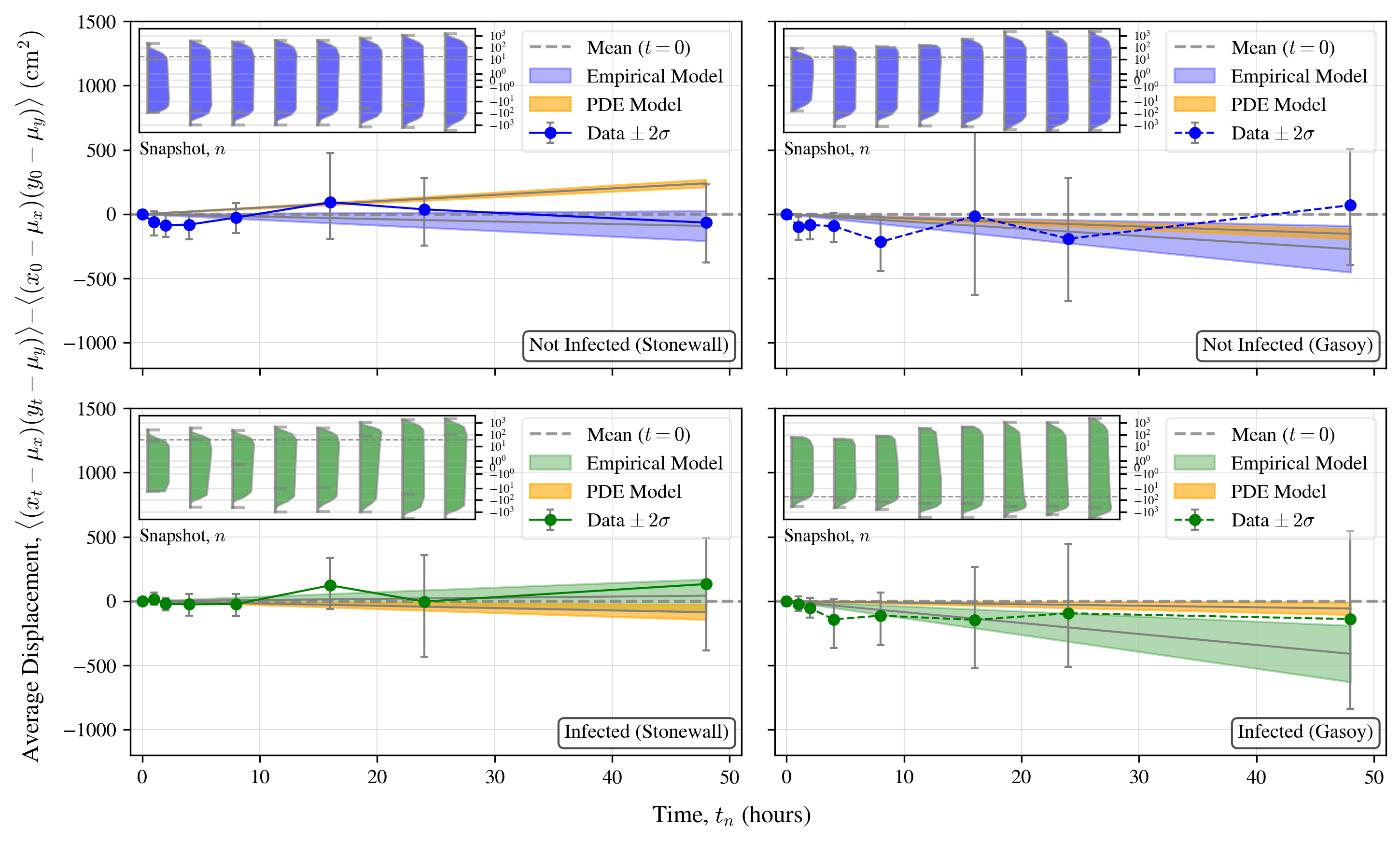}
    \caption{Similar to Figure~\ref{fig:caterpillars_violin_r} and Figure~\ref{fig:cats_violin_xy} except that here we use $\langle(x-\mu_x)(y-\mu_y)\rangle$ to estimate the $\hat{D}_{xy}$ cross-terms.}
    \label{fig:cats_violin_cross_term}
\end{figure*}

\begin{figure*}
    \centering
    \includegraphics[width=\linewidth]{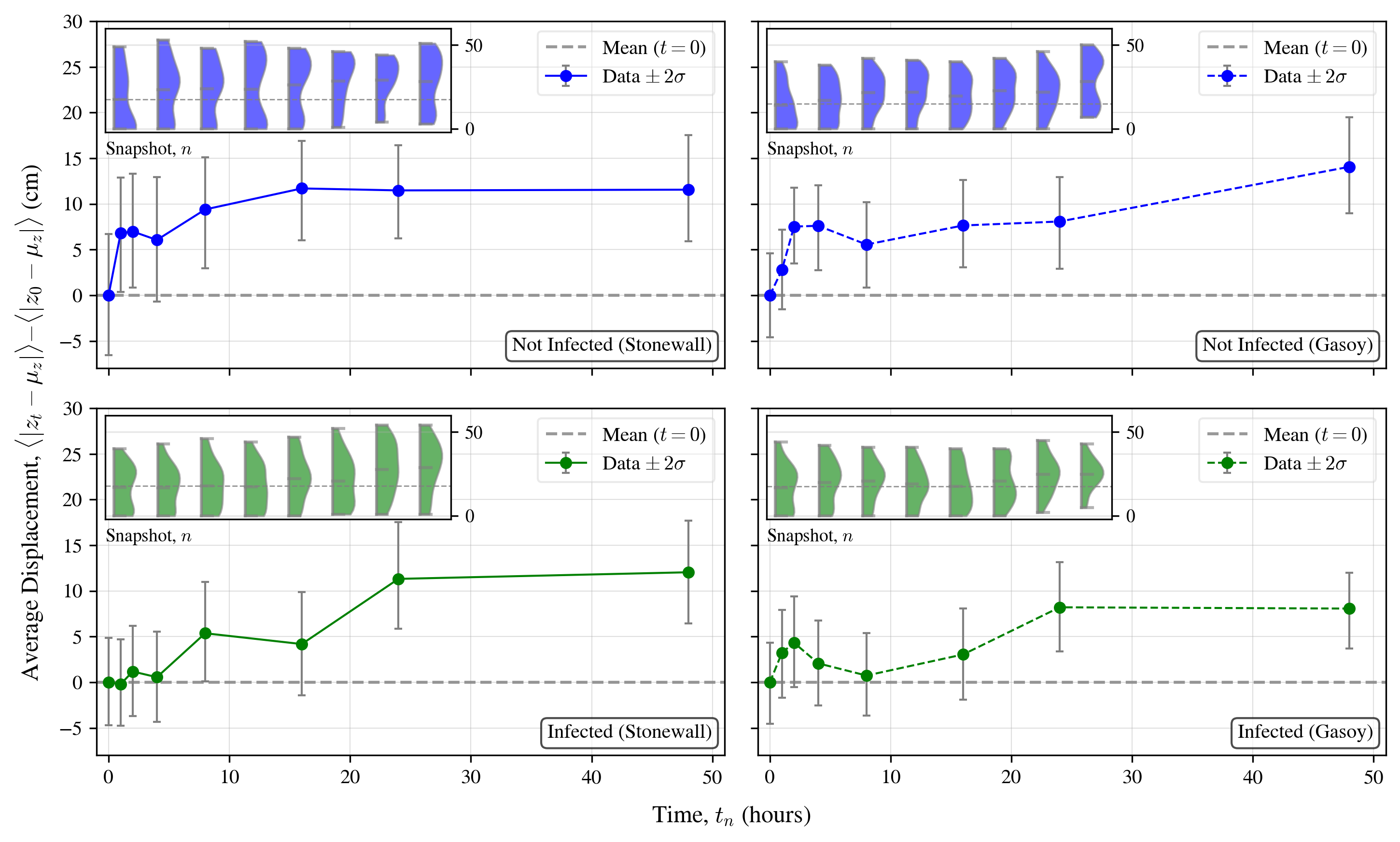}
    \caption{Similar to Figure~\ref{fig:caterpillars_violin_r} and Figure~\ref{fig:cats_violin_xy}, except that here we plot averaged vertical displacements.}
    \label{fig:cats_violin_z}
\end{figure*}

\newpage

\begin{figure*}
    \centering
    \includegraphics[width=\linewidth]{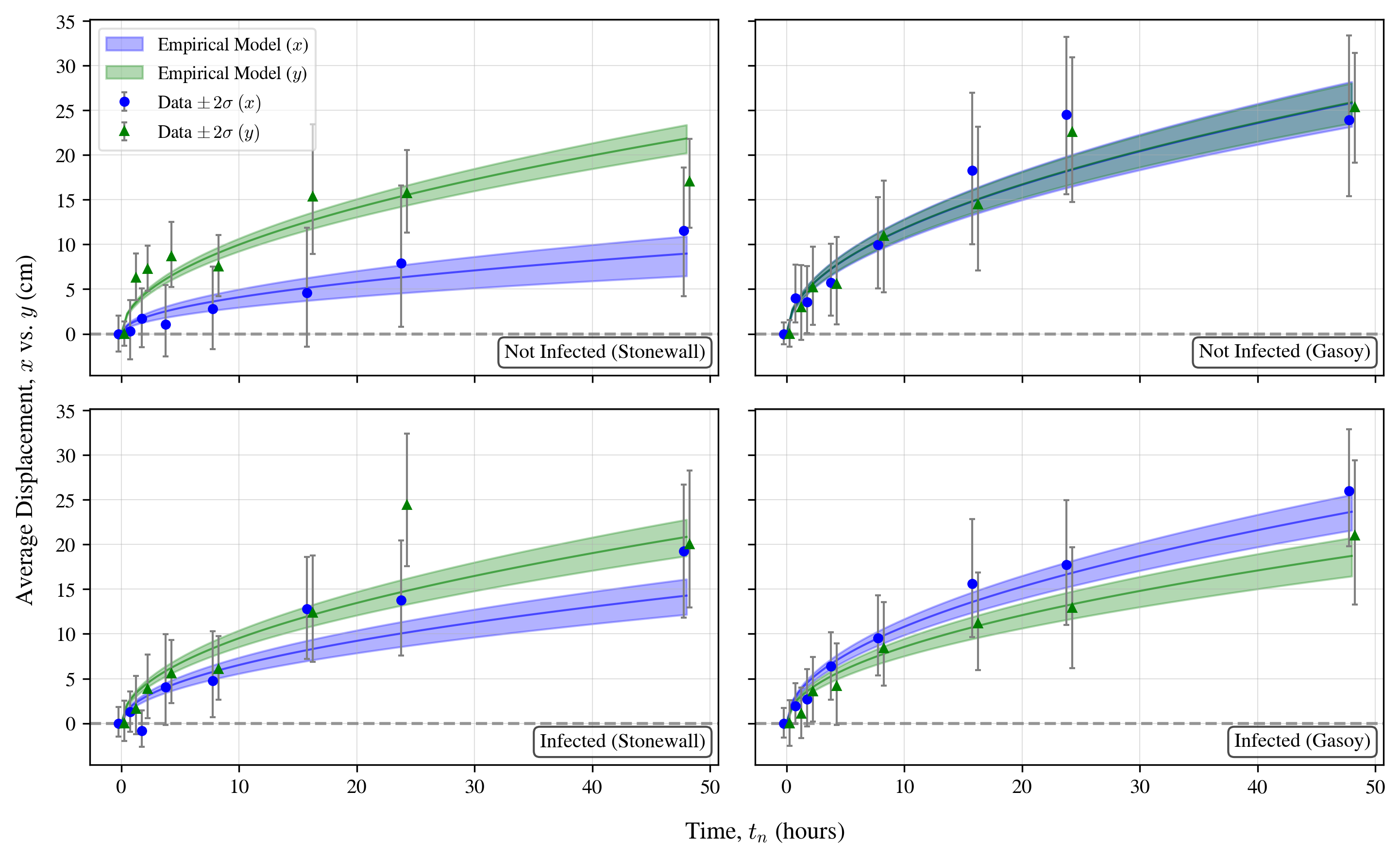}
    \caption{Comparison of $x$ and $y$ diffusion rates from the empirical data, using the same empirical models as in Figure~\ref{fig:cats_violin_xy}.}
    \label{fig:empirical_Dx_vs_Dy}
\end{figure*}

\end{document}